\documentclass{article}

\usepackage[eandd, final]{neurips_2026}

\usepackage[utf8]{inputenc} %
\usepackage[T1]{fontenc}    %
\usepackage{hyperref}       %
\usepackage{url}            %
\usepackage{booktabs}       %
\usepackage{amsfonts, amsmath}       %
\usepackage{nicefrac}       %
\usepackage{microtype}      %
\usepackage{xcolor}         %
\usepackage{hyperref}       %

\RequirePackage{algorithm}
\RequirePackage{algorithmic}
\usepackage{multirow}
\usepackage{graphicx}
\usepackage{tikz}
\usetikzlibrary{positioning,calc,fit,backgrounds,shadows.blur,arrows.meta}
\usepackage{booktabs} 
\usepackage{standalone}
\usepackage{pgfplots} %
\pgfplotsset{compat=1.18}

\usepackage{framed} %
\usepackage[table]{xcolor}
\usepackage[most]{tcolorbox}

\DeclareMathOperator{\R}{\mathbb{R}}
\DeclareMathOperator{\N}{\mathbb{N}}

\newcommand{\Loss}{\mathcal{L}}

\usepackage{siunitx}

\definecolor{cardframe}{rgb}{0.1, 0.1, 0.1}

\newtcolorbox{evalcard}[1][]{
  enhanced,
  breakable,
  colback=white,
  colframe=cardframe,
  title=\textbf{EvalCard: #1},
  fonttitle=\bfseries,
  coltitle=white,
  arc=8pt,
  width=14.75cm,
  boxrule=1pt,
  before skip=2pt, after skip=2pt,
  attach boxed title to top left={
    yshift=-3mm, xshift=5mm
  },
  boxed title style={
    size=small,
    colback=cardframe,
    colframe=cardframe,
    sharp corners=south
  },
}

\usepackage{cancel}

\title{PINNeval: A Comprehensive Evaluation Standard for Physics-Informed Neural Networks}

\author{%
  Kevin von Bargen\quad Hamid Ebrahimy\quad Tim Römer  \\
  Joint Lab Artificial Intelligence and Data Science,\\
  Osnabrück University, Osnabrück, Germany \\
  \texttt{\{kvonbargen,hamid.ebrahimy,troemer\}@uos.de} \\
}

\begin{document}

\maketitle

\begin{abstract}
The Physics-Informed Neural Network (PINN) literature has produced a rapidly expanding catalogue of methods.
We divide a PINN training pipeline systematically into four components and observe that many existing benchmarks study these components in isolation. These components are domain-importance adaptation, loss balancing, neural network architecture, and optimization scheme. Moreover, the studies set the surrounding and not considered components of the pipeline to a minimal default that is no longer representative of current best practice approaches.
Conclusions drawn under such conditions may not transfer to competitive pipelines.
In this work, we propose a new evaluation protocol that 
(1) treats the four components as an integrated system; 
(2) crucially requires competitive implementations of the components not under investigation; 
(3) specifies the metrics and reporting format; and 
(4) is accompanied by structured documentation that ensures the traceability of evaluations across studies.
Based on this comprehensive protocol, we release PINNpipe, an innovative JAX package containing 18 different methods from the current literature as well as nine efficiently implemented benchmark problems.
Using PINNpipe, we illustrate our evaluation protocol in relation to established benchmark studies and report results that challenge common assumptions.
\end{abstract}

\section{Introduction}

Physics-Informed Neural Networks (PINNs) reformulate the solution of partial differential equations (PDEs) as an optimization of a residual-based objective evaluated on a neural surrogate~\citep{Raissi.2019}.
PINNs are mesh-free, differentiable, and applicable to forward, inverse, and parametric formulations of the same governing equations.
These properties have driven substantial interest across scientific computing and engineering.
Their practical performance, however, is consistently limited by the difficulty of training. 
Slow convergence, sensitivity to hyperparameters, and stagnation at unfaithful solutions are routinely reported for stiff, multi-scale, or high-dimensional problems \citep{krishnapriyan2021characterizing}. 
In Section~\ref{experiments:optimizers}, we revisit their early observation that the bottleneck lies not in network expressivity but in the surrounding training pipeline under modern optimizers and architectures.
In response, the PINN literature has produced numerous methods that target components of a training pipeline. We divide such a pipeline  into four key components: 
\begin{itemize}
    \item 
    \emph{domain-importance adaptation}, which governs where in the problem domain residuals are evaluated and how individual locations are weighted;
    \item 
    \emph{loss balancing}, which governs how multiple loss terms are weighted during training;
    \item 
    \emph{neural network architecture}, which determines the hypothesis class of the neural surrogate;
    \item 
    \emph{optimization scheme}, which specifies the optimizers and their schedule.
\end{itemize}
Beyond these four components, advanced training techniques such as \emph{temporal marching}~\citep{Timemarching} and \emph{training curricula}~\citep{krishnapriyan2021characterizing} provide extensions for specific PDE families, 
in particular time-dependent problems on long horizons. However, these are optional rather than constitutive and are therefore omitted from our discussion.

Given the rapid growth of PINN method variants, a rigorous and standardized assessment is essential for meaningful comparisons and progress.
Several benchmark studies have been conducted in response.
Targeted studies examine individual components. For example, the architectures of PINNs and neural operators are studied in \citet{Pdenneval} 
and optimizers in \citet{SOAPWang}.
A more comprehensive study, PINNacle~\citep{PINNacle}, also covers diverse methods of a PINN training pipeline.
However, many efforts share a common methodological limitation.
Specifically, they evaluate methods with a surrounding pipeline that is set to a minimal default.
As a result, methods that pursue different objectives within the pipeline, and that are intended to be used jointly, are placed in direct comparison against one another.
PINNacle, the study most closely comparable to ours, exemplifies this pattern: 
loss-balancing methods and domain-importance adaptation methods are treated as a single methodological category and evaluated against each other rather than in combination, which conflates two of the four essential components into a single comparison axis.
Concretely, the training signal arising from any region of the domain is shaped jointly by an effective importance density and by the scaling of the associated loss terms. 
This density can be realized through denser sampling or through pointwise weighting.

In general, these limitations stem from two fundamental issues. First, the literature lacks a standardized definition of a PINN training pipeline that explicitly accounts for its four key components. 
Second, no comprehensive evaluation protocol establishes the prerequisites under which comparative claims about a method become meaningful and transferable across settings.

In the absence of such a protocol, benchmarking studies often fix the surrounding pipeline to a vanilla default and insert the candidate method into that setting. 
This setting is no longer representative of best practice, and conclusions drawn under it can deviate substantially from those obtained when the same method is evaluated within a competitive instance of the pipeline. 
See Section~\ref{sec:related} for details.

Against this background, we use a unified definition of the PINN optimization problem and propose a four-component decomposition of the related training pipeline. 
Building on this, we present an evaluation protocol that treats the pipeline as a coupled system. 
The protocol requires a competitive implementation of the components not currently under investigation.
It further requires that the configuration of the surrounding pipeline is specified along with each comparison result, 
and that metrics and reporting formats follow a fixed specification under which claims become directly comparable across studies. 
The formal definition of our pipeline and protocol is given in Section~\ref{sec:protocol}.

Our contributions can be summarized as follows:
\begin{itemize}
    \item 
    We present a four-component decomposition of a PINN training pipeline as an explicit basis for evaluation.
    \item 
    We propose PINNeval, an evaluation protocol that captures dependencies between components and requires a competitive rather than a minimal instantiation of the surrounding pipeline.
    \item
    We supplement the protocol with a structured documentation format that records the configuration and assumptions of an evaluation, so that comparative statements across different studies can be verified.
    \item 
    We apply the protocol in case studies covering all four components and obtain results that revise commonly held expectations about specific methods, 
    including negative results that contradict  assumptions underlying some earlier evaluations.
    \item 
    We implement the protocol in PINNpipe, a JAX package with composable modules for the four components and advanced training techniques, and efficient implementations of 18 methods together with nine benchmark problems.
\end{itemize}

The paper is organized as follows.
Section~\ref{sec:related} situates our contribution against existing benchmarks and component-specific studies.
Section~\ref{sec:protocol} formalizes the PINN training pipeline and the evaluation protocol.
Section~\ref{sec:experiments} demonstrates the protocol on case studies covering each component of the pipeline and introduces PINNpipe, our JAX package.
Section~\ref{sec:discussion} discusses the limitations of the present scope and possible extensions of the protocol to inverse and operator-learning settings.

\section{Related Work}
\label{sec:related}

The literature on benchmarking PINNs can be broadly divided into two main categories. 
The first category consists of comparative evaluations: works that survey multiple existing methods across a shared set of problems without proposing a new one. Their goal is to map the landscape of available components and characterize how they perform relative to one another.
Within this category, PDEBench \citep{Pdebench} provides a systematic evaluation over a broad range of PDEs, but it primarily targets neural operator methods. Here PINNs are included only as one method among many and at a limited level of detail. 
Similarly, PDENNEval \citep{Pdenneval} narrows the scope to a comparative study of the underlying neural network architectures used in PINNs and neural operators, but essentially leaves the other components of a PINN training pipeline outside its scope. 
PINNacle \citep{PINNacle}, while presented as a comprehensive benchmark, adopts an evaluation protocol whose design choices substantially restrict the conclusions that can be drawn from its results.
Each variant under investigation is evaluated as a single modification to the vanilla PINN baseline, with all other pipeline components left at their defaults.
Additionally, PINNacle groups loss-balancing and domain-importance adaptation methods into a single comparison category rather than treating them as complementary components designed to function jointly.
Conclusions drawn under this design do not reflect configurations representative of established practice like in \citet{wang2023expert}.

In the second category, research articles propose a new approach by fixing a surrounding pipeline of their own choice and compare the new method against existing alternatives for the same pipeline component. 
Within this category, several individual components of the PINN training pipeline have been considered.
For domain-importance adaptation, \citet{Toscano.2026} introduce a sampling rule whose objective coincides with that of RAD \citep{RAR}. 
However, this rule is combined with a GradNorm-based loss-balancing scheme of the type proposed in \citet{wang2023expert}, without comparison against RAD combined with the same loss-balancing scheme.
The empirical comparison therefore reflects the joint effect of two pipeline components, and any improvement over alternative pipelines cannot be attributed unambiguously to the proposed domain-importance adaptation rule. 
The study also reports the numerical precision used in other evaluations, but does not specify the precision used in its own pipeline.
The released code uses float64 when combined with SSBroyden, while the field's default is float32, a difference that has been shown to materially affect PINN convergence \citep{xu2025fp64}.
The reported improvements therefore cannot be attributed unambiguously to the proposed method, to the combination with the loss-balancing scheme, or to the precision regime. 
Also within the scope of domain-importance adaptation, \citet{jeong2025mitigating} provide a careful comparison of such methods. However, the reported errors indicate that the surrounding pipeline is not competitive with current practice. Conclusions about the relative merits of domain-importance adaptation methods, when drawn under such a pipeline, therefore have limited transferability.
For the loss balancing component, \citet{relobralo} introduce ReLoBRaLo, adopt a competitive surrounding pipeline, and provide a comparison of weighting schemes, but omit the well-established GradNorm-based method proposed by \citet{wang2023expert}. 
As a result, their comparison covers only a subset of the loss-balancing methods relevant to a comprehensive evaluation of this pipeline component.
For optimization schemes, \citet{SOAPWang} present a thorough evaluation of quasi-second-order optimizers. 
However, quasi-Newton methods are excluded on the grounds that they rely on full-batch training, float64 precision, and small network sizes, and are therefore claimed to be “impractical for the complex PDE benchmarks considered in this work.” 

Although each of these studies contributes valuable insights within its chosen scope, the recurring pattern across both categories is that conclusions are drawn under pipeline configurations that are not representative of established practice, omit relevant competing methods, or leave key implementation details unspecified. 
As a result, it remains difficult to assess the relative merits of individual pipeline components, and the reported improvements cannot be clearly attributed to the methodological contribution under investigation.
Taken together, the limitations of both categories motivate the need for a comprehensive evaluation standard for PINNs. 
Our proposed framework, PINNeval, not only compares methods across benchmark problems, but also explicitly accounts for the modular structure of the PINN training pipeline. 
In particular, PINNeval provides a transparent evaluation protocol, documents all relevant implementation and numerical details, and assesses methods under competitive pipeline configurations.
\clearpage
\section{Evaluation Protocol}
\label{sec:protocol}
\subsection{General PINN Framework}
\label{sec:general_pinn_framework}
Let $\Omega \subset \mathbb{R}^d$ be a spatial domain with boundary
$\partial \Omega$, let $T \in \mathbb{R}_{\geq 0}$, and let
$\mathcal{N}[\cdot]$ and $\mathcal{B}[\cdot]$ denote a differential
operator and a boundary operator, respectively. For a given initial
condition $g : \Omega \to \mathbb{R}$ and boundary condition
$h : [0, T] \times \partial \Omega \to \mathbb{R}$, an
\emph{initial-boundary value problem for time-dependent PDEs (IBVP)}
seeks a solution $u : [0, T] \times \Omega \to \mathbb{R}$ satisfying
the \emph{governing equation}
\begin{equation}
    \partial_t u(t, x) + \mathcal{N}[u](t, x) = 0,
    \quad (t, x) \in (0, T] \times \Omega, \label{eq:PDE}
\end{equation}
the \emph{initial condition}
\begin{equation}
    u(0, x) = g(x), \quad x \in \Omega,
\end{equation}
and the \emph{boundary condition}
\begin{equation}
    \mathcal{B}[u](t, x) = h(t, x),
    \quad (t, x) \in [0, T] \times \partial \Omega.
\end{equation}
The operator $\mathcal{N}$ may be linear or nonlinear, and $\mathcal{B}$
corresponds to a Dirichlet, Neumann, Robin, or periodic boundary condition.

The PINN approach approximates $u(t, x)$ by a neural network $u_\theta(t, x)$. Training $u_\theta$ on an IBVP requires two formal ingredients: subsets of the respective domains on which the equations are enforced, and training objectives that act on those subsets.

Let $\mathcal{E}$ denote the set of equations of the IBVP, comprising the governing equation, the initial condition, and the boundary condition. 
To each equation $e \in \mathcal{E}$, we associate a \emph{residual function} $r_e(t, x; \theta)$ that measures the pointwise violation of $e$ by $u_\theta$. 
Each equation is enforced on a subset $D_e \subset [0, T] \times \Omega$, e.g., an initial condition is enforced on $D_{\text{init}} = \{0\} \times \Omega$. 
From these domain sets $D_e$, collocation sets $C_e$ are constructed or sampled to evaluate the residuals $r_e(t,x;\theta)$. The loss of equation $e$ for $u_\theta$ is defined as
\begin{equation}
\Loss_e(\theta)
=  \frac{1}{|C_e|}
    \sum_{(t,x)\in C_e}
      r_e(t,x;\theta)^2,
\end{equation}
and the composite training objective aggregates the \emph{physics-informed losses}:
\begin{equation}
\Loss(\theta)
  = \sum_{e\in\mathcal{E}}
      \Loss_e(\theta).
\end{equation}
Despite correctly capturing the learning objective, this formulation is
known to suffer from poor convergence and training
instabilities~\citep{ModifiedMLP}. To prepare for the four-component
decomposition that follows, we extend the basic objective to admit
pointwise and equation-level weighting. Given pointwise weights
$\omega(t, x)$ for $(t, x) \in C_e$ and equation-level weights
$\lambda_e$ for $e \in \mathcal{E}$, the training objective becomes
\begin{subequations}\label{eq:loss}
\begin{equation}
\Loss(\omega, \lambda, \theta)
= \sum_{e\in\mathcal{E}} \lambda_e \, \Loss_e(\omega,\theta),
\label{eq:loss-total}
\end{equation}
\begin{equation}
\Loss_e(\omega, \theta)
= \frac{1}{|C_e|} \sum_{(t,x)\in C_e}
  \omega(t,x)\, r_e(t,x;\theta)^2.
\label{eq:loss-e}
\end{equation}
\end{subequations}

\subsection{The Four-Component Decomposition of the PINN Training Pipeline}
\label{sec:compon}

Two of the four PINN training pipeline components act directly on this loss structure.
The first modifies the per-equation loss~\eqref{eq:loss-e} by adapting
the pointwise weights $\omega(t, x)$ and the sampled collocation set
$C_e$. We refer to this component as \emph{domain-importance adaptation}.
The second modifies the composite loss~\eqref{eq:loss-total} through
the equation-level weights $\lambda_e$ that balance the contributions
of different residuals. We refer to this as \emph{loss balancing}.

\paragraph{Domain-Importance Adaptation.}
Within this component, two core approaches have emerged for adaptively modifying the pointwise weights $\omega$ or the collocation sets $C_e$:
\begin{enumerate}
    \item \textbf{Adaptive sampling:} Methods that modify the sets $C_e$ by adding or resampling points from the domains $D_e$, typically biased toward high-residual regions.
    \item \textbf{Importance weighting:} Methods that attach dynamic, trainable, or scheduled weights to individual points or segments of the domain.
\end{enumerate}
\newpage
This unified view is reinforced by \citet{Toscano.2026}, who show that within domain-importance adaptation, the choice between adaptive sampling and importance weighting can be computational rather than conceptual. Established methods such as RAD, an adaptive sampling scheme, and RBA \citep{RBA-PINN}, an importance weighting scheme, emerge as alternative realizations of the same underlying variational objective.

\paragraph{Loss Balancing.}
This component addresses the multi-task nature of the composite loss $\Loss(\theta)$, where individual loss terms can differ by orders of magnitude in their gradient norms and pull the optimizer in conflicting directions. Methods in this component adjust the scalar weights $\lambda_e$ through different rules.

\begin{figure}[htbp]
  \centering
  \definecolor{cDIA}{HTML}{0072B2}   %
\definecolor{cArch}{HTML}{E69F00}   %
\definecolor{cLB}{HTML}{009E73}   %
\definecolor{cOpt}{HTML}{56B4E9}  %

\begin{tikzpicture}[
    neuron/.style={circle, draw=black!55, line width=0.4pt, minimum size=0.7cm, fill=white},
    input/.style ={circle, draw=black!55, line width=0.4pt, minimum size=0.7cm, fill=white, font=\small},
    bias/.style  ={circle, draw=black!55, line width=0.4pt, minimum size=0.7cm, fill=white, font=\small},
    cbox/.style={
        rounded corners=4pt, draw=#1, line width=1pt, fill=#1!7,
        align=center, inner sep=10pt,
        blur shadow={shadow xshift=0pt, shadow yshift=-1pt, shadow opacity=15},
    },
    auxbox/.style={
        rounded corners=4pt, draw=black!45, line width=0.6pt, fill=black!3,
        align=center, inner sep=8pt,
    },
    chead/.style={font=\bfseries\footnotesize, text=#1},
    arrow/.style={->, line width=0.7pt, >=Stealth, draw=black!75},
    feedbackarrow/.style={->, line width=0.7pt, >=Stealth, draw=black!75, dashed},
    connection/.style={-, thin, draw=black!25},
]

\node[cbox=cDIA, minimum width=4.6cm] (Data) at (-5.0, -0.28) {%
    \emph{Adaptive Sampling}\\[2pt]
    $\{C_e\}_{e\in\mathcal{E}}$\\
    {\footnotesize collocation sets}\\[12pt]
    \emph{Importance Weighting}\\[2pt]
    $\omega(t,x)$\\
    {\footnotesize pointwise weights}};
\node[chead=cDIA, anchor=south] at ([yshift=4pt]Data.north)
    {Domain-Importance Adaptation};

\node[input] (x1) at (-0.5, 2)  {$x_1$};
\node[input] (x2) at (-0.5, 0)  {$x_2$};
\node[input] (x3) at (-0.5,-2)  {$x_3$};
\node[neuron] (h1) at (1.5, 2.5) {};
\node[neuron] (h2) at (1.5, 1.5) {};
\node at (1.5, 0.5) {$\vdots$};
\node[neuron] (h3) at (1.5,-0.5) {};
\node[neuron] (h4) at (1.5,-1.5) {};
\node[bias]   (h_bias) at (1.5,-2.5) {+1};
\node[neuron] (h2_1) at (3.5, 2.5) {};
\node[neuron] (h2_2) at (3.5, 1.5) {};
\node at (3.5, 0.5) {$\vdots$};
\node[neuron] (h2_3) at (3.5,-0.5) {};
\node[neuron] (h2_4) at (3.5,-1.5) {};
\node[bias]   (h2_bias) at (3.5,-2.5) {+1};
\node[neuron] (o1) at (5.5, 1) {$u_1$};
\node[neuron] (o2) at (5.5,-1) {$u_2$};
\foreach \i in {1,2,3} {\foreach \j in {1,2,3,4} {\draw[connection] (x\i) -- (h\j);}}
\foreach \i in {1,2,3,4} {\foreach \j in {1,2,3,4} {\draw[connection] (h\i) -- (h2_\j);}}
\foreach \j in {1,2,3,4} {\draw[connection] (h_bias) -- (h2_\j);}
\foreach \i in {1,2,3,4} {\draw[connection] (h2_\i) -- (o1); \draw[connection] (h2_\i) -- (o2);}
\draw[connection] (h2_bias) -- (o1);
\draw[connection] (h2_bias) -- (o2);

\begin{scope}[on background layer]
    \node[cbox=cArch, fit=(x1)(x3)(o1)(o2)(h_bias)(h2_bias),
          inner sep=12pt, name=NNbox]{};
\end{scope}
\node[chead=cArch, anchor=south] at ([yshift=4pt]NNbox.north)
    {Neural Architecture};

\node[cbox=cLB, minimum width=5.8cm, minimum height=2.9cm] (LB) at (11.25, -0.28) {%
    $\{\mathcal{L}_e(\theta)\}_{e\in\mathcal{E}}
     \quad\text{and}\quad
     \{\lambda_e\}_{e\in\mathcal{E}}$\\[2pt]
    {\footnotesize per-equation losses\quad\;equation-level weights}\\[10pt]
    $\displaystyle\mathcal{L}(\theta;\lambda)
     = \sum_{e\in\mathcal{E}} \lambda_e\,\mathcal{L}_e(\theta)$};
\node[chead=cLB, anchor=south] at ([yshift=4pt]LB.north)
    {Loss Balancing};

\node[cbox=cOpt, minimum width=6.2cm, minimum height=1.9cm] (Opt) at (2.52, -5.2) {%
    \emph{Update Rules}\\[4pt]
    $\theta_{k+1} = \theta_k - \alpha_k\, \nabla\mathcal{L}(\theta_k)$\\
    $\theta_{k+1} = \theta_k - \alpha_k\, H_k^{-1}\, \nabla\mathcal{L}(\theta_k)$};
\node[chead=cOpt, anchor=south] at ([yshift=4pt,xshift=-5.2pt]Opt.north)
    {Optimizer Scheme};

\draw[arrow] (Data.east)   -- (NNbox.west);
\draw[arrow] (NNbox.east)  -- (LB.west);
\draw[arrow] (LB.south) |- (Opt.east);
\draw[arrow] (Opt.north) -- (NNbox.south);
\end{tikzpicture}
  \caption{Schematic of the four-component PINN training pipeline.\label{fig:nn}}
\end{figure}
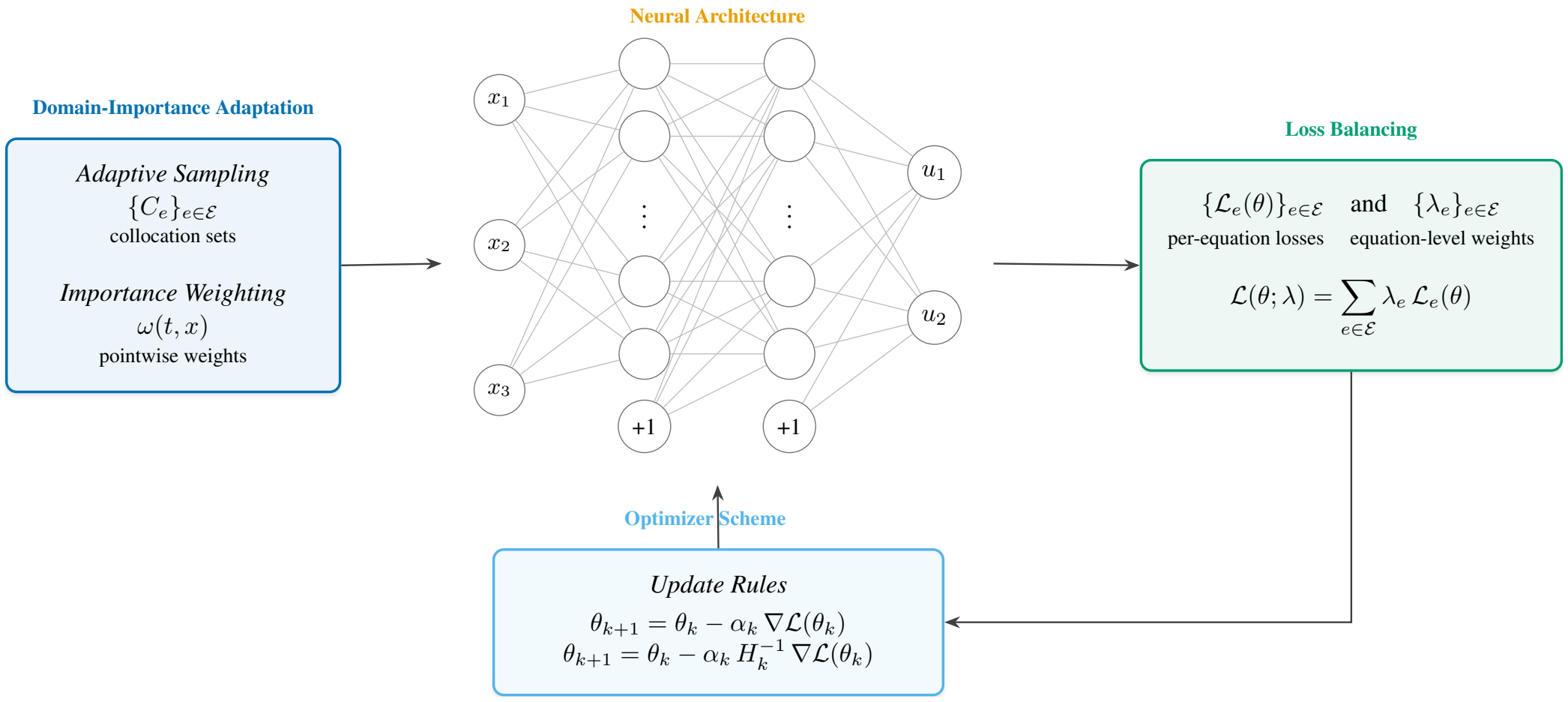

These two components distinguish a PINN training pipeline from standard machine learning pipelines. 
The remaining two components are the \textbf{neural network architecture}, which acts as the neural surrogate $u_\theta$, and the \textbf{optimization scheme} that minimizes $\Loss(\theta)$. They are shared with general machine learning practice and are discussed in Appendix~\ref{Appendix:neural_networks} and Appendix~\ref{Appendix:optimizer}, respectively. 
All methods considered in this paper are embedded into this framework. See Appendix~\ref{Appendix:dia} for domain-importance adaptation and Appendix~\ref{Appendix:lossbalancing} for loss balancing. An overview of the full pipeline is shown in Figure~\ref{fig:nn}.

\subsection{PINNeval: A Fair and Competitive Evaluation Protocol}

Building on the four-component decomposition described in Section~\ref{sec:compon} and the limitations identified in Section~\ref{sec:related}, the proposed protocol is guided by three main principles:

\begin{enumerate}
    \item 
    \textbf{Competitive surrounding pipeline:} 
    PINNeval is built around the principle that a method introduced in one component is evaluated within competitive choices of the remaining components. 
    A component choice is competitive at the time of evaluation if the evidence does not show an established alternative that consistently achieves either better accuracy at comparable computational cost or comparable accuracy at lower cost. 
    The evidence may come from the literature or from controlled benchmark experiments, and it must refer to a problem class comparable to that of the evaluation and use the same metrics as the evaluation. 
    Given the four-component decomposition, checking a PINN training pipeline against this definition amounts to four focused checks, one per component. 
    
    \item 
    \textbf{Coupled evaluation of components:} Methods in different components of the PINN training pipeline interact with each other. 
    Consequently, comparisons that vary one component while holding the others fixed are conditional on the surrounding configuration, i.e., the choices of the remaining three components.  
    The PINNeval protocol therefore requires the configuration of all four components to be reported with every comparative result, making it possible to assess the conditions under which a conclusion was obtained.
    The competitiveness requirement of the first principle makes the estimated conditional effect informative about the compared methods.
    \item 
    \textbf{Cost reporting beyond iteration counts:} Iteration counts alone are insufficient as a cost metric, since different methods are not always directly comparable. For example, the cost of a quasi-Newton step is not directly comparable to that of an Adam step, and a parameter-rich architecture incurs a higher per-iteration cost than a small Multilayer Perceptron (MLP). Accordingly, in the PINNeval protocol the runtime is included alongside the iteration counts and the hardware used, in the spirit of \citet{McGreivy.2024}.
\end{enumerate}

\subsection{Protocol Specification}

Under the protocol, the following items are specified, controlled, and reported, organized into per-component and evaluation-level requirements.

\paragraph{Per-Component Requirements.}
\begin{enumerate}
    \item \textbf{Model specification:}
    Network topology, depth, width, activation functions, and feature mappings (e.g., Fourier features).

    \item \textbf{Optimization scheme:}
    Optimizer or sequence of optimizers, learning-rate schedule, optimizer-specific hyperparameters, stopping criterion, and total training budget.

    \item \textbf{Loss balancing:}
    Method for balancing the equation-level loss terms via the weights $\lambda_e$, update frequency of the balancing scheme, and method-specific hyperparameters.

    \item \textbf{Domain-importance adaptation:}
   Adaptive sampling or importance weighting method, sampling distribution and collocation set sizes per equation, update frequency of the method, and method-specific hyperparameters.
\end{enumerate}

\paragraph{Evaluation-Level Requirements.}
\begin{enumerate}\setcounter{enumi}{4}
    \item \textbf{Claim under evaluation:}
    Each evaluation begins with an explicit statement of the claim under investigation and a specification of the PDE problems used to support it. The full PDE specification is referenced within the protocol.

    \item \textbf{Comparison conditions:}
    Each evaluation specifies the baselines used and the justification for their selection, the components held fixed across the comparison, and an explicit statement that the components not currently under evaluation are instantiated competitively.
    
    \item \textbf{Metrics:}
    The relative $L^2$ error is reported as the error metric, and runtime alongside iteration count as the computational-cost metric. They are consistently reported across all compared methods.

    \item \textbf{Numerical precision:}
    The floating-point precision used (e.g., float32 or float64) is reported consistently across all compared methods.

    \item \textbf{Robustness and statistical reporting:} The number of independent runs, the variability across runs, any observed instability or non-convergence, and, if a method is highlighted as best, the significance test supporting it are reported.

    \item \textbf{Reproducibility and limitations:}
    The protocol reports all implementation-relevant settings required for reproduction, the surrounding-pipeline configurations that were tested, and the conditions under which the conclusions do not generalize.
\end{enumerate}

\subsection{The EvalCard Documentation Format}
\label{sec:Evalcard}
To strengthen consistent and comparable reporting of evaluation setups, a unified description format is needed. Building on the card-based reporting style of \citet{dhar2025evalcardsframeworkstandardizedevaluation}, we introduce a dedicated \emph{EvalCard} for the proposed evaluation protocol. 
Each card summarizes, in a uniform layout, the methods compared, the architectural and optimization choices held constant across all benchmarks, and the evaluation-level conditions specified by the protocol. 
Benchmark-specific configurations that are independent of the compared methods, such as feature scales, training budget, and advanced training techniques, are reported alongside the card to avoid visual clutter.

A template is given in Appendix~\ref{Appendix:Evalcard}; filled-in cards are provided in Figures~\ref{fig:evalcard_dia}, ~\ref{fig:evalcard_optimizer}, ~\ref{fig:evalcard_loss_balancing}, and~\ref{fig:evalcard_nn}.

\section{Experiments}
\label{sec:experiments}

We conduct three evaluation studies, each targeting one component of the PINN training pipeline as well as one optimizer-and-network study, a coupling of these two components.
One MLP with random Fourier features serves as the common anchor of all four studies.
Performance differences in every study are therefore attributable to the component under evaluation and its surrounding pipeline rather than to an uncontrolled change in network expressivity.
A dedicated study of neural network topology and size is provided in Appendix~\ref{Appendix:nn_study}. 

\subsection{Domain-Importance Adaptation Study}
\label{sec:experiments-dia}
This study addresses the two gaps left by \citet{Toscano.2026} and \citet{jeong2025mitigating}: placing vRBA against a representative selection of domain-importance adaptation methods, and evaluating LAS within a competitive surrounding pipeline. Within a fixed pipeline, we compare Random-R, LAS, vRBA, R3~\citep{r3}, and RAD. 
Method definitions, pseudocode, and hyperparameters used are provided in Appendix~\ref{Appendix:dia}.
\begin{figure}[htbp]
  \centering
  \begin{evalcard}[Domain-Importance Adaptation]
\vspace{0.5em}
\textbf{Claim under Evaluation.}
The choice of domain-importance adaptation method materially affects the
final relative $L^2$ error of a PINN when evaluated on a competitive
surrounding pipeline. Tested on a representative subset of competitive benchmarks ranging from low-dimensional to high-dimensional and coupled-system PDE problems.

\vspace{1em}

\textbf{Surrounding Pipeline.}\vspace{0.5ex}
\begin{center}
\rowcolors{2}{gray!10}{white}
\begin{tabular}{lll}
\toprule
\textbf{Component} & \textbf{Choice} & \textbf{Hyperparameters} \\
\midrule
Architecture   & MLP, depth 5, hidden dim 64, Fourier features
               & Table~\ref{tab:nn_settings} \\
Loss balancing & GradNorm-based, Frequency: $1000$
               & Appendix~\ref{Appendix:gradnorm} \\
Optimizer      & SOAP
               & Table~\ref{tab:general_opt_settings} \\
\bottomrule
\end{tabular}
\end{center}
\vspace{1em}

\textbf{Domain-Importance Adaptation.}
\begin{center}
\rowcolors{2}{gray!10}{white}
\begin{tabular}{llllll}
\toprule
Hyperparameter & Random-R & vRBA & LAS & R3 & RAD \\
\midrule
Sample size &  \multicolumn{5}{c}{8192} \\
Frequency &  \multicolumn{5}{c}{1} \\
Additional Hyperparameters  &  \multicolumn{5}{c}{See defaults in Appendix~\ref{Appendix:dia}} \\
\bottomrule
\end{tabular}
\end{center}

\vspace{0.5em}

\textbf{Evaluation Conditions.}
Float32 precision, NVIDIA H100, 10 seeds. The median values are reported. 
The standard deviation and results for all nine benchmarks are provided in Appendix~\ref{appendix:dia_full}.
Significance test (bold): exact two-sided Mann–Whitney $U$ over seeds, Holm-corrected, $\alpha = 0.05$.
\vspace{1em}

\textbf{Results.}

\vspace{0.5ex}
\rowcolors{2}{gray!10}{white}
\resizebox{\textwidth}{!}{
\begin{tabular}{lcccccccc}
& \multicolumn{4}{c}{\textbf{Relative $L^2$ error}}
& \multicolumn{4}{c}{\textbf{Runtime [min]}} \\
\cmidrule(lr){2-5}\cmidrule(lr){6-9}
\textbf{Method} & Wave & KdV & RT & Heat & Wave & KdV & RT & Heat \\
\midrule
Random-R & $7.93\times 10^{-5}$ & $\mathbf{4.54\times 10^{-4}}$ & $\mathbf{1.68\times 10^{-2}}$ & $3.90\times 10^{-4}$ & $10.5$ & $10.1$ & $29.8$ & $5.4$ \\
LAS     & ${7.85\times 10^{-5}}$ & $5.05\times 10^{-4}$ & $2.30\times 10^{-2}$ & $\mathbf{2.39\times 10^{-4}}$ & $11.2$ & $11.9$ & $34.8$ & $7.0$ \\
vRBA    & $1.70\times 10^{-4}$ & $8.39\times 10^{-4}$ & $2.27\times 10^{-2}$ & $1.21\times 10^{-3}$ & $11.3$ & $11.1$ & $31.6$ & $5.8$ \\
R3      & $2.55\times 10^{-4}$ & $1.43\times 10^{-3}$ & $2.71\times 10^{-2}$ & $5.64\times 10^{-4}$ & $11.0$ & $11.0$ & $31.1$ & $6.0$ \\
RAD     & $2.02\times 10^{-4}$ & $1.61\times 10^{-3}$ & $5.94\times 10^{-2}$ & $2.02\times 10^{-3}$ & $12.2$ & $13.0$ & $34.9$ & $7.0$ \\
\end{tabular}}
\end{evalcard}

  \caption{EvalCard for the domain-importance adaptation study.}
  \label{fig:evalcard_dia}
\end{figure}

Figure~\ref{fig:evalcard_dia} summarizes the results.
The runtime overhead of all domain-importance adaptation methods is small relative to Random-R, so the choice of a domain-importance adaptation method is not notably constrained by cost. 
In terms of accuracy, the picture is more nuanced. Random-R sampling is best on KdV and RT and lies within a small factor of the best on Wave and Heat. 
LAS matches or outperforms Random-R on Wave and Heat, confirming that its benefits transfer to a competitive pipeline, although the margin is small and inconsistent. 
The remaining three methods mostly perform worse than Random-R and LAS on these four PDEs, with RAD trailing on most of them. 
We note that the adaptive methods run at the hyperparameter values proposed by the authors who introduced them. In their original studies, these methods match or outperform Random-R within surrounding pipelines built on other optimizers, in particular Adam, whereas this study uses SOAP.
Full results across all nine PDEs as well as the standard deviation between different seeds are reported in Appendix~\ref{appendix:dia_full}.

\subsection{Optimizer and Neural Network Study}
\label{experiments:optimizers}
This study compares two pipelines that differ in optimization scheme, numerical precision, and neural network architecture, in order to test the claim of \citet{SOAPWang}. The results in Figure~\ref{fig:evalcard_optimizer} refute this claim, as the two-stage MLP pipeline (SOAP followed by SSBroyden2) attains both a lower median relative $L^2$ error and a markedly shorter runtime than the single-stage PirateNet-SOAP baseline.

\begin{figure}[h!]
  \centering
  \begin{evalcard}[Neural Network and Optimizer]
\vspace{0.5em}
\textbf{Claim under Evaluation.}
Small neural networks with float64 precision and quasi-Newton optimizers are impractical for complex PDE benchmarks such as the Rayleigh--Taylor instability or the Kuramoto--Sivashinsky equation considered in our benchmark suite.

\vspace{1em}

\textbf{Surrounding Pipeline.}\vspace{0.5ex}
\begin{center}
\rowcolors{2}{gray!10}{white}
\begin{tabular}{lll}
\toprule
\textbf{Component} & \textbf{Choice} & \textbf{Hyperparameters} \\
\midrule
Loss balancing & GradNorm-based, Frequency: $1000$ & Appendix~\ref{Appendix:gradnorm} \\
Domain-Importance Adaptation      &  LAS, Sample size: $8192$, Frequency: $1$ & Appendix~\ref{Appendix:las}\\
\bottomrule
\end{tabular}
\end{center}

\vspace{0.5em}

\textbf{Neural Network Architectures and Optimizer Schemes.}
\begin{center}
\rowcolors{2}{gray!10}{white}
\begin{tabular}{lllllll}
\toprule
\textbf{Label} &\textbf{Architecture} & \textbf{Depth} & \textbf{Width} & \textbf{Opt. scheme}  & \textbf{Optimizers}  & \textbf{Precision}   \\
\midrule
P1 & PirateNet & 3 & 256 & One-stage & SOAP & float32 \\
P2 & MLP & 5 & 64 & Two-stage & SOAP \& SSBroyden2 & float64 \\
\bottomrule
\end{tabular}
\end{center}
Additional neural network settings are provided in Table~\ref{tab:nn_settings}. Optimizer settings for both pipelines are provided in Table~\ref{tab:optimizer_study_settings}.

\vspace{0.5em}

\textbf{Evaluation Conditions.}
NVIDIA H100, 5 seeds. Reported variability: median $\pm$ standard deviation. 
Significance test (bold): exact two-sided Mann–Whitney $U$ over seeds, Holm-corrected, $\alpha = 0.05$.
\vspace{1em}

\textbf{Results.}\vspace{0.5ex}

\begin{center}
\rowcolors{2}{gray!10}{white}
\begin{tabular}{lcccc}
& \multicolumn{2}{c}{\textbf{Relative $L^2$ error}}
& \multicolumn{2}{c}{\textbf{Runtime [min]}} \\
\cmidrule(lr){2-3}\cmidrule(lr){4-5}
Benchmark  & P1 & P2 & P1 & P2 \\
\midrule
Wave & ${2.82 \pm 0.8 \times 10^{-5}}$ & $\mathbf{{2.36 \pm 0.8 \times 10^{-7}}}$ & $111.22$ & $\mathbf{4.65}$ \\
KdV & ${1.93 \pm 0.1 \times 10^{-4}}$ & $\mathbf{{2.31 \pm 1.2 \times 10^{-5}}}$ & $88.16$ & $\mathbf{7.39}$ \\
RT & ${8.80 \pm 0.8 \times 10^{-3}}$ & ${{8.61 \pm 0.1 \times 10^{-3}}}$ & $513.50$ & $\mathbf{70.27}$ \\
Heat & ${2.62 \pm 0.2 \times 10^{-4}}$ & $\mathbf{{1.25 \pm 0.6 \times 10^{-8}}}$ & $49.39$ & $\mathbf{7.29}$ \\
AC & ${2.04 \pm 0.5 \times 10^{-5}}$ & $\mathbf{{9.20 \pm 3.4 \times 10^{-6}}}$ & $297.46$ & $\mathbf{9.68}$ \\
Burgers & ${2.51 \pm 0.9 \times 10^{-5}}$ & $\mathbf{{8.59 \pm 10.8 \times 10^{-8}}}$ & $41.56$ & $\mathbf{3.10}$ \\
GL & ${4.77 \pm 1.6 \times 10^{-3}}$ & ${{3.68 \pm 1.2 \times 10^{-3}}}$ & $519.22$ & $\mathbf{18.02}$ \\
Poisson & ${4.82 \pm 0.4 \times 10^{-6}}$ & $\mathbf{{1.14 \pm 0.4 \times 10^{-7}}}$ & $25.30$ & $\mathbf{5.04}$ \\
KS & ${4.51 \pm 0.7 \times 10^{-2}}$ & $\mathbf{{7.26 \pm 3.6 \times 10^{-3}}}$ & $1100.85$ & $\mathbf{70.33}$ \\
\end{tabular}
\end{center}
\end{evalcard}

  \caption{EvalCard for the optimizer and neural network study.}
  \label{fig:evalcard_optimizer}
\end{figure}

We note that the MLP of the two-stage pipeline is the anchor MLP. 
In a single-axis ablation (Appendix~\ref{Appendix:ablation_nn_optim}), float64 precision alone yields no measurable gain, and replacing PirateNet by the MLP under SOAP alone mostly increases the error. 
Adding the quasi-Newton stage to the same MLP at the same precision, however, reduces the error by up to a factor of $1.7 \times 10^4$ (Heat) and accounts for the accuracy gain of the two-stage pipeline. 
The larger error of the MLP under SOAP alone is therefore a trainability limit of single-stage SOAP optimization, not a capacity limit. 
Float64 precision is a prerequisite of the quasi-Newton stage \citep{SOAPWang} rather than an independent contributor. 
The moderate size of the MLP makes this stage feasible, since SSBroyden2 stores a dense inverse-Hessian approximation with $O(d^2)$ memory for $d$ trainable parameters.
Per-stage iteration counts and additional hyperparameters are reported in Table~\ref{tab:optimizer_study_settings} and the peak VRAM usage in Appendix~\ref{Appendix:vram_results}.

We further highlight the efficiency of our implementation on the more demanding PDEs: on Ginzburg--Landau, the PirateNet-SOAP pipeline in \texttt{jaxpi} required $746.0$ minutes on our hardware, whereas PINNpipe completes the same benchmark in $519.2$ minutes despite employing the more expensive LAS domain-importance adaptation in place of Random-R. Comparable speed-ups are also observed on the Rayleigh--Taylor benchmark.

\subsection{Loss Balancing Study}
\label{sec:experiments-lb}
This study closes the gap left by \citet{relobralo}, whose comparison of ReLoBRaLo against several adaptive schemes omits the GradNorm-based loss balancing of \citet{wang2023expert}. 
The evaluation conditions, network architecture, and optimizer specified in Figure~\ref{fig:evalcard_dia} are used throughout, with Random-R as the domain-importance adaptation method.

A  filled-in EvalCard is provided in Figure~\ref{fig:evalcard_loss_balancing}.
Method definitions, pseudocode, and hyperparameters used are given in Appendix~\ref{Appendix:lossbalancing}. 
We compare GradNorm-based balancing \citep{wang2023expert}, ReLoBRaLo, SoftAdapt~\citep{SoftAdapt}, GradNorm-MTL~\citep{Gradnorm_mtl}, and a fixed-weight baseline.

\begin{figure}[htbp]
  \centering
  \centering
\definecolor{cGN}{HTML}{0072B2}   %
\definecolor{cSA}{HTML}{E69F00}   %
\definecolor{cRL}{HTML}{56B4E9}   %
\definecolor{cFX}{HTML}{009E73}   %
\definecolor{cMT}{HTML}{F0E442}   %

\begin{tikzpicture}
\begin{axis}[
    ybar=1pt,
    bar width=4.5pt,
    width=1\textwidth,
    height=6.0cm,
    xtick pos=bottom,
    ymode=log,
    ymin=2e-6, ymax=20,
    ylabel={Relative $L^2$ error},
    symbolic x coords={Wave, AC, KS, Poisson, Burgers, GL, Heat, KdV, RT},
    xtick=data,
    enlarge x limits=0.1,
    ymajorgrids=true,
    ylabel near ticks,
    major grid style={dashed, gray!30},
    tick label style={font=\small},
    label style={font=\small},
    legend style={
        font=\small,
        at={(0.5,-0.15)}, anchor=north,
        legend columns=5,
        draw=none,
        /tikz/every even column/.append style={column sep=0.6em},
    },
    legend cell align=left,
    error bars/y dir=both,
    error bars/y explicit,
    every axis plot/.append style={fill opacity=0.85, draw=black!60, line width=0.3pt},
]

\addplot+[fill=cGN] coordinates {
  (Wave,    7.93e-5) +- (0, 2.6e-5)
  (AC,      3.52e-5) +- (0, 0.9e-5)
  (KS,      8.43e-2) +- (0, 4.6e-2)
  (Poisson, 2.93e-5) +- (0, 0.6e-5)
  (Burgers, 7.21e-6) +- (0, 2.2e-6)
  (GL,      4.06e-2) +- (0, 0.5e-2)
  (Heat,    3.90e-4) +- (0, 0.4e-4)
  (KdV,     4.54e-4) +- (0, 0.1e-4)
  (RT,      1.68e-2) +- (0, 0.2e-2)
};

\addplot+[fill=cSA] coordinates {
  (Wave,    4.15e-2) +- (0, 0.3e-2)
  (AC,      2.48e-4) +- (0, 1.8e-4)
  (KS,      1.45e0 ) +- (0, 0.1e0 )
  (Poisson, 6.38e-5) +- (0, 1.1e-5)
  (Burgers, 1.17e-5) +- (0, 0.2e-5)
  (GL,      1.75e-1) +- (0, 0.3e-1)
  (Heat,    3.49e-4) +- (0, 0.3e-4)
  (KdV,     1.05e-3) +- (0, 0.1e-3)
  (RT,      1.92e-2) +- (0, 0.5e-2)
};

\addplot+[fill=cRL] coordinates {
  (Wave,    1.61e-3) +- (0, 0.2e-3)
  (AC,      1.76e-4) +- (0, 1.2e-4)
  (KS,      1.47e-1) +- (0, 1.4e-1)
  (Poisson, 3.57e-5) +- (0, 0.9e-5)
  (Burgers, 1.02e-5) +- (0, 0.2e-5)
  (GL,      5.70e-2) +- (0, 1.0e-2)
  (Heat,    3.53e-4) +- (0, 0.4e-4)
  (KdV,     5.14e-4) +- (0, 0.4e-4)
  (RT,      1.69e-2) +- (0, 0.4e-2)
};

\addplot+[fill=cFX] coordinates {
  (Wave,    9.21e-4) +- (0, 10.4e-4)
  (AC,      3.30e-5) +- (0, 1.0e-5)
  (KS,      5.63e-2) +- (0, 3.5e-2)
  (Poisson, 3.02e-5) +- (0, 0.7e-5)
  (Burgers, 1.04e-5) +- (0, 0.3e-5)
  (GL,      4.58e-2) +- (0, 0.5e-2)
  (Heat,    3.65e-4) +- (0, 0.4e-4)
  (KdV,     4.40e-4) +- (0, 0.3e-4)
  (RT,      1.87e-2) +- (0, 0.3e-2)
};

\addplot+[fill=cMT] coordinates {
  (Wave,    9.09e-4) +- (0, 2.2e-4)
  (AC,      3.61e-5) +- (0, 0.8e-5)
  (KS,      2.60e-1) +- (0, 1.2e-1)
  (Poisson, 3.17e-5) +- (0, 0.7e-5)
  (Burgers, 1.00e-5) +- (0, 0.2e-5)
  (GL,      4.55e-2) +- (0, 0.7e-2)
  (Heat,    3.60e-4) +- (0, 0.4e-4)
  (KdV,     6.32e0 ) +- (0, 3.2e0 )
  (RT,      1.73e-2) +- (0, 0.3e-2)
};

\legend{GradNorm-based, SoftAdapt, ReLoBRaLo, Fixed, GradNorm-MTL}
\end{axis}
\end{tikzpicture}
  \caption{Relative $L^2$ error of the loss-balancing methods on nine
  representative PDE benchmarks (log scale, lower is better). Bars show the
  median over 10 seeds and indicate one standard deviation.}
  \label{fig:loss_balancing_main}
\end{figure}
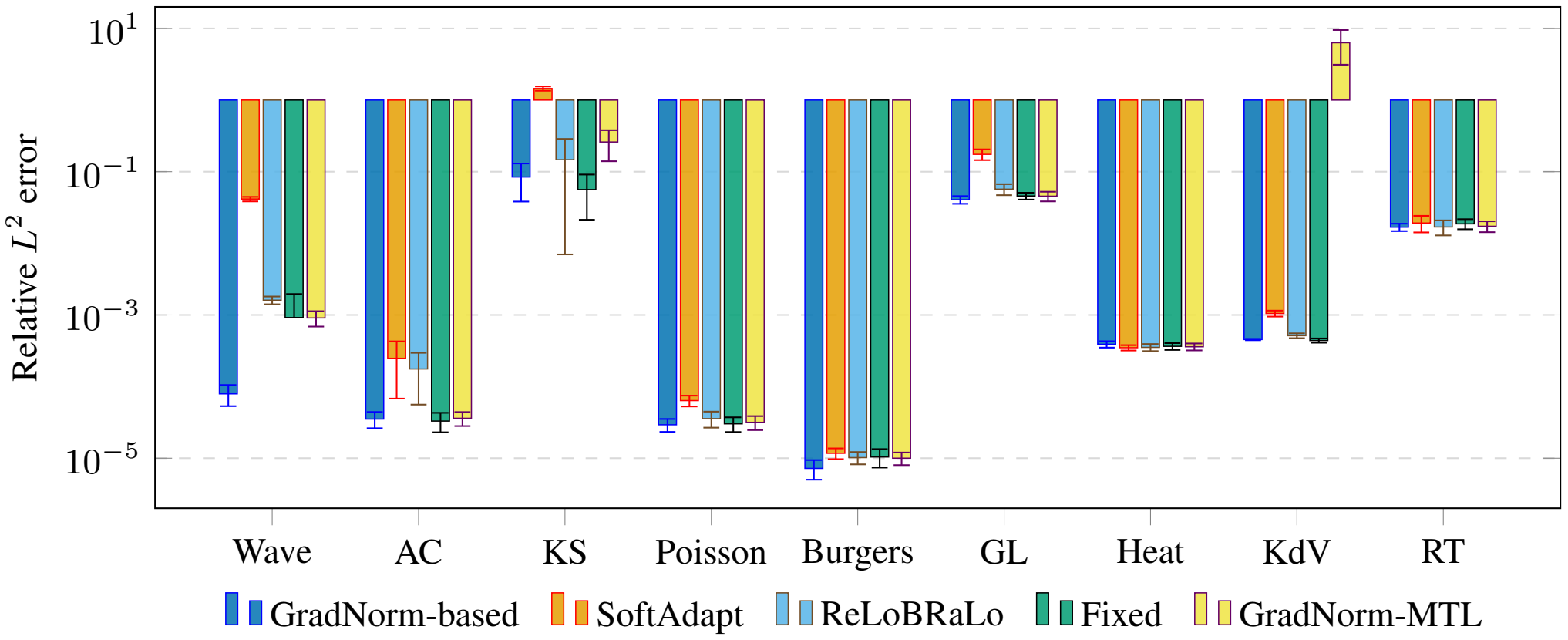
Figure~\ref{fig:loss_balancing_main} shows that no method dominates 
across all nine benchmarks, but two trends are clear. GradNorm-based
balancing performs best overall, achieving the lowest median error on 
five of the nine benchmarks (Wave, Poisson, Burgers, GL, RT) and 
remaining competitive on the others. The fixed-weight baseline is also a
reliable choice and performs best on KS, AC and KdV. Two methods show robustness 
failures: 
SoftAdapt fails to train on Wave and KS, and GradNorm-MTL collapses 
on KdV. Full numerical results, including runtimes, are provided in Appendix~\ref{Appendix:loss_balancing_results}.

\subsection{PINNpipe: a Modular PINN Evaluation Package}
\label{sec:library}
To support reproducibility and lower the barrier for future studies, we release the \emph{PINNpipe} package (\url{https://github.com/KevinvonBargen/pinnpipe}), which contains all reported experiments.
Its internal structure reflects the four-component decomposition of the PINN training pipeline. 
Within the JAX-based package, each component is implemented as a self-contained module with a uniform interface, so that new methods can be added in isolation and benchmarked against established baselines without modifying the surrounding pipeline. 
This positions PINNpipe as a foundation for developing new PINN methods and evaluating them under controlled and standardized conditions.

\section{Conclusions and Future Work}
\label{sec:discussion}

In this work, we introduced an evaluation protocol that (1) treats the PINN training pipeline as a coupled system of four components; (2) requires a competitive implementation of those components not currently under investigation; (3) specifies the metrics and reporting format; and (4) uses the EvalCard documentation template to record the configuration of each evaluation.
We released PINNpipe, a JAX package that implements the protocol together with 18 methods and nine benchmark problems. 
Four case studies illustrated the protocol on the four components of the PINN training pipeline, and they revise the claim of \citet{SOAPWang} that quasi-Newton methods are impractical for complex PDEs.
Coupled with the moderate-size MLPs and the float64 precision on which quasi-Newton methods operate, and embedded in a competitive pipeline, a two-stage optimization scheme outperforms the configuration recommended by the authors on the most demanding benchmarks of that work.
By using the same neural network instance across these studies, we further showed that the limiting factor was not the expressivity of the neural surrogate but the surrounding training pipeline.

Our benchmark suite covers a broad range of real-world, chaotic, and multi-scale problems, but it does not include problems on complex geometries. 
Other challenging classes of PDEs likewise remain outside its scope.
The suite is chosen to be representative of the problems in the studies we re-examine. 
The evaluation standard itself is not restricted to this suite.
We restricted the present study to PINNs and to forward problems. 
We note that inverse formulations and neural-operator methods share enough of the pipeline structure that the protocol and the EvalCard format are applicable with adjustments, and we view the corresponding extensions as a natural next step. 
Finally, the protocol depends on the surrounding pipeline being instantiated competitively. 
As the field continues to evolve, the reference instantiation must be updated regularly. 
PINNeval is designed to make such updates straightforward, but no set of rules can replace the community’s ongoing efforts to maintain a current notion of competitive best practice.

\begin{ack}
This research was supported by the Lower Saxony Ministry of Science and Culture (MWK), funded
through the zukunft.niedersachsen program of the Volkswagen Foundation, Germany (ZN4072). The
computations were carried out on the high-performance computing cluster of Osnabrück University,
funded by the Deutsche Forschungsgemeinschaft (DFG, German Research Foundation) project
number 456666331. The authors declare no competing interests.
\end{ack}

\bibliographystyle{abbrvnat}
\bibliography{bib}

\begin{thebibliography}{30}
\providecommand{\natexlab}[1]{#1}
\providecommand{\url}[1]{\texttt{#1}}
\expandafter\ifx\csname urlstyle\endcsname\relax
  \providecommand{\doi}[1]{doi: #1}\else
  \providecommand{\doi}{doi: \begingroup \urlstyle{rm}\Url}\fi

\bibitem[Al-Baali(1998)]{ssbroyden}
M.~Al-Baali.
\newblock Numerical experience with a class of self-scaling quasi-{N}ewton
  algorithms.
\newblock \emph{Journal of Optimization Theory and Applications}, 96\penalty0
  (3):\penalty0 533--553, 1998.
\newblock \doi{10.1023/A:1022608410710}.

\bibitem[Anagnostopoulos et~al.(2024)Anagnostopoulos, Toscano, Stergiopulos,
  and Karniadakis]{RBA-PINN}
S.~J. Anagnostopoulos, J.~D. Toscano, N.~Stergiopulos, and G.~E. Karniadakis.
\newblock Residual-based attention in physics-informed neural networks.
\newblock \emph{Computer Methods in Applied Mechanics and Engineering},
  421:\penalty0 116805, 2024.
\newblock ISSN 00457825.
\newblock \doi{10.1016/j.cma.2024.116805}.

\bibitem[Bischof and Kraus(2025)]{relobralo}
R.~Bischof and M.~A. Kraus.
\newblock Multi-objective loss balancing for physics-informed deep learning.
\newblock \emph{Computer Methods in Applied Mechanics and Engineering},
  439:\penalty0 117914, 2025.
\newblock ISSN 0045-7825.
\newblock \doi{10.1016/j.cma.2025.117914}.
\newblock URL
  \url{https://www.sciencedirect.com/science/article/pii/S0045782525001860}.

\bibitem[Chen et~al.(2018)Chen, Badrinarayanan, Lee, and
  Rabinovich]{Gradnorm_mtl}
Z.~Chen, V.~Badrinarayanan, C.-Y. Lee, and A.~Rabinovich.
\newblock {GradNorm}: Gradient normalization for adaptive loss balancing in
  deep multitask networks.
\newblock In J.~Dy and A.~Krause, editors, \emph{Proceedings of the 35th
  International Conference on Machine Learning, ICML 2018,
  Stockholmsm{\"a}ssan, Stockholm, Sweden, July 10-15, 2018}, volume~80 of
  \emph{Proceedings of Machine Learning Research}, pages 794--803. PMLR, 2018.

\bibitem[Daw et~al.(2023)Daw, Bu, Wang, Perdikaris, and Karpatne]{r3}
A.~Daw, J.~Bu, S.~Wang, P.~Perdikaris, and A.~Karpatne.
\newblock Mitigating propagation failures in physics-informed neural networks
  using retain-resample-release ({R}3) sampling.
\newblock In A.~Krause, E.~Brunskill, K.~Cho, B.~Engelhardt, S.~Sabato, and
  J.~Scarlett, editors, \emph{Proceedings of the 40th International Conference
  on Machine Learning}, volume 202 of \emph{Proceedings of Machine Learning
  Research}, pages 7264--7302. PMLR, 23--29 Jul 2023.
\newblock URL \url{https://proceedings.mlr.press/v202/daw23a.html}.

\bibitem[Dhar et~al.(2025)Dhar, {Sanchez Villegas}, Karamolegkou, Schiavone,
  Yuan, Chen, Li, Frank, {De Grazia}, Swain, Brandl, Hershcovich, Søgaard, and
  Elliott]{dhar2025evalcardsframeworkstandardizedevaluation}
R.~Dhar, D.~{Sanchez Villegas}, A.~Karamolegkou, A.~Schiavone, Y.~Yuan,
  X.~Chen, J.~Li, S.~Frank, L.~{De Grazia}, M.~Swain, S.~Brandl,
  D.~Hershcovich, A.~Søgaard, and D.~Elliott.
\newblock {EvalCards}: A framework for standardized evaluation reporting, 2025.
\newblock URL \url{https://arxiv.org/abs/2511.21695}.

\bibitem[Dong and Ni(2021)]{periodic}
S.~Dong and N.~Ni.
\newblock A method for representing periodic functions and enforcing exactly
  periodic boundary conditions with deep neural networks.
\newblock \emph{Journal of Computational Physics}, 435:\penalty0 110242, 2021.
\newblock ISSN 0021-9991.
\newblock \doi{10.1016/j.jcp.2021.110242}.
\newblock URL
  \url{https://www.sciencedirect.com/science/article/pii/S0021999121001376}.

\bibitem[Driscoll et~al.(2014)Driscoll, Hale, and Trefethen]{chebfun}
T.~A. Driscoll, N.~Hale, and L.~N. Trefethen, editors.
\newblock \emph{Chebfun Guide}.
\newblock Pafnuty Publications, Oxford, 2014.

\bibitem[Hao et~al.(2024)Hao, Liu, Lu, Lu, Su, Su, Wang, Xia, Yao, Zhang, and
  Zhu]{PINNacle}
Z.~Hao, S.~Liu, F.~Lu, L.~Lu, C.~Su, H.~Su, Z.~Wang, Z.~Xia, J.~Yao, Y.~Zhang,
  and J.~Zhu.
\newblock {PINNacle}: A comprehensive benchmark of physics-informed neural
  networks for solving {PDEs}.
\newblock In A.~Globerson, L.~Mackey, D.~Belgrave, A.~Fan, U.~Paquet,
  J.~Tomczak, and C.~Zhang, editors, \emph{Advances in Neural Information
  Processing Systems 37}, pages 76721--76774, San Diego, California, USA, 2024.
  {Neural Information Processing Systems Foundation, Inc. (NeurIPS)}.
\newblock ISBN 979-8-3313-1438-5.
\newblock \doi{10.52202/079017-2442}.

\bibitem[Harris et~al.(2020)Harris, Millman, van~der Walt, Gommers, Virtanen,
  Cournapeau, Wieser, Taylor, Berg, Smith, Kern, Picus, Hoyer, van Kerkwijk,
  Brett, Haldane, del Río, Wiebe, Peterson, Gérard-Marchant, Sheppard, Reddy,
  Weckesser, Abbasi, Gohlke, and Oliphant]{harris2020array}
C.~R. Harris, K.~J. Millman, S.~J. van~der Walt, R.~Gommers, P.~Virtanen,
  D.~Cournapeau, E.~Wieser, J.~Taylor, S.~Berg, N.~J. Smith, R.~Kern, M.~Picus,
  S.~Hoyer, M.~H. van Kerkwijk, M.~Brett, A.~Haldane, J.~F. del Río, M.~Wiebe,
  P.~Peterson, P.~Gérard-Marchant, K.~Sheppard, T.~Reddy, W.~Weckesser,
  H.~Abbasi, C.~Gohlke, and T.~E. Oliphant.
\newblock Array programming with {NumPy}.
\newblock \emph{Nature}, 585\penalty0 (7825):\penalty0 357--362, 2020.
\newblock \doi{10.1038/s41586-020-2649-2}.
\newblock URL \url{https://doi.org/10.1038/s41586-020-2649-2}.

\bibitem[Heydari et~al.(2019)Heydari, Thompson, and Mehmood]{SoftAdapt}
A.~A. Heydari, C.~A. Thompson, and A.~Mehmood.
\newblock {SoftAdapt}: Techniques for adaptive loss weighting of neural
  networks with multi-part loss functions.
\newblock arXiv preprint arXiv:1912.12355, 2019.

\bibitem[Hunter(2007)]{Hunter2007}
J.~D. Hunter.
\newblock Matplotlib: A {2D} graphics environment.
\newblock \emph{Computing in Science \& Engineering}, 9\penalty0 (3):\penalty0
  90--95, 2007.
\newblock \doi{10.1109/MCSE.2007.55}.

\bibitem[Jeong et~al.(2025)Jeong, Seo, and Hwang]{jeong2025mitigating}
M.~Jeong, G.~Seo, and E.~Hwang.
\newblock Mitigating instability in high residual adaptive sampling for {PINNs}
  via {L}angevin dynamics.
\newblock In \emph{The Thirty-ninth Annual Conference on Neural Information
  Processing Systems}, 2025.

\bibitem[Kassam and Trefethen(2005)]{KassamTrefethen2005}
A.-K. Kassam and L.~N. Trefethen.
\newblock Fourth-order time-stepping for stiff {PDEs}.
\newblock \emph{SIAM Journal on Scientific Computing}, 26\penalty0
  (4):\penalty0 1214--1233, 2005.
\newblock \doi{10.1137/S1064827502410633}.

\bibitem[Krishnapriyan et~al.(2021)Krishnapriyan, Gholami, Zhe, Kirby, and
  Mahoney]{krishnapriyan2021characterizing}
A.~S. Krishnapriyan, A.~Gholami, S.~Zhe, R.~M. Kirby, and M.~W. Mahoney.
\newblock Characterizing possible failure modes in physics-informed neural
  networks.
\newblock In \emph{Advances in Neural Information Processing Systems},
  volume~34, pages 26548--26560, 2021.

\bibitem[McGreivy and Hakim(2024)]{McGreivy.2024}
N.~McGreivy and A.~Hakim.
\newblock Weak baselines and reporting biases lead to overoptimism in machine
  learning for fluid-related partial differential equations.
\newblock \emph{Nature Machine Intelligence}, 6\penalty0 (10):\penalty0
  1256--1269, 2024.
\newblock ISSN 2522-5839.
\newblock \doi{10.1038/s42256-024-00897-5}.

\bibitem[Raissi et~al.(2019)Raissi, Perdikaris, and Karniadakis]{Raissi.2019}
M.~Raissi, P.~Perdikaris, and G.~E. Karniadakis.
\newblock Physics-informed neural networks: A deep learning framework for
  solving forward and inverse problems involving nonlinear partial differential
  equations.
\newblock \emph{Journal of Computational Physics}, 378:\penalty0 686--707,
  2019.
\newblock ISSN 00219991.
\newblock \doi{10.1016/j.jcp.2018.10.045}.

\bibitem[Takamoto et~al.(2022)Takamoto, Praditia, Leiteritz, MacKinlay,
  Alesiani, Pfl\"{u}ger, and Niepert]{Pdebench}
M.~Takamoto, T.~Praditia, R.~Leiteritz, D.~MacKinlay, F.~Alesiani,
  D.~Pfl\"{u}ger, and M.~Niepert.
\newblock {PDEBench}: An extensive benchmark for scientific machine learning.
\newblock In S.~Koyejo, S.~Mohamed, A.~Agarwal, D.~Belgrave, K.~Cho, and A.~Oh,
  editors, \emph{Advances in Neural Information Processing Systems}, volume~35,
  pages 1596--1611. Curran Associates, Inc., 2022.
\newblock URL
  \url{https://proceedings.neurips.cc/paper_files/paper/2022/file/0a9747136d411fb83f0cf81820d44afb-Paper-Datasets_and_Benchmarks.pdf}.

\bibitem[Tancik et~al.(2020)Tancik, Srinivasan, Mildenhall, Fridovich-Keil,
  Raghavan, Singhal, Ramamoorthi, Barron, and Ng]{Fourier-features}
M.~Tancik, P.~P. Srinivasan, B.~Mildenhall, S.~Fridovich-Keil, N.~Raghavan,
  U.~Singhal, R.~Ramamoorthi, J.~T. Barron, and R.~Ng.
\newblock Fourier features let networks learn high frequency functions in low
  dimensional domains.
\newblock In \emph{Advances in Neural Information Processing Systems},
  volume~33, pages 7537--7547, 2020.

\bibitem[Toscano et~al.(2026)Toscano, Chen, Oommen, Darbon, and
  Karniadakis]{Toscano.2026}
J.~D. Toscano, D.~T. Chen, V.~Oommen, J.~Darbon, and G.~E. Karniadakis.
\newblock A variational framework for residual-based adaptivity in neural {PDE}
  solvers and operator learning.
\newblock \emph{npj Artificial Intelligence}, 2\penalty0 (1):\penalty0 32,
  2026.
\newblock ISSN 3005-1460.
\newblock \doi{10.1038/s44387-026-00084-4}.

\bibitem[Urb{\'a}n et~al.(2025)Urb{\'a}n, Stefanou, and Pons]{urban2025pinns}
J.~F. Urb{\'a}n, P.~Stefanou, and J.~A. Pons.
\newblock Unveiling the optimization process of physics informed neural
  networks: How accurate and competitive can {PINNs} be?
\newblock \emph{Journal of Computational Physics}, 523:\penalty0 113656, 2025.
\newblock \doi{10.1016/j.jcp.2024.113656}.

\bibitem[Vyas et~al.(2025)Vyas, Morwani, Zhao, Kwun, Shapira, Brandfonbrener,
  Janson, and Kakade]{SOAP}
N.~Vyas, D.~Morwani, R.~Zhao, M.~Kwun, I.~Shapira, D.~Brandfonbrener,
  L.~Janson, and S.~M. Kakade.
\newblock {SOAP}: Improving and stabilizing {S}hampoo using {A}dam.
\newblock In \emph{The Thirteenth International Conference on Learning
  Representations ({ICLR})}, 2025.
\newblock URL \url{https://openreview.net/forum?id=IDxZhXrpNf}.

\bibitem[Wang et~al.(2021)Wang, Teng, and Perdikaris]{ModifiedMLP}
S.~Wang, Y.~Teng, and P.~Perdikaris.
\newblock Understanding and mitigating gradient flow pathologies in
  physics-informed neural networks.
\newblock \emph{SIAM Journal on Scientific Computing}, 43\penalty0
  (5):\penalty0 A3055--A3081, 2021.
\newblock \doi{10.1137/20M1318043}.

\bibitem[Wang et~al.(2023)Wang, Sankaran, Wang, and Perdikaris]{wang2023expert}
S.~Wang, S.~Sankaran, H.~Wang, and P.~Perdikaris.
\newblock An expert's guide to training physics-informed neural networks.
\newblock \emph{arXiv preprint arXiv:2308.08468}, 2023.
\newblock \doi{10.48550/arXiv.2308.08468}.

\bibitem[Wang et~al.(2024)Wang, Li, Chen, and Perdikaris]{PirateNet}
S.~Wang, B.~Li, Y.~Chen, and P.~Perdikaris.
\newblock {PirateNets}: Physics-informed deep learning with residual adaptive
  networks.
\newblock \emph{Journal of Machine Learning Research}, 25\penalty0
  (402):\penalty0 1--51, 2024.

\bibitem[Wang et~al.(2025)Wang, Bhartari, Li, and Perdikaris]{SOAPWang}
S.~Wang, A.~K. Bhartari, B.~Li, and P.~Perdikaris.
\newblock Gradient alignment in physics-informed neural networks: A
  second-order optimization perspective.
\newblock In \emph{Advances in Neural Information Processing Systems}, 2025.

\bibitem[Wei et~al.(2024)Wei, Liu, Cen, Zhou, Chen, and Zou]{Pdenneval}
P.~Wei, M.~Liu, J.~Cen, Z.~Zhou, L.~Chen, and Q.~Zou.
\newblock {PDENNEval}: A comprehensive evaluation of neural network methods for
  solving {PDEs}.
\newblock In K.~Larson, editor, \emph{Proceedings of the Thirty-Third
  International Joint Conference on Artificial Intelligence, {IJCAI-24}}, pages
  5181--5189. International Joint Conferences on Artificial Intelligence
  Organization, Aug. 2024.
\newblock \doi{10.24963/ijcai.2024/573}.
\newblock URL \url{https://doi.org/10.24963/ijcai.2024/573}.
\newblock Main Track.

\bibitem[Wight and Zhao(2020)]{Timemarching}
C.~L. Wight and J.~Zhao.
\newblock Solving {A}llen--{C}ahn and {C}ahn--{H}illiard equations using the
  adaptive physics informed neural networks.
\newblock arXiv preprint arXiv:2007.04542, 2020.

\bibitem[Wu et~al.(2023)Wu, Zhu, Tan, Kartha, and Lu]{RAR}
C.~Wu, M.~Zhu, Q.~Tan, Y.~Kartha, and L.~Lu.
\newblock A comprehensive study of non-adaptive and residual-based adaptive
  sampling for physics-informed neural networks.
\newblock \emph{Computer Methods in Applied Mechanics and Engineering},
  403:\penalty0 115671, 2023.
\newblock ISSN 0045-7825.
\newblock \doi{10.1016/j.cma.2022.115671}.
\newblock URL
  \url{https://www.sciencedirect.com/science/article/pii/S0045782522006260}.

\bibitem[Xu et~al.(2025)Xu, Liu, Nassereldine, and Xiong]{xu2025fp64}
C.~Xu, D.~Liu, A.~Nassereldine, and J.~Xiong.
\newblock {FP64} is all you need: Rethinking failure modes in physics-informed
  neural networks.
\newblock In \emph{Advances in Neural Information Processing Systems}, 2025.

\end{thebibliography}

\appendix

\newpage
\section*{Appendix Overview}

The Appendix is organized as follows:
\begin{itemize}
  \item 
  \textbf{Appendix~\ref{Appendix:dia}} describes domain-importance adaptation strategies,
  including definitions and pseudocode.
  \item
  \textbf{Appendix \ref{Appendix:lossbalancing}} details the loss balancing methods considered in
  this work.
  \item 
  \textbf{Appendix~\ref{Appendix:neural_networks}} summarizes the neural network architectures and
  input representations used throughout the experiments.
  \item
  \textbf{Appendix~\ref{Appendix:optimizer}} provides definitions and hyperparameters of the optimizers and method alongside them used in this work.
  \item
  \textbf{Appendix~\ref{Appendix:Evalcard}} provides a template for the evaluation card proposed in Section~\ref{sec:Evalcard} and filled-in cards appear in multiple places throughout this work.
  \item 
  \textbf{Appendix~\ref{Appendix:benchmarks}} provides the definitions of all benchmark PDEs,
  including governing equations and reference solutions and their accuracies.
  \item 
  \textbf{Appendix~\ref{Appendix:technicals-pdes}} reports detailed training hyperparameter settings and optimization schedules.
  \item 
  \textbf{Appendix~\ref{Appendix:study_results}} contains extended results for the large-scale method
  comparison study.
  \item
  \textbf{Appendix~\ref{Appendix:ablation_nn_optim}} separates the optimizer, precision and architecture axes of the optimizer and network study in a single-axis ablation.
  \item 
  \textbf{Appendix~\ref{Appendix:vram_results}} presents a GPU memory measurement methodology
  and peak usage results.
  \item 
  \textbf{Appendix~\ref{Appendix:PINNpipe}} describes the JAX library used in this work.
\end{itemize}

\section{Domain-Importance Adaptation}
\label{Appendix:dia}
In this section we introduce the most common methods to adapt the importance inside the training domains of PINNs.
For each equation $e \in \mathcal{E}$, a set of collocation points $C_e$ is selected from a sampling domain $D_e$.
Along with $C_e$, adaptive weights $\omega(t,x)$ are defined.
The loss term associated with equation $e$ is then evaluated on $C_e$ via
\[
\Loss_e(\omega, \theta) 
=\frac{1}{|C_e|}
    \sum_{(t,x)\in C_e}
    \omega(t,x)\,
      r_e(t,x;\theta)^2.
\]
Since the choice of the collocation set $C_e \subset D_e$ is entirely flexible as well as the weight we assign to each point, numerous sampling strategies have been developed to improve the accuracy and efficiency of PINN training.
A central idea behind many of these methods is to increase the density of points in regions where the current residual is large.
This leads to two broad categories of adaptive strategies:

\begin{enumerate}
    \item \textbf{Adaptive sampling:} methods that modify the collocation sets $C_e$ by adding or resampling points from the domains $D_e$, typically biased toward high-residual regions.
    \item \textbf{Importance weighting:} methods that attach dynamic, trainable, or scheduled weights to individual points or segments of the domain.
\end{enumerate}

The representative methods discussed in this section are:
\begin{enumerate}
  \item 
  \textbf{Residual-based Adaptive Refinement (RAR)}:
        Periodically augments $C_e$ by adding the highest-residual points drawn from $D_e$.
  \item 
  \textbf{Residual-Based Adaptive Distribution (RAD)}:
        Periodically resamples the entire collocation set $C_e$ from a probability distribution
        proportional to the current residuals.
  \item 
  \textbf{R3}:
        On each iteration a subset of $C_e$ is resampled based on the mean of the absolute $r_e(t,x;\theta)$.
    \item 
  \textbf{Langevin Adaptive Sampling (LAS)}:
        Each point in $C_e$ gets updated based on the gradient w.r.t $(x,t)$ of $r_e(\theta)$, perturbed by Gaussian noise.  
      \item 
  \textbf{Variational Residual-Based Adaptivity (vRBA)}:
        Adaptive weights $\omega(t,x)$ are used and define a resampling rule for $C_e$.
\end{enumerate}

Methods 1--4 change $C_e$ (sampling-based adaptivity). vRBA (method 5) defines pointwise weights that are used either directly in the loss or to resample $C_e$. In other words, sampling adaptivity alters the geometry of the training set, while weight adaptivity reshapes the loss landscape over a set of points.

In addition to these adaptive strategies, we also include a non-adaptive baseline sampling procedure for comparison: Random-R (new samples drawn each iteration from $D_e$).
\subsection{Residual-based Adaptive Refinement}

The Residual-based Adaptive Refinement (RAR) strategy~\citep{RAR} is one of the earliest adaptive sampling
approaches for PINNs.  
For each equation $e \in \mathcal{E}$, the method begins by drawing an initial collocation set
$C_e$ of size $n_e$ uniformly from the corresponding sampling domain $D_e$.

During training, RAR periodically identifies regions with large residuals by drawing a temporary
candidate set $\tilde C_e \subset D_e$ and evaluating the pointwise residuals
$r_e(t,x;\theta)$ on these candidates.  
A fixed number $l_e$ of the highest absolute residual points from $\tilde C_e$ are then added to the
collocation set $C_e$.  
In this way, global coverage from the original uniform sampling is preserved, while progressively
focusing more sampling density in difficult regions of the domain.

RAR is a \emph{sampling-based} adaptive method because the set $C_e$ grows over time as new informative
points are added.

\begin{algorithm}[H]
  \caption{Residual-based Adaptive Refinement (RAR)}
  \begin{algorithmic}[1]
    \REQUIRE Initial sizes $n_e \in \mathbb{N}$, number of added points $l_e \in \mathbb{N}$,
             number of warm-up iterations $N_0 \in \mathbb{N}$, resampling interval $N_{\mathrm{resample}} \in \N$.
      \STATE Sample initial collocation points $C_e \subset D_e$ with $|C_e| = n_e$.
    \STATE Train the neural network for $N_0$ iterations.
    \WHILE{not converged}
    \STATE Sample temporary candidate points $\tilde C_e \subset D_e$.
    \STATE Evaluate residuals $r_e(t,x;\theta)$ for all $(t,x)\in \tilde C_e$.
    \STATE Gather the $l_e$ points $(t,x)\in\tilde C_e$ with the largest absolute residuals in $\hat{C}_e$.
    \STATE Update collocation set:
          \[
          C_e \gets C_e \cup \hat{C}_e.
          \]
      \STATE Train the neural network for $N_{\mathrm{resample}}$ iterations.
    \ENDWHILE
  \end{algorithmic}
\end{algorithm}
\subsection{Residual-Based Adaptive Distribution}
The Residual-Based Adaptive Distribution (RAD) strategy \citep{RAR} is a sampling-based 
adaptive method that, unlike RAR, \emph{resamples the entire collocation set} at regular 
intervals.  
For each equation $e \in \mathcal{E}$, an initial collocation set $C_e$ of size $n_e$ is 
drawn uniformly from the sampling domain $D_e$.

During training, RAD periodically constructs a new collocation set by sampling from a
residual-weighted probability distribution over $D_e$.  
Points with larger current residuals $r_e(t,x;\theta)$ receive higher probability, thereby
biasing the sampling process toward regions where the condition is currently violated the most.
To maintain sufficient exploration of the domain, the distribution is augmented with a
positive constant~$c$.

Formally, RAD draws new points from a density of the form
\[
p_e(t,x)
\;\propto\;
\frac{ r_e(t,x;\theta)^k }
     { \mathbb{E}_{(t,x)\sim D_e}\bigl[\,r_e(t,x;\theta)^k\,\bigr] }
\;+\; c,
\qquad c,k \ge 0.
\]
The exponent $k$ controls the degree of exploitation
(stronger focus on high-residual regions),
while the constant $c$ ensures exploration by preventing the probability density
from vanishing outside these regions.

\begin{algorithm}[H]
  \caption{Residual-Based Adaptive Distribution (RAD)}
  \begin{algorithmic}[1]
    \REQUIRE Initial collocation size $n_e \in \mathbb{N}$,
             warm-up iterations $N_0 \in \mathbb{N}$,
             exploration coefficient $c \ge 0$,
             exploitation exponent $k \ge 0$,
             resampling interval $N_{\mathrm{resample}} \in \N$
    \STATE Sample initial collocation set $C_e \subset D_e$ with $|C_e| = n_e$.
    \STATE Train the neural network for $N_0$ iterations.
    \WHILE{not converged}
    \STATE Construct the residual-weighted density
          \[
          p_e(t,x)
          \;\propto\;
          \frac{ r_e(t,x;\theta)^k }
               { \mathbb{E}_{D_e}[\,r_e^k\,] }
          + c.
          \]
    \STATE Draw a new collocation set
          \[
          C_e \ \sim p_e(t,x)
          \quad \text{with} \quad |C_e| = n_e.
          \]
      \STATE Train the neural network for $N_{\mathrm{resample}}$ iterations.
    \ENDWHILE
  \end{algorithmic}
\end{algorithm}

\subsection{LAS}
\label{Appendix:las}

The Langevin Adaptive Sampling (LAS) strategy \citep{jeong2025mitigating} is a sampling-based
adaptive method that, unlike RAR, RAD, and R3, does not draw new collocation
points from $D_e$ at each step but instead \emph{evolves} the existing points
by applying a discretized overdamped Langevin diffusion to the squared
residual.
For each equation $e \in \mathcal{E}$, an initial collocation set $C_e$ of size
$n_e$ is drawn uniformly from the sampling domain $D_e$.

At each training iteration, every point $(t,x) \in C_e$ is updated by performing
$l_L \in \N$ Langevin steps. A single Langevin step takes the form
\[
(t,x)
\;\gets\;
(t,x)
+ \frac{\tau}{2}\,
  \nabla_{(t,x)}\bigl|r_e(t,x;\theta)\bigr|^2
+ \beta\sqrt{\tau}\,z,
\qquad
z \sim \mathcal{N}(0,I),
\]
where $\tau > 0$ is the Langevin step size, $\beta > 0$ is the sampling
concentration coefficient, and $z$ is a vector of independent standard Gaussian
noise of the same dimension as $(t,x)$.
The drift term
$\tfrac{\tau}{2}\,\nabla_{(t,x)}|r_e|^2$ pulls each point toward local maxima
of the squared residual, while the diffusion term $\beta\sqrt{\tau}\,z$
injects exploration noise.

For fixed network parameters $\theta$, the limit $\tau \to 0$ and $l_L \to
\infty$ formally recovers the continuous Langevin diffusion whose stationary
density on $D_e$ is
\[
p_e^{\infty}(t,x)
\;\propto\;
\exp\!\Bigl(\tfrac{|r_e(t,x;\theta)|^2}{\beta^2}\Bigr).
\]
The coefficient $\beta$ therefore plays the same role as the exponent $k$ in
RAD: small $\beta$ concentrates $C_e$ in regions of large residual, whereas
large $\beta$ approaches uniform sampling on $D_e$.
In contrast to RAD, however, this residual-weighted distribution is realized
\emph{implicitly} through the dynamics, so no normalization of $|r_e|^k$ over
$D_e$ is required.

\begin{algorithm}[H]
  \caption{Langevin Adaptive Sampling (LAS)}
  \begin{algorithmic}[1]
    \REQUIRE Collocation size $n_e \in \mathbb{N}$,
             Langevin step size $\tau > 0$,
             concentration coefficient $\beta > 0$,
             number of Langevin steps per iteration $l_L \in \mathbb{N}$.
    \STATE Sample initial collocation set $C_e \subset D_e$ with $|C_e| = n_e$.
    \WHILE{not converged}
      \FOR{$l = 1,\dots,l_L$}
        \FORALL{$(t,x) \in C_e$}
          \STATE Compute the residual gradient
                 $\nabla_{(t,x)}|r_e(t,x;\theta)|^2$.
          \STATE Sample $z \sim \mathcal{N}(0,I)$.
          \STATE Update the point via
                \[
                (t,x)
                \gets
                (t,x)
                + \frac{\tau}{2}\,
                  \nabla_{(t,x)}|r_e(t,x;\theta)|^2
                + \beta\sqrt{\tau}\,z.
                \]
        \ENDFOR
      \ENDFOR
      \STATE Train the neural network for one iteration on the updated $C_e$.
    \ENDWHILE
  \end{algorithmic}
\end{algorithm}
The values used are $\beta = 0.2$, $l_L = 1$, and $\tau = 2 \cdot 10^{-3}$.

\subsection{R3}
\label{Appendix:r3}
The Retain--Resample--Release (R3) sampling strategy \citep{r3} is a sampling-based
adaptive method that, in contrast to RAR and RAD, does not require an auxiliary candidate set $\tilde C_e \subset D_e$ to identify high-residual regions.
For each equation $e \in \mathcal{E}$, an initial collocation set $C_e$ of size
$n_e$ is drawn uniformly from the sampling domain $D_e$.

At every iteration, R3 evaluates the residual on the current collocation set and
splits it into two subsets using the mean of the absolute residuals as a
threshold:
\[
\tau_e(\theta)
\;=\;
\frac{1}{|C_e|}
\sum_{(t,x)\in C_e}
\bigl|r_e(t,x;\theta)\bigr|.
\]
Points whose absolute residual exceeds $\tau_e$ form the \emph{retained subset}
\[
C_e^{\mathrm{r}}
\;:=\;
\bigl\{
(t,x)\in C_e \;\big|\; |r_e(t,x;\theta)| > \tau_e(\theta)
\bigr\},
\]
while the remaining points $C_e \setminus C_e^{\mathrm{r}}$ are considered
sufficiently minimized and are dropped.
To keep the size of the collocation set constant, the dropped points are replaced
by a freshly drawn uniform sample
\[
C_e^{\mathrm{s}} \subset D_e
\quad\text{with}\quad
|C_e^{\mathrm{s}}| = n_e - |C_e^{\mathrm{r}}|,
\]
and the updated collocation set used in the next training step is
\[
C_e
\;\gets\;
C_e^{\mathrm{r}} \cup C_e^{\mathrm{s}}.
\]

This update rule realizes three complementary effects.
Points lying in regions whose residual remains above $\tau_e$ persist in $C_e$
across iterations and accumulate as long as those regions are not resolved
(\emph{retain}); points that have been driven below the mean residual are
released and replaced by uniform draws (\emph{release}); and because $|r_e|>\tau_e$ is
a strict inequality, the resampled subset $C_e^{\mathrm{s}}$ is always non-empty,
so a background density of uniform points over $D_e$ is maintained at every
iteration (\emph{resample}).
In contrast to other domain-importance adaptation methods considered, R3 does not introduce any additional hyperparameter.

\begin{algorithm}[H]
  \caption{Retain--Resample--Release (R3)}
  \begin{algorithmic}[1]
    \REQUIRE Collocation size $n_e \in \mathbb{N}$.
    \STATE Sample initial collocation set $C_e \subset D_e$ with $|C_e| = n_e$.
    \WHILE{not converged}
      \STATE Evaluate residuals $r_e(t,x;\theta)$ for all $(t,x)\in C_e$.
      \STATE Compute the threshold
            \[
            \tau_e(\theta)
            = \frac{1}{|C_e|}
              \sum_{(t,x)\in C_e}
              |r_e(t,x;\theta)|.
            \]
      \STATE Form the retained subset
            \[
            C_e^{\mathrm{r}}
            \gets
            \bigl\{(t,x)\in C_e \;\big|\;
                   |r_e(t,x;\theta)| > \tau_e(\theta)\bigr\}.
            \]
      \STATE Draw a fresh uniform sample
            \[
            C_e^{\mathrm{s}} \subset D_e
            \quad\text{with}\quad
            |C_e^{\mathrm{s}}| = n_e - |C_e^{\mathrm{r}}|.
            \]
      \STATE Update the collocation set:
            \[
            C_e \gets C_e^{\mathrm{r}} \cup C_e^{\mathrm{s}}.
            \]
      \STATE Train the neural network for one iteration on the updated $C_e$.
    \ENDWHILE
  \end{algorithmic}
\end{algorithm}

\subsection{vRBA}
\label{Appendix:vRBA}

The variational Residual-Based Adaptivity (vRBA) framework \citep{Toscano.2026}
differs from the previously discussed methods in that it does not prescribe
a single update rule but instead provides a \emph{generative recipe}
parameterized by a convex, increasing potential function
$\Phi : \mathbb{R}_+ \to \mathbb{R}_+$.
Different choices of $\Phi$ recover existing schemes as special cases:
the linear potential $\Phi(r) = r$ corresponds to uniform sampling,
the quadratic potential $\Phi(r) = r^2 + 1$ recovers RAD- and RBA-type
weighting, and the exponential potential $\Phi(r) = e^{r}$ produces a
softmax-style attention over the residuals.

For each equation $e \in \mathcal{E}$, an initial collocation set $C_e$ of size $n_e$ is drawn uniformly from the sampling domain $D_e$, and a non-negative pointwise weight $\lambda_{e,(t,x)} \ge 0$ is associated with each $(t,x) \in C_e$.
At every training iteration, vRBA constructs a target weight at each point
that is proportional to $\Phi'$ evaluated at the rescaled residual.
For a temperature parameter $\epsilon_e^{(k)} > 0$, the target weight at
iteration $k$ is
\[
q^{(k+1)}_{e,(t,x)}
\;\propto\;
\Phi'\!\left(\frac{|r_e(t,x;\theta)|}{\epsilon_e^{(k)}}\right),
\qquad (t,x) \in C_e.
\]

To stabilize the weights against stochastic fluctuations and to retain a
baseline of exploration, the target $q^{(k+1)}_{e,(t,x)}$ is combined convexly with the uniform density $p_u$ on $D_e$ through a stabilization
parameter $\phi \in [0,1]$, and smoothed with an exponential moving
average. Let $\gamma_e \in [0,1)$ denote the EMA memory and
$\eta_e^{*} > 0$ the EMA learning rate. The weight update reads
\[
\lambda_{e,(t,x)}
\;\gets\;
\gamma_e\,\lambda_{e,(t,x)}
+ \eta_e^{*}\!\left(
\phi\, q^{(k+1)}_{e,(t,x)}
+ (1-\phi)\, p_u(t,x)
\right),
\qquad (t,x) \in C_e.
\]
By construction this guarantees
$\lambda_{e,(t,x)} \in [0,\, \eta_e^{*}/(1-\gamma_e)]$
throughout training.

The temperature $\epsilon_e^{(k)}$ governs the sharpness of the induced
distribution and is updated at every iteration according to an annealing
schedule that depends on $\Phi$.
For the exponential potential $\Phi(r) = e^{r}$, a logarithmic decay
\[
\epsilon_e^{(k+1)}
= \frac{c\,\max_{(t,x)\in C_e}|r_e(t,x;\theta)|}{\log(2 + k)},
\qquad c > 0,
\]
is used.
For polynomial potentials $\Phi(r) = r^{p}$ with $p > 1$, the optimal
$\epsilon_e^{(k)}$ is available in closed form by enforcing the
normalization condition
$\int_{D_e}\Phi'(|r_e|/\epsilon_e)\,p_u(\mathrm{d}x) = 1$.
For other admissible potentials (e.g.\ $\Phi(r) = \cosh(r)$,
$\Phi(r) = e^{r^2}$, or $\Phi(r) = (1+r)\log(1+r) - r$), the same
normalization constraint is solved at each iteration via a few
Newton--Raphson steps.

The smoothed weights $\lambda_{e,(t,x)}$ can be used in two complementary
ways:

\paragraph{Importance weighting.}
The weights are applied directly inside the equation loss,
\[
\Loss_e(\theta, \lambda_e)
=
\frac{1}{|C_e|}
\sum_{(t,x)\in C_e}
\bigl(
\lambda_{e,(t,x)}\, r_e(t,x;\theta)
\bigr)^{2},
\]
and the collocation set $C_e$ is kept fixed throughout training.
In this mode, vRBA acts as a weight-based adaptive method.

\paragraph{Adaptive resampling.}
The weights are normalized to a probability mass function on $C_e$,
\[
q_{e,(t,x)}
=
\frac{\lambda_{e,(t,x)}}
     {\sum_{(t',x')\in C_e} \lambda_{e,(t',x')}},
\]
and a new collocation set is resampled from $q_e$.
The unweighted loss
\[
\Loss_e(\theta)
=
\frac{1}{|C_e|}\sum_{(t,x)\in C_e} r_e(t,x;\theta)^{2}
\]
is then evaluated on the resampled points.
Through this resampling rule, vRBA defines a sampling-based adaptive
method that can be combined with the candidate-construction strategies
of RAR and RAD.

\begin{algorithm}[H]
  \caption{Variational Residual-Based Adaptivity (vRBA)}
  \begin{algorithmic}[1]
    \REQUIRE Convex increasing potential $\Phi$,
             collocation size $n_e \in \mathbb{N}$,
             EMA memory $\gamma_e \in [0,1)$,
             EMA learning rate $\eta_e^{*} > 0$,
             stabilization parameter $\phi \in [0,1]$,
             mode $\in \{\text{weighting},\,\text{resampling}\}$.
    \STATE Sample initial collocation set $C_e \subset D_e$ with
           $|C_e| = n_e$.
    \STATE Initialize $\lambda_{e,(t,x)} \gets 0$ for all
           $(t,x) \in C_e$.
    \WHILE{not converged}
      \STATE Evaluate residuals $r_e(t,x;\theta)$ for all
             $(t,x)\in C_e$.
      \STATE Update the temperature $\epsilon_e^{(k)}$ from
             $\{r_e(t,x;\theta)\}_{(t,x)\in C_e}$
             according to the annealing rule associated with $\Phi$.
      \STATE Compute the target weights
            \[
            q^{(k+1)}_{e,(t,x)}
            \;\propto\;
            \Phi'\!\left(\frac{|r_e(t,x;\theta)|}{\epsilon_e^{(k)}}\right),
            \qquad (t,x) \in C_e.
            \]
      \STATE Update the smoothed weights
            \[
            \lambda_{e,(t,x)}
            \gets
            \gamma_e\,\lambda_{e,(t,x)}
            + \eta_e^{*}\!\bigl(
                \phi\, q^{(k+1)}_{e,(t,x)}
                + (1-\phi)\, p_u(t,x)
              \bigr).
            \]
      \IF{mode $=$ weighting}
        \STATE Compute the weighted loss
              $\Loss_e(\theta,\lambda_e)
                = \tfrac{1}{|C_e|}\sum_{(t,x)\in C_e}
                  (\lambda_{e,(t,x)}\, r_e(t,x;\theta))^{2}$.
      \ELSE
        \STATE Form
              $q_{e,(t,x)}
                = \lambda_{e,(t,x)} \big/
                  \sum_{(t',x')\in C_e}\lambda_{e,(t',x')}$.
        \STATE Resample $C_e \sim q_e$ with $|C_e| = n_e$.
        \STATE Compute the unweighted loss
              $\Loss_e(\theta)
                = \tfrac{1}{|C_e|}\sum_{(t,x)\in C_e}
                  r_e(t,x;\theta)^{2}$.
      \ENDIF
      \STATE Train the neural network for one iteration on the current
             loss.
    \ENDWHILE
  \end{algorithmic}
\end{algorithm}
The values used are $\gamma_e = 0.999$,
$\eta_e^{*} = 10^{-2}$, $\phi = 1$, and $c = 1$ in the logarithmic
annealing schedule used for the exponential potential.

\clearpage

\section{Loss Balancing Strategies}
\label{Appendix:lossbalancing}
Training PINNs requires the simultaneous optimization of numerous heterogeneous
loss components arising from the governing equations, boundary and initial
conditions, and possibly additional observational data. These components
typically operate on different physical scales and may exhibit widely varying
optimization dynamics. As a result, certain losses may dominate the gradient
signal, while others receive insufficient attention, ultimately leading to poor
convergence, imbalance between constraints, or even complete failure of
training. This motivates the use of adaptive loss balancing strategies.

A key observation is that each physical constraint in a PINN generally consists of
\emph{multiple distinct equations or residual functions}. 
For example, a multi-dimensional PDE system or a vector-valued boundary condition may produce
several independent residual terms, each requiring accurate enforcement. 
Instead of treating entire categories such as ``PDE'', ``BC'', or ``IC'' as single loss
terms, we adopt an equation-wise perspective: let
\[
\mathcal{E} = \{1,\dots,m\}
\]
denote the set of all individual equations or residual components in the model.
Each $e\in\mathcal{E}$ is associated with its own collocation set $C_e$ and
loss value
\[
\Loss_e(\theta) = \frac{1}{|C_e|}
\sum_{(t,x)\in C_e} r_e(t,x;\theta)^2,
\]
where $r_e$ denotes the corresponding residual function. The total weighted loss
used for training is then
\[
\Loss(\theta;\lambda)
  = \frac{1}{m}\sum_{e\in\mathcal{E}} \lambda_e\, \Loss_e(\theta),
\]
so that every equation receives its own adaptive weight $\lambda_e$. This
equation-wise viewpoint naturally aligns PINNs with the language of multitask
learning, where each equation constitutes a separate task with its own scale and
convergence behavior.

The literature on adaptive loss balancing, however, is not uniform in its
terminology. In particular, the name \emph{GradNorm} has been used to describe
two different algorithms. To avoid ambiguity, we adopt the following naming
convention throughout this paper:
\begin{itemize}
    \item \textbf{GradNorm (Multitask).}  
    Introduced by \citet{Gradnorm_mtl}, this method balances task losses by
    equalizing gradient magnitudes relative to their training rates. It was not
    originally designed for PINNs but is applicable in equation-wise form.

    \item \textbf{Learning Rate Annealing (Original).}  
    Proposed by \citet{ModifiedMLP} specifically for PINNs, this method adjusts
    the weights of boundary and initial condition equations using gradient-norm
    ratios relative to the PDE equations.

    \item \textbf{GradNorm-based}/ Learning Rate Annealing (Revised). 
    A simplified variant used in the \texttt{jaxpi} framework. Although
    internally referred to as ``gradnorm'', it is conceptually distinct from
    GradNorm (Multitask). We refer to this method as \emph{GradNorm-based}.
\end{itemize}

In the remainder of this section we present all loss balancing strategies using
the unified equation-wise notation above, ensuring compatibility with
multi-dimensional, multiscale, and multi-equation PDE problems, before applying
them in our benchmark study.

\subsection{GradNorm (Multitask)}

GradNorm \citep{Gradnorm_mtl} adaptively balances task losses based on their
gradient magnitudes and relative training progress. Interpreting each equation
$e\in\mathcal{E}$ as a separate task, the method updates the weights
$\lambda_e$ so that harder or slower-learning equations receive larger
weights.

For each equation $e$, define the gradient norm and averages
\[
G_e = \bigl\|\nabla_\theta \bigl(\lambda_e \Loss_e(\theta)\bigr)\bigr\|_2,
\qquad
\overline{G} = \frac{1}{m} \sum_{j\in\mathcal{E}} G_j.
\]
To measure training progress, let
\[
\rho_e = \frac{\Loss_e(\theta)}{\Loss_e(\theta^{(0)})},
\qquad
\overline{\rho} = \frac{1}{m} \sum_{j\in\mathcal{E}} \rho_j,
\qquad
\tilde{\rho}_e = \frac{\rho_e}{\overline{\rho}}.
\]
Given a hyperparameter $\alpha \ge 0$, the target gradient is
\[
\widehat{G}_e = \overline{G}\,\tilde{\rho}_e^{\alpha}.
\]
GradNorm updates $\lambda$ by minimizing the auxiliary objective
\[
\Loss_{\mathrm{grad}}(\lambda)
  = \sum_{e\in\mathcal{E}}
      \bigl|G_e - \widehat{G}_e\bigr|.
\]

\begin{algorithm}[H]
    \caption{GradNorm (Multitask, equation-wise)}
    \begin{algorithmic}[1]
    \REQUIRE Initial parameters $\theta^{(0)}$, weights $\lambda_e^{(0)} = 1$, hyperparameter $\alpha\ge0$
    \WHILE{not converged}
      \STATE Compute losses $\Loss_e(\theta)$ for all $e\in\mathcal{E}$
      \STATE Compute gradient norms $G_e = \|\nabla_\theta(\lambda_e \Loss_e)\|_2$
      \STATE Compute $\overline{G}$ and the ratios $\rho_e$, $\overline{\rho}$, $\tilde{\rho}_e$
      \STATE Compute targets $\widehat{G}_e = \overline{G}\,\tilde{\rho}_e^{\alpha}$
      \STATE Form $\displaystyle \Loss_{\mathrm{grad}} = \sum_e |G_e - \widehat{G}_e|$
      \STATE Update $\lambda = \{\lambda_e\}$ by a gradient step on $\Loss_{\mathrm{grad}}$
      \STATE Optionally normalize $\lambda$ (e.g.\ $\frac{1}{m}\sum_e \lambda_e = 1$)
      \STATE Update $\theta$ using the weighted loss
        $\displaystyle \Loss(\theta;\lambda) = \frac{1}{m}\sum_e \lambda_e \Loss_e(\theta)$
    \ENDWHILE
    \end{algorithmic}
\end{algorithm}

\subsection{Learning Rate Annealing (Original)}

Learning Rate Annealing was introduced by \citet{ModifiedMLP} as a
simple heuristic to rebalance the PDE residual loss and the data-fitting losses
(initial and boundary conditions, and possibly additional data). The total loss
is written as
\[
\Loss(\theta)
  = \Loss_r(\theta)
    + \sum_{i=1}^m \lambda_i \, \Loss_i(\theta),
\]
where $\Loss_r$ denotes the (single) PDE residual loss and
$\{\Loss_i\}_{i=1}^m$ are data-fitting losses. The weight of the PDE loss is
fixed to~1, while the weights $\{\lambda_i\}$ of the data-fitting terms are
adapted during training.

At each iteration, the method compares the gradient of $\Loss_r$ with those of
the data-fitting losses. Denoting by $\theta_j$ the $j$-th network parameter,
\citet{ModifiedMLP} define
\[
\gamma_r = \max_j \bigl|\partial_{\theta_j} \Loss_r(\theta)\bigr|,
\qquad
\gamma_i = \frac{1}{|\theta|}
            \sum_j \bigl|\partial_{\theta_j} \Loss_i(\theta)\bigr|,
\]
i.e.\ $\gamma_r$ is the maximum absolute gradient component of the PDE residual
loss, and $\gamma_i$ is the mean absolute gradient component of $\Loss_i$. The
instantaneous target weight for loss $\Loss_i$ is
\[
\widehat{\lambda}_i = \frac{\gamma_r}{\gamma_i},
\]
and the actual weight is updated via an exponential moving average
\[
\lambda_i \leftarrow (1-\alpha)\,\lambda_i + \alpha\,\widehat{\lambda}_i,
\]
with a smoothing parameter $\alpha \in (0,1)$.

\emph{Remark.}
In this original formulation, all PDE residuals are aggregated into a single
scalar loss~$\Loss_r$, and only the data-fitting losses $\{\Loss_i\}$ receive
adaptive weights. As a consequence, Learning Rate Annealing (Original) does not
distinguish between different PDE equations or scales. This limitation
motivated the more general revised variant discussed in the following
subsection.

\subsection{Learning Rate Annealing (Revised)/ GradNorm-based}
\label{Appendix:gradnorm}
A revised version of Learning Rate Annealing is implemented in the
\texttt{jaxpi} framework, where it is internally referred to as
\texttt{gradnorm}. In contrast to the original group-wise formulation, this
variant assigns a separate weight to each equation $e\in\mathcal{E}$ based on
its gradient magnitude, and thus naturally fits our equation-wise setting.

At a given iteration, we define for every equation $e\in\mathcal{E}$ the
gradient magnitude
\[
\gamma_e = \bigl\|\nabla_\theta \Loss_e(\theta)\bigr\|_2,
\]
and the average gradient magnitude
\[
\Gamma = \frac{1}{|\mathcal{E}|}\sum_{j\in\mathcal{E}} \gamma_j.
\]
For $\epsilon>0$, the instantaneous target weight for equation $e$ is
\[
\widehat{\lambda}_e = \frac{\Gamma}{\gamma_e + \epsilon \,\Gamma},
\]
so that equations with comparatively small gradients receive larger target
weights. The actual weights are updated via an exponential moving average
\[
\lambda_e \leftarrow \alpha\,\lambda_e + (1-\alpha)\,\widehat{\lambda}_e,
\qquad \alpha \in (0,1),
\]
and the parameters $\theta$ are updated using the weighted loss
\[
\Loss(\theta;\lambda)
  = \frac{1}{m}\sum_{e\in\mathcal{E}} \lambda_e \Loss_e(\theta).
\]
Typical choices for the hyperparameters are $\epsilon=10^{-5}$ and $\alpha=0.9$.
\begin{algorithm}[H]
  \caption{GradNorm-based (Revised, equation-wise)}
  \begin{algorithmic}[1]
    \REQUIRE Initial parameters $\theta^{(0)}$,
             initial weights $\lambda_e^{(0)} = 1$ for all $e\in\mathcal{E}$,
             $\epsilon>0$,
             smoothing parameter $\alpha \in (0,1)$
    \WHILE{not converged}
      \STATE Compute losses $\Loss_e(\theta)$ for all $e\in\mathcal{E}$
      \STATE Compute gradient magnitudes
             $\gamma_e = \|\nabla_\theta \Loss_e(\theta)\|_2$ for all $e$
      \STATE Compute average gradient magnitude
             $\Gamma = \frac{1}{|\mathcal{E}|}\sum_{j\in\mathcal{E}} \gamma_j$
      \STATE Compute target weights $\widehat{\lambda}_e = \Gamma / (\gamma_e + \epsilon \, \Gamma)$
      \STATE Compute weights 
      $\lambda_e = \alpha\,\lambda_e + (1-\alpha)\,\widehat{\lambda}_e $ 
      \STATE Update $\theta$ using the weighted loss
        $\displaystyle \Loss(\theta;\lambda) = \frac{1}{m}\sum_e \lambda_e \Loss_e(\theta)$
    \ENDWHILE
  \end{algorithmic}
\end{algorithm}

\emph{Remark.}
Unlike the original Learning Rate Annealing, this revised variant operates at
the level of individual equations and treats PDE, BC, IC, and data equations in
a unified way, without collapsing PDE residuals into a single loss.

\subsection{SoftAdapt}

SoftAdapt \citep{SoftAdapt} balances loss terms based on their recent change
during training. At each iteration, losses that have decreased less (or even
increased) compared to the previous iteration receive larger weights via a
softmax over their temporal differences.

Let $\Loss_e^{(n)} = \Loss_e(\theta^{(n)})$ denote the loss of equation
$e\in\mathcal{E}$ at training iteration $n$, and let $\Loss_e^{(n-1)}$ be its
value at the previous iteration. For a temperature parameter $\mathcal{T} > 0$,
we define the score
\[
s_e^{(n)} = \Loss_e^{(n)} - \Loss_e^{(n-1)},
\]
and compute the weights via a softmax
\[
\lambda_e^{(n)}
  = \frac{\exp\bigl(\mathcal{T}\, s_e^{(n)}\bigr)}%
         {\sum_{j\in\mathcal{E}}
            \exp\bigl(\mathcal{T}\, s_j^{(n)}\bigr)},
\qquad e\in\mathcal{E}.
\]
The parameters $\theta$ are then updated using the weighted loss
\[
\Loss(\theta^{(n)};\lambda^{(n)})
  = \frac{1}{m}\sum_{e\in\mathcal{E}} \lambda_e^{(n)} \Loss_e^{(n)}.
\]

\begin{algorithm}[H]
  \caption{SoftAdapt (equation-wise)}
  \begin{algorithmic}[1]
    \REQUIRE Temperature $\mathcal{T}>0$, initial parameters $\theta^{(0)}$, $\Loss_e^{(0)} = 0$ for all $e \in \mathcal{E}$
    \STATE Set iteration counter $n \gets 1$
    \WHILE{not converged}
      \STATE Compute current losses $\Loss_e^{(n)} = \Loss_e(\theta^{(n)})$
      \STATE Compute scores $s_e^{(n)} = \Loss_e^{(n)} - \Loss_e^{(n-1)}$
      \STATE Compute weights
        $\displaystyle \lambda_e^{(n)}
          = \frac{\exp(\mathcal{T}\, s_e^{(n)})}
                 {\sum_{j\in\mathcal{E}}\exp(\mathcal{T}\, s_j^{(n)})}$
      \STATE Update parameters using
        $\displaystyle \Loss(\theta^{(n)};\lambda^{(n)})
          = \frac{1}{m}\sum_{e\in\mathcal{E}}
              \lambda_e^{(n)} \Loss_e^{(n)}$
      \STATE Set $\Loss_e^{(n-1)} \gets \Loss_e^{(n)}$ for all $e$
      \STATE $n \gets n+1$
    \ENDWHILE
  \end{algorithmic}
\end{algorithm}

\subsection{Relative Loss Balancing with Random Lookback (ReLoBRaLo)}

ReLoBRaLo \citep{relobralo} is a SoftAdapt-style method that operates on loss
values rather than gradients and combines two ideas: (i) relative loss
balancing based on ratios of current to past losses, and (ii) a random
look-back mechanism that occasionally refers back to the initial losses. This
is intended to improve robustness and avoid getting stuck in imbalanced
configurations.

Let $\Loss_e^{(n)} = \Loss_e(\theta^{(n)})$ denote the loss of equation
$e\in\mathcal{E}$ at training iteration $n$, and let $m = |\mathcal{E}|$ be
the number of loss components. For two iterations $n_1 > n_2 \ge 0$ and a
temperature parameter $\mathcal{T}>0$, ReLoBRaLo defines the balanced weights

\begin{equation}\label{eq:relobralo_bal}
    \lambda^{\mathrm{bal}}_e(n_1,n_2)
  = m \cdot
    \frac{\exp\!\left(\dfrac{\Loss_e^{(n_1)}}{\mathcal{T}\,\Loss_e^{(n_2)}}\right)}%
         {\sum_{j\in\mathcal{E}}
            \exp\!\left(\dfrac{\Loss_j^{(n_1)}}{\mathcal{T}\,\Loss_j^{(n_2)}}\right)}
\qquad \text{for }e\in\mathcal{E}.
\end{equation}

In practice, ReLoBRaLo uses two specific look-back choices: 
(i) from the initial iteration $0$ and 
(ii) from the previous iteration $n-1$. 
The algorithm updates the weights $\lambda_e^{(n)}$ via:
\[
\lambda^{\mathrm{hist}}_e(n)
  = \rho^{(n)} \lambda_e^{(n-1)}
    + \bigl(1-\rho^{(n)}\bigr)\,\lambda^{\mathrm{bal}}_e(n,0)
\]
\[
\lambda_e^{(n)}
  = \alpha\,\lambda^{\mathrm{hist}}_e(n)
    + (1-\alpha)\,\lambda^{\mathrm{bal}}_e(n,n-1)
\]
where $\alpha\in(0,1)$ is a smoothing parameter and $\rho^{(n)}$ is a
Bernoulli random variable with $\mathbb{E}[\rho^{(n)}]$ chosen close to~1.
The parameters $\theta$ are then updated using the weighted loss
\[
\Loss(\theta^{(n)};\lambda^{(n)})
  = \frac{1}{m}\sum_{e\in\mathcal{E}}
      \lambda_e^{(n)} \Loss_e^{(n)}.
\]

\begin{algorithm}[H]
  \caption{Relative Loss Balancing with Random Lookback (ReLoBRaLo)}
  \begin{algorithmic}[1]
    \REQUIRE Initial parameters $\theta^{(0)}$, temperature $\mathcal{T}>0$,
             smoothing parameter $\alpha\in(0,1)$,
             look-back probability $p\in(0,1)$
    \STATE Initialize $\lambda_e^{(0)} \gets 1$ for all $e\in\mathcal{E}$
    \STATE Compute initial losses $\Loss_e^{(0)} = \Loss_e(\theta^{(0)})$
    \STATE Set iteration counter $n \gets 1$
    \WHILE{not converged}
      \STATE Compute current losses $\Loss_e^{(n)} = \Loss_e(\theta^{(n)})$
      \STATE Compute $\lambda^{\mathrm{bal}}_e(n,0)$ for all $e$ via Eq.~(\ref{eq:relobralo_bal})
      \STATE Sample $\rho^{(n)} \sim \mathrm{Bernoulli}(p)$
      \STATE Compute
        $\lambda^{\mathrm{hist}}_e(n)
          = \rho^{(n)} \lambda_e^{(n-1)}
            + \bigl(1-\rho^{(n)}\bigr)\lambda^{\mathrm{bal}}_e(n,0)$
      \STATE Compute $\lambda^{\mathrm{bal}}_e(n,n-1)$ for all $e$ via Eq.~(\ref{eq:relobralo_bal})
      \STATE Update
        $\lambda_e^{(n)}
          = \alpha\,\lambda^{\mathrm{hist}}_e(n)
            + (1-\alpha)\,\lambda^{\mathrm{bal}}_e(n,n-1)$
      \STATE Update $\theta^{(n+1)}$ using an optimizer step on
        $\displaystyle \Loss(\theta^{(n)};\lambda^{(n)})
          = \frac{1}{m}\sum_{e\in\mathcal{E}} \lambda_e^{(n)} \Loss_e^{(n)}$
      \STATE $n \gets n+1$
    \ENDWHILE
  \end{algorithmic}
\end{algorithm}

\section{Neural Network Architectures}
\label{Appendix:neural_networks}
In this section we introduce the most common and promising architectures of neural networks in PINNs. Apart from the traditionally used MLP, we will also use the Modified MLP introduced by \citet{ModifiedMLP} and the PirateNet \citep{PirateNet} which builds on the idea of Modified MLPs and residual neural networks.\\
In addition to architectural changes, a very promising approach is the inclusion of random Fourier features \citep{Fourier-features}.
As our basic vanilla model we choose the MLP with tanh as its activation function.
It is by far the most used network architecture in PINN studies and publications and is therefore included.

\subsection{Multi-layer Perceptrons}
\label{MLP}
As default function approximator, which maps the spatio-temporal domain to the output space, we use MLPs. 
A layer of an MLP consists of
\begin{enumerate}
    \item affine map $\R^{{d}_{i-1}}\to \R^{d_i}, \quad x \mapsto W_i x+b_i$
    \item $\sigma_i$ is a non-polynomial function from $\R^{d_i} \to \R^{d_i}$
    \item together they form: $f_i (x) = \sigma_i (W_i x + b_i)$.
\end{enumerate}
The functions $\sigma_i$ are commonly called activation functions and are usually element-wise.
The composition of the $L$ layers is given by
\begin{equation}
    u_\theta = f_{MLP} = f_L \circ f_{L-1} \circ ... \circ f_1
\end{equation}
and represents the MLP as one function.
The learnable parameters called $\theta$ are the parameters of the affine maps. Explicitly this means $\theta = \{ (b_i, W_i)\}_{i=1}^L$. 

\subsection{Modified MLP}

Inspired by attention and convolutions, the \emph{Modified MLP} \citep{ModifiedMLP}
enriches a standard MLP by introducing two global feature maps that extract
input-dependent invariants and propagate this information to every hidden layer.
Throughout this section we assume that all hidden layers have the same width
$d_1 = \dots = d_L =: d$.

The global features are defined by
\[
U : \R^{d_0} \to \R^{d}, 
\qquad 
U(x) = \sigma_1\bigl(W_U x + b_U\bigr),
\]
\[
V : \R^{d_0} \to \R^{d}, 
\qquad 
V(x) = \sigma_1\bigl(W_V x + b_V\bigr).
\]

The first hidden layer is given by the standard MLP layer map
\[
H_1(x) = \sigma_1\bigl(W_1 x + b_1\bigr).
\]

For $i = 1,\dots,L-1$ we define the map
\[
Z_i(x) = \sigma_i\bigl(W_i \cdot H_i(x) +b_i\bigr),
\]
and the next hidden layer by
\[
H_{i+1}(x) 
= \bigl(1 - Z_i(x)\bigr) \odot U(x)
  +       Z_i(x)  \odot V(x),
\]
where $\odot$ denotes element-wise multiplication.

The output of the Modified MLP is computed by a final affine map,
\[
u_\theta = f_{\mathrm{Modified MLP}}(x)
= W_L \cdot H_L(x) + b_L.
\]

\subsection{Random Fourier features}
\label{Appendix:RFF}
To enhance the representational capacity of the networks and to mitigate spectral
bias, we employ \emph{random Fourier features} (RFF) as an input embedding.
Since all hidden layers in our architectures have width $d$, we require $d$ to be
an even number.  
We then define the RFF embeddings as a map
\[
\Phi : \R^{d_0} \to \R^{d},
\qquad
\Phi(x) =
\begin{pmatrix}
\cos(Bx) \\
\sin(Bx)
\end{pmatrix},
\]
where the trigonometric functions are applied element-wise.
The matrix $B$ has dimensions
\[
B \in \R^{(d/2)\times d_0},
\]
so that $\Phi(x)$ has dimension $d$.
Each entry of $B$ is sampled independently from a Gaussian distribution
$\mathcal{N}(0,s^2)$, where the parameter $s>0$ controls the frequency spectrum.

The embedded coordinates $\Phi(x)$ replace the raw input $x$ and serve as the 
new input representation for all architectures.
In particular, in the Modified MLP formulation the global feature maps and first
hidden layer become
\[
U(x) = \sigma_1\!\bigl( W_U \cdot \Phi(x) +b_U \bigr), 
\qquad
V(x) = \sigma_1\!\bigl(  W_V \cdot \Phi(x) +b_V \bigr),
\]
\[
H_1(x) = \sigma_1\!\bigl(  W_1 \cdot \Phi(x) +b_1 \bigr).
\]

\subsection{PirateNet}
Let $d_{\mathrm{in}}$ denote the feature dimension of the network input
(e.g., after an optional embedding), and let $d_h$ be the hidden width of the
bottleneck layers.

As in the Modified MLP, PirateNet uses two global feature maps
\[
U, V : \R^{d_{\mathrm{in}}} \to \R^{d_h},
\qquad
U(x) = \sigma\bigl(W_U x + b_U \bigr), 
\quad
V(x) = \sigma\bigl(W_V x + b_V \bigr),
\]
which are computed once from the input $x$ and shared across all blocks.

A PirateNet consists of $L$ blocks, each defining a map
\[
f_k : \R^{d_{\mathrm{in}}} \to \R^{d_{\mathrm{in}}}, \qquad k = 1,\dots,L.
\]
Given $x \in \R^{d_{\mathrm{in}}}$, the $k$-th block applies a three-layer
bottleneck with two gating operations:
\begin{align*}
    H_{1,k} &: \R^{d_{\mathrm{in}}} \to \R^{d_h},
    & H_{1,k}(x) &= \sigma\bigl( W_{1,k} x + b_{1,k} \bigr), \\[4pt]
    G_{1,k} &: \R^{d_{\mathrm{in}}} \to \R^{d_h},
    & G_{1,k}(x) &= H_{1,k}(x) \odot U(x)
                   + \bigl(1 - H_{1,k}(x)\bigr) \odot V(x), \\[8pt]
    H_{2,k} &: \R^{d_{\mathrm{in}}} \to \R^{d_h},
    & H_{2,k}(x) &= \sigma\bigl(W_{2,k} G_{1,k}(x) + b_{2,k}\bigr), \\[4pt]
    G_{2,k} &: \R^{d_{\mathrm{in}}} \to \R^{d_h},
    & G_{2,k}(x) &= H_{2,k}(x) \odot U(x)
                   + \bigl(1 - H_{2,k}(x)\bigr) \odot V(x), \\[8pt]
    H_{3,k} &: \R^{d_{\mathrm{in}}} \to \R^{d_{\mathrm{in}}},
    & H_{3,k}(x) &= \sigma\bigl(W_{3,k} G_{2,k}(x) + b_{3,k}\bigr),
\end{align*}
where the affine maps satisfy
\[
W_{1,k} \in \R^{d_{\mathrm{in}} \times d_h},\qquad
W_{2,k} \in \R^{d_{h} \times d_h},\qquad
W_{3,k} \in \R^{d_h \times d_{\mathrm{in}}},
\]
and $\odot$ denotes element-wise multiplication.

The residual output of the block is
\[
f_k(x)
= \alpha_k\, H_{3,k}(x)
  + (1 - \alpha_k)\, x,
\]
where $\alpha_k \in \R$ is a trainable scalar parameter controlling the balance
between nonlinear transformation and identity propagation.

The entire PirateNet is the composition of its blocks,
\[
u_\theta = f_{\mathrm{PirateNet}}
  = f_L \circ f_{L-1} \circ \cdots \circ f_1.
\]

\subsection{Hard periodic boundary conditions}
\label{HardBC}

In order to impose periodic boundary conditions exactly rather than as a soft
penalty in the loss, we follow the construction of \citet{periodic}. The idea is
to compose the network with a coordinate embedding that is itself exactly
periodic. The resulting network then inherits exact periodicity together with
the periodicity of all its derivatives.

Let the input be split as $x = (x_p, x_{np}) \in \R^{p} \times \R^{d_0 - p}$,
where $x_p = (x_1,\dots,x_p)$ are the directions in which the solution is
required to be periodic with periods $L_1,\dots,L_p > 0$, and $x_{np}$
collects the remaining (non-periodic) coordinates. For each periodic
direction $j = 1,\dots,p$ we set $\omega_j = 2\pi / L_j$ and define the
elementary periodic embedding
\[
\phi_j : \R \to \R^{2},
\qquad
\phi_j(x_j) = \bigl(\cos(\omega_j x_j),\; \sin(\omega_j x_j)\bigr).
\]

The full periodic embedding is then the map
\[
\Pi : \R^{d_0} \to \R^{d_0 + p},
\qquad
\Pi(x) = \bigl(\phi_1(x_1),\,\dots,\,\phi_p(x_p),\, x_{np}\bigr).
\]
By construction, $\Pi$ is smooth and exactly periodic in each periodic
coordinate, i.e.
\[
\Pi(x_1,\dots,x_j + L_j,\dots,x_{d_0}) \;=\; \Pi(x)
\qquad \text{for every } j = 1,\dots,p.
\]

The periodic embedding $\Pi(x)$ replaces the raw input $x$ as the
representation passed to any of the architectures introduced above,
\[
u_\theta(x) \;=\; f\bigl(\Pi(x)\bigr),
\]
where $f$ is an MLP, a Modified MLP, or a PirateNet, now with input dimension
$d_0 + p$ instead of $d_0$. Since $f$ acts on $x$ only through $\Pi(x)$ and
$\Pi$ is exactly $L_j$-periodic in $x_j$, the network output $u_\theta$ is
itself exactly periodic in $x_1,\dots,x_p$, and so are all its derivatives.
The periodic boundary conditions are therefore enforced as a hard constraint,
and no corresponding penalty term is needed in the PINN loss.

When random Fourier features are additionally employed, the periodic
embedding is applied first and the RFF map afterwards, so that the input
representation seen by the network becomes
\[
(\Phi \circ \Pi) : \R^{d_0} \to \R^{d},
\]
with $\Phi$ as defined in Section~\ref{Appendix:RFF}. The matrix $B$
in the RFF map then has dimension $B \in \R^{(d/2) \times (d_0 + p)}$, while
periodicity of $u_\theta$ in $x_1,\dots,x_p$ is preserved because $\Phi$
acts only through the already-periodic $\Pi(x)$.

\newpage

\section{Optimizer}
\label{Appendix:optimizer}
\subsection{Shampoo with Adam in the Preconditioner's Eigenbasis (SOAP)}

SOAP \citep{SOAP} is a quasi-second-order optimizer that runs AdamW in a
slowly changing rotated coordinate basis derived from the eigenvectors of
Shampoo's Kronecker-factored preconditioner. Conceptually, SOAP can be viewed
as Adam applied not to the raw gradient but to a \emph{whitened} gradient,
where the whitening rotation is given by the eigendecomposition of two
layer-wise gradient-covariance matrices accumulated in the style of Shampoo.
The rotation is refreshed only every $T_{\mathrm{precond}}$ steps via a
single step of orthogonal iteration, while the Adam moments are updated at
every step inside the rotated basis. This combines the curvature awareness
of Shampoo with the per-coordinate adaptivity and computational simplicity
of Adam, and adds only a single hyperparameter ($T_{\mathrm{precond}}$)
relative to AdamW.

We apply SOAP layer-wise to the 2D weight matrices
$W_i\in\mathbb{R}^{m_i\times n_i}$, $i=1,\dots,L$. The 1D parameters $b_i$
are updated with standard AdamW. Let
$G_i^{(n)} = \nabla_{W_i}\Loss(\theta^{(n)}) \in \mathbb{R}^{m_i\times n_i}$
denote the gradient of layer $i$ at iteration $n$. For brevity we drop the
layer index $i$ in what follows; the updates are applied independently per
2D layer. For each such layer, SOAP maintains the first moment
$M^{(n)}\in\mathbb{R}^{m\times n}$ in the original basis, the second moment
$V^{(n)}\in\mathbb{R}^{m\times n}$ in the rotated basis, the Kronecker
factors $\mathbf{L}^{(n)}\in\mathbb{R}^{m\times m}$ and
$\mathbf{R}^{(n)}\in\mathbb{R}^{n\times n}$, and the orthogonal eigenvector
estimates $Q_L^{(n)}\in\mathbb{R}^{m\times m}$,
$Q_R^{(n)}\in\mathbb{R}^{n\times n}$.

The two Kronecker factors are exponential moving averages of the left and
right gradient covariances,
\begin{equation}\label{eq:SOAP_LR}
  \mathbf{L}^{(n)} = \beta_2\,\mathbf{L}^{(n-1)} + (1-\beta_2)\,
                     G^{(n)} G^{(n)\top},
  \qquad
  \mathbf{R}^{(n)} = \beta_2\,\mathbf{R}^{(n-1)} + (1-\beta_2)\,
                     G^{(n)\top} G^{(n)},
\end{equation}
and form the layerwise Shampoo preconditioner
$\mathbf{L}^{(n)}\otimes\mathbf{R}^{(n)}$ in the vectorized parameter space.
Rather than inverting this preconditioner, SOAP works in its eigenbasis:
every $T_{\mathrm{precond}}$ iterations the eigenvector estimates are
refreshed via a single step of orthogonal iteration,
\begin{equation}\label{eq:SOAP_Q}
  Q_L^{(n)} = \mathrm{qr}\!\bigl(\mathbf{L}^{(n)} Q_L^{(n-1)}\bigr),
  \qquad
  Q_R^{(n)} = \mathrm{qr}\!\bigl(\mathbf{R}^{(n)} Q_R^{(n-1)}\bigr),
\end{equation}
where $\mathrm{qr}(\cdot)$ returns the orthogonal factor of a QR
decomposition. On iterations that are not multiples of
$T_{\mathrm{precond}}$, $Q_L$ and $Q_R$ are kept fixed. The estimates are
initialized at $n=1$ from a full eigendecomposition of
$\mathbf{L}^{(1)},\mathbf{R}^{(1)}$.

Inside the rotated basis defined by $Q_L^{(n)},Q_R^{(n)}$, SOAP performs an
Adam update. The first moment is accumulated in the original basis and
projected each step, while the second moment is accumulated directly in the
rotated basis,
\begin{align}
  M^{(n)}  &= \beta_1\,M^{(n-1)} + (1-\beta_1)\,G^{(n)},
              \label{eq:SOAP_M}\\[2pt]
  G'^{(n)} &= Q_L^{(n)\top} G^{(n)} Q_R^{(n)},
   \qquad
  M'^{(n)} = Q_L^{(n)\top} M^{(n)} Q_R^{(n)},
              \label{eq:SOAP_proj}\\[2pt]
  V^{(n)}  &= \beta_2\,V^{(n-1)} + (1-\beta_2)\,
              G'^{(n)} \odot G'^{(n)},
              \label{eq:SOAP_V}
\end{align}
where $\odot$ denotes the Hadamard product. With bias-corrected moments
$\widehat{M}'^{(n)} = M'^{(n)} / (1-\beta_1^{n})$ and
$\widehat{V}^{(n)}    = V^{(n)}  / (1-\beta_2^{n})$, the Adam direction in
the rotated basis is
\begin{equation}\label{eq:SOAP_adam}
  N'^{(n)} = \frac{\widehat{M}'^{(n)}}{\sqrt{\widehat{V}^{(n)}} + \epsilon},
\end{equation}
with the square root and division taken element-wise. The update is rotated
back to the original basis and applied,
\begin{equation}\label{eq:SOAP_step}
  W^{(n+1)} = W^{(n)} - \alpha\, Q_L^{(n)} N'^{(n)} Q_R^{(n)\top}.
\end{equation}
We use weight decay $0$ throughout, so the AdamW decoupled-decay term
$-\alpha\lambda W^{(n)}$ in equation~(\ref{eq:SOAP_step}) vanishes and is
omitted. Setting $Q_L^{(n)}=I_m$ and $Q_R^{(n)}=I_n$ in
equations~(\ref{eq:SOAP_M})--(\ref{eq:SOAP_step}) recovers AdamW exactly,
which is the update we use for the bias parameters $b_i$.

\begin{algorithm}[H]
  \caption{SOAP step for a single 2D layer $W\in\mathbb{R}^{m\times n}$.}
  \begin{algorithmic}[1]
    \REQUIRE Initial weights $W^{(0)}$, learning rate $\alpha$,
             EMA decays $\beta_1,\beta_2\in[0,1)$,
             stability constant $\epsilon>0$,
             preconditioning frequency $T_{\mathrm{precond}}\in\mathbb{N}$
    \STATE Initialize $M^{(0)}\gets 0$, $V^{(0)}\gets 0$,
           $\mathbf{L}^{(0)}\gets 0$, $\mathbf{R}^{(0)}\gets 0$
    \STATE Initialize $Q_L^{(0)}\gets I_m$, $Q_R^{(0)}\gets I_n$
    \STATE $n \gets 1$
    \WHILE{not converged}
      \STATE Compute gradient $G^{(n)} = \nabla_W \Loss(\theta^{(n)})$
      \STATE $M^{(n)} \gets \beta_1 M^{(n-1)} + (1-\beta_1)\,G^{(n)}$
        \hfill\COMMENT{first moment, original basis}
      \STATE $\mathbf{L}^{(n)} \gets \beta_2\mathbf{L}^{(n-1)} + (1-\beta_2)\,G^{(n)} G^{(n)\top}$
      \STATE $\mathbf{R}^{(n)} \gets \beta_2\mathbf{R}^{(n-1)} + (1-\beta_2)\,G^{(n)\top} G^{(n)}$
      \IF{$n = 1$}
        \STATE Initialize $Q_L^{(n)}, Q_R^{(n)}$ from full eigendecomposition of $\mathbf{L}^{(n)}, \mathbf{R}^{(n)}$
      \ELSIF{$n \bmod T_{\mathrm{precond}} = 0$}
        \STATE $Q_L^{(n)} \gets \mathrm{qr}(\mathbf{L}^{(n)} Q_L^{(n-1)})$
          \hfill\COMMENT{one step of orthogonal iteration}
        \STATE $Q_R^{(n)} \gets \mathrm{qr}(\mathbf{R}^{(n)} Q_R^{(n-1)})$
      \ELSE
        \STATE $Q_L^{(n)} \gets Q_L^{(n-1)}$,\quad $Q_R^{(n)} \gets Q_R^{(n-1)}$
      \ENDIF
      \STATE $G'^{(n)} \gets Q_L^{(n)\top} G^{(n)} Q_R^{(n)}$
        \hfill\COMMENT{rotate gradient into eigenbasis}
      \STATE $M'^{(n)} \gets Q_L^{(n)\top} M^{(n)} Q_R^{(n)}$
      \STATE $V^{(n)} \gets \beta_2 V^{(n-1)} + (1-\beta_2)\,G'^{(n)}\odot G'^{(n)}$
        \hfill\COMMENT{second moment, rotated basis}
      \STATE $\widehat{M}'^{(n)} \gets M'^{(n)} / (1-\beta_1^{n})$
      \STATE $\widehat{V}^{(n)}    \gets V^{(n)}  / (1-\beta_2^{n})$
      \STATE $N'^{(n)} \gets \widehat{M}'^{(n)} \big/ \bigl(\sqrt{\widehat{V}^{(n)}} + \epsilon\bigr)$
        \hfill\COMMENT{Adam step in rotated basis}
      \STATE $W^{(n+1)} \gets W^{(n)} - \alpha\,Q_L^{(n)} N'^{(n)} Q_R^{(n)\top}$
        \hfill\COMMENT{rotate back and update}
      \STATE $n \gets n+1$
    \ENDWHILE
  \end{algorithmic}
\end{algorithm}

\paragraph{Default settings.}
The defaults recommended in the reference SOAP implementation~\citep{SOAP} are $\beta_1 = \beta_2 = 0.95$, $\epsilon = 10^{-8}$, and $T_{\mathrm{precond}} = 10$. 
The recommended learning rate in the reference implementation is $\alpha = 3\times 10^{-3}$. 
The reference implementation also uses a decoupled weight decay of $10^{-2}$ by default; 
in our experiments we set the weight decay to $0$ because we employ an LR schedule manually, so the corresponding term is omitted from equation~(\ref{eq:SOAP_step}).
The following settings are used throughout this work:
\begin{table}[H]
  \centering
  \label{tab:SOAP_defaults}
  \caption{Default SOAP hyperparameters used in this work.}
  \begin{tabular}{lll}
    \toprule
    Symbol & Description & Default \\
    \midrule
    $\alpha$               & learning rate                       & $3\times 10^{-3}$ \\
    $\beta_1$              & EMA decay for first moment $M$      & $0.95$ \\
    $\beta_2$              & EMA decay for $V$, $\mathbf{L}$, $\mathbf{R}$ & $0.95$ \\
    $\epsilon$             & numerical stability constant        & $10^{-8}$ \\
    $T_{\mathrm{precond}}$ & eigenbasis update period            & $2$ \\
    \bottomrule
  \end{tabular}
\end{table}

\subsection{Self-Scaled Broyden (SSBroyden)}

SSBroyden \citep{ssbroyden} is a full-memory quasi-Newton optimizer that
maintains a dense approximation $H^{(n)}\in\mathbb{R}^{d\times d}$ of the
inverse Hessian $\nabla^2\Loss(\theta^{(n)})^{-1}$, where $d=\dim\theta$ is
the total number of trainable parameters. At each iteration the method
flattens $\theta^{(n)}$ into a vector, computes the search direction
$p^{(n)} = -H^{(n)}\,g^{(n)}$ from the gradient
$g^{(n)} = \nabla\Loss(\theta^{(n)}) \in \mathbb{R}^d$, performs a Wolfe
line search along $p^{(n)}$ to find a step length $\alpha^{(n)} > 0$
(Section~\ref{sec:ssb_ls}), takes the step
$\theta^{(n+1)} = \theta^{(n)} + \alpha^{(n)} p^{(n)}$, and finally refines
$H^{(n)}$ to $H^{(n+1)}$ via a self-scaled rank-2 update. Because $H^{(n)}$
is stored explicitly, SSBroyden has $\mathcal{O}(d^2)$ memory and per-step
cost, which restricts it to the moderate-$d$ regime typical of PINN-style
problems. Recent work by \citet{urban2025pinns} has demonstrated that
SSBroyden, when combined with a strong-Wolfe line search, attains relative
errors on standard PINN benchmarks that are several orders of magnitude
below those obtained with Adam, BFGS or L-BFGS, motivating its use as the
quasi-Newton optimizer in this work.

The classical Broyden family is a one-parameter convex combination of the
BFGS ($\theta=0$) and DFP ($\theta=1$) updates applied directly to the
inverse Hessian. The \emph{self-scaled} extension multiplies the
DFP-component of the update by the reciprocal of a positive scaling factor
$\tau^{(n)}$ before the rank-1 BFGS-style outer product
$\rho^{(n)} s^{(n)} s^{(n)\top}$ is added; this is intended to control the
conditioning of $H^{(n+1)}$ across iterations. The variants
SSBroyden$\{1,2,3\}$ in the literature differ only in how the Broyden
parameter $\theta^{(n)}$ is chosen at each step inside an admissible
interval; below we describe the SSBroyden2 selection rule, which is the
one used in our implementation.

For iteration $n$ define the curvature pair $(s^{(n)}, y^{(n)})$ and the
auxiliary vector $u^{(n)} = H^{(n)} y^{(n)}$,
\begin{equation}\label{eq:ssb_curv}
   s^{(n)} = \alpha^{(n)}\,p^{(n)},\qquad
   y^{(n)} = g^{(n+1)} - g^{(n)},\qquad
   \rho^{(n)} = \frac{1}{y^{(n)\top} s^{(n)}},
\end{equation}
together with the scalars
\begin{equation}\label{eq:ssb_scalars}
   h^{(n)} = \rho^{(n)}\,y^{(n)\top} u^{(n)},
   \qquad
   b^{(n)} = -\alpha^{(n)} \rho^{(n)}\, s^{(n)\top} g^{(n)},
   \qquad
   a^{(n)} = b^{(n)}\,h^{(n)} - 1,
\end{equation}
and the bounds
\begin{equation}\label{eq:ssb_bounds}
   \rho_-^{(n)}    = \min\!\left\{1,\; h^{(n)}\!\left(1 - \sqrt{\tfrac{|a^{(n)}|}{1+|a^{(n)}|}}\right)\right\},\qquad
   \theta_-^{(n)}  = \frac{\rho_-^{(n)} - 1}{a^{(n)}},\qquad
   \theta_+^{(n)}  = \frac{1}{\rho_-^{(n)}}.
\end{equation}
The interval $[\theta_-^{(n)}, \theta_+^{(n)}]$ is the admissible range for
the Broyden parameter; SSBroyden2 selects
\begin{equation}\label{eq:ssb_theta}
   \theta^{(n)} = \max\!\left\{\theta_-^{(n)},\,
                  \min\!\left\{\theta_+^{(n)},\,
                  \frac{1 - b^{(n)}}{b^{(n)}}\right\}\right\}.
\end{equation}
With $\sigma^{(n)} = 1 + a^{(n)}\theta^{(n)}$ and
$\rho_*^{(n)} = \min\{1,\,1/b^{(n)}\}$, and assuming $d>1$, the self-scaling
factor is
\begin{equation}\label{eq:ssb_tau}
  \tau^{(n)} =
  \begin{cases}
    \min\!\Bigl\{\rho_*^{(n)} |\sigma^{(n)}|^{1/(1-d)},\; \sigma^{(n)}\Bigr\}
       & \text{if } \theta^{(n)} \le 0,\\[2pt]
    \rho_*^{(n)}\, \min\!\Bigl\{|\sigma^{(n)}|^{1/(1-d)},\; 1/\theta^{(n)}\Bigr\}
       & \text{if } \theta^{(n)} > 0.
  \end{cases}
\end{equation}

Finally, define the auxiliary vector $v^{(n)}$ and convex coefficient
$\phi^{(n)}$ by
\begin{equation}\label{eq:ssb_vphi}
  v^{(n)} = \rho^{(n)}\, s^{(n)} - \frac{u^{(n)}}{y^{(n)\top} u^{(n)}},
  \qquad
  \phi^{(n)} = \frac{1 - \theta^{(n)}}{1 + a^{(n)}\theta^{(n)}}.
\end{equation}
The inverse Hessian approximation is then updated as
\begin{equation}\label{eq:ssb_update}
  H^{(n+1)} = \frac{1}{\tau^{(n)}}
              \!\left[H^{(n)} - \frac{u^{(n)} u^{(n)\top}}{y^{(n)\top} u^{(n)}}
                              + \phi^{(n)}\,(y^{(n)\top} u^{(n)})\, v^{(n)} v^{(n)\top}\right]
              + \rho^{(n)}\, s^{(n)} s^{(n)\top}.
\end{equation}
With $\theta^{(n)}=0$ and $\tau^{(n)}=1$, equation~\eqref{eq:ssb_update}
reduces to the standard BFGS update on the inverse Hessian; with
$\theta^{(n)}=1$ it reduces to DFP.

\subsubsection{Wolfe line search}\label{sec:ssb_ls}

Each step length $\alpha^{(n)}$ is obtained by a strong-Wolfe line search on
the scalar function
$\varphi(\alpha) = \Loss(\theta^{(n)} + \alpha p^{(n)})$. A trial step
length $\alpha$ is accepted when both
\begin{align}
  \varphi(\alpha) &\le \varphi(0) + c_1\,\alpha\,\varphi'(0)
     &&\text{(Armijo / sufficient decrease)},
     \label{eq:wolfe_armijo}\\
  |\varphi'(\alpha)| &\le c_2\,|\varphi'(0)|
     &&\text{(strong curvature condition)},
     \label{eq:wolfe_curv}
\end{align}
with $0 < c_1 < c_2 < 1$. We use the JAX implementation of the single-step
strong-Wolfe line search of Mor\'e and Thuente as the inner primitive, and
wrap it in a three-stage cascade: each stage uses the same $c_1$ but a
progressively tighter (i.e.\ smaller) curvature tolerance $c_2$, and a
stage is only invoked if the previous one returned a failure flag --- e.g.\
exceeded its iteration cap without finding any $\alpha$ satisfying both
\eqref{eq:wolfe_armijo} and \eqref{eq:wolfe_curv}. The schematic is given
in Algorithm~\ref{alg:ssb_ls}.

\begin{algorithm}[H]
  \caption{Wolfe line-search cascade used inside SSBroyden.}
  \label{alg:ssb_ls}
  \begin{algorithmic}[1]
    \REQUIRE Iterate $\theta^{(n)}$, direction $p^{(n)}$,
             value $\Loss(\theta^{(n)})$, gradient $g^{(n)}$
    \REQUIRE Armijo constant $c_1$;
             curvature tolerances $c_2^{(1)} > c_2^{(2)} > c_2^{(3)}$;
             per-stage iteration caps $K_1, K_2, K_3$
    \STATE \textbf{Stage 1.} Run strong-Wolfe line search with $(c_1, c_2^{(1)})$, max $K_1$ trials.
    \IF{stage 1 succeeded}
      \STATE \textbf{return} $(\alpha^{(n)},\,\Loss(\theta^{(n+1)}),\,g^{(n+1)})$
    \ENDIF
    \STATE \textbf{Stage 2.} Run strong-Wolfe line search with $(c_1, c_2^{(2)})$, max $K_2$ trials.
    \IF{stage 2 succeeded}
      \STATE \textbf{return} $(\alpha^{(n)},\,\Loss(\theta^{(n+1)}),\,g^{(n+1)})$
    \ENDIF
    \STATE \textbf{Stage 3.} Run strong-Wolfe line search with $(c_1, c_2^{(3)})$, max $K_3$ trials.
    \STATE \textbf{return} stage-3 result (success or failure flag).
  \end{algorithmic}
\end{algorithm}

In our implementation we set $c_1 = 10^{-4}$ for all three stages, with
curvature tolerances $c_2^{(1)} = 0.9$, $c_2^{(2)} = 0.8$, $c_2^{(3)} = 0.5$
and per-stage cap $K_1 = K_2 = K_3 = 25$. 

\begin{algorithm}[H]
  \caption{Self-Scaled Broyden (SSBroyden2) step on the flattened parameter vector $\theta\in\mathbb{R}^d$.}
  \label{alg:ssbroyden}
  \begin{algorithmic}[1]
    \REQUIRE Initial parameters $\theta^{(0)}$, initial inverse Hessian $H^{(0)} = I_d$
    \REQUIRE Wolfe parameters $c_1, c_2^{(1)}, c_2^{(2)}, c_2^{(3)}, K_1, K_2, K_3$
    \REQUIRE Curvature-collapse threshold $\varepsilon>0$, fallback constant $\rho_{\mathrm{fb}}$
    \STATE $n \gets 0$
    \WHILE{not converged}
      \STATE Compute $\Loss(\theta^{(n)})$ and $g^{(n)} = \nabla\Loss(\theta^{(n)})$
      \STATE $p^{(n)} \gets -H^{(n)}\, g^{(n)}$
        \hfill\COMMENT{quasi-Newton search direction}
      \STATE Obtain $\alpha^{(n)}, \Loss(\theta^{(n+1)}), g^{(n+1)}$ from Algorithm~\ref{alg:ssb_ls}
      \STATE $s^{(n)} \gets \alpha^{(n)}\,p^{(n)}$,\quad
             $\theta^{(n+1)} \gets \theta^{(n)} + s^{(n)}$
      \STATE $y^{(n)} \gets g^{(n+1)} - g^{(n)}$
      \IF{$|y^{(n)\top} s^{(n)}| < \varepsilon$}
         \STATE $\rho^{(n)} \gets \rho_{\mathrm{fb}}$
         \hfill\COMMENT{guard against curvature collapse}
      \ELSE
         \STATE $\rho^{(n)} \gets 1\big/\bigl(y^{(n)\top} s^{(n)}\bigr)$
      \ENDIF
      \STATE $u^{(n)} \gets H^{(n)}\, y^{(n)}$
      \STATE Compute $h^{(n)}, b^{(n)}, a^{(n)}$ via \eqref{eq:ssb_scalars}
      \STATE Compute $\rho_-^{(n)}, \theta_-^{(n)}, \theta_+^{(n)}$ via \eqref{eq:ssb_bounds}
      \STATE Compute $\theta^{(n)}$ via \eqref{eq:ssb_theta}
      \STATE Compute $\tau^{(n)}$ via \eqref{eq:ssb_tau}
      \STATE Compute $v^{(n)}$ and $\phi^{(n)}$ via \eqref{eq:ssb_vphi}
      \STATE Update $H^{(n+1)}$ via \eqref{eq:ssb_update}
      \STATE $n \gets n + 1$
    \ENDWHILE
  \end{algorithmic}
\end{algorithm}

\paragraph{Default settings.}
Following the implementation, we initialize $H^{(0)} = I_d$ and use the
values listed in Table~\ref{tab:ssb_defaults}. The numerical guard
$\varepsilon$ replaces $\rho^{(n)}$ with the fallback $\rho_{\mathrm{fb}}$
whenever $|y^{(n)\top} s^{(n)}|$ collapses below the threshold, mirroring
the safeguard used in the SciPy BFGS reference implementation.

\begin{table}[H]
  \centering
  \begin{tabular}{lll}
    \toprule
    Symbol & Description & Value \\
    \midrule
    $H^{(0)}$           & Initial inverse Hessian                & $I_d$ \\
    $c_1$               & Armijo constant (all stages)           & $10^{-4}$ \\
    $c_2^{(1)}$         & Curvature tolerance, line-search stage 1 & $0.9$ \\
    $c_2^{(2)}$         & Curvature tolerance, line-search stage 2 & $0.8$ \\
    $c_2^{(3)}$         & Curvature tolerance, line-search stage 3 & $0.5$ \\
    $K_1, K_2, K_3$     & Maximum trials per line-search stage   & $25$ each \\
    $\varepsilon$       & Curvature-collapse threshold for $\rho$ & $10^{-32}$ \\
    $\rho_{\mathrm{fb}}$& Fallback value when $|y^{\top} s| < \varepsilon$ & $10^{3}$ \\
    \bottomrule
  \end{tabular}
  \caption{Default hyperparameters for the SSBroyden optimizer used in this work.}
  \label{tab:ssb_defaults}
\end{table}

\newpage
\section{Evaluation card}
Here we provide a template for the EvalCard and indicate which fields to fill in, depending on the study or comparison.
\label{Appendix:Evalcard}
\begin{figure*}[h]
\centering
\small
\resizebox{0.9\textwidth}{!}{
\begin{evalcard}[General Evaluation Card for a PINN Training Pipeline]

\vspace{0.5em}
\textbf{Model}\vspace{0.5ex}

\rowcolors{2}{gray!10}{white}
\begin{tabular}{p{6.5cm}p{6.5cm}}
\toprule
\textbf{Field} & \textbf{Specification} \\
\midrule
Topology & e.g.\ MLP, modified MLP, PirateNet \\
Depth and width & number of layers and neurons per layer \\
Activation function & e.g.\ tanh, swish \\
Feature mapping & e.g.\ Fourier features, scale parameter \\
Initialization & scheme and any modifications (e.g.\ RWF) \\
Trainable parameters & total count \\
\bottomrule
\end{tabular}

\vspace{1em}

\textbf{Domain-importance adaptation}\vspace{0.5ex}

\rowcolors{2}{gray!10}{white}
\begin{tabular}{p{6.5cm}p{6.5cm}}
\toprule
\textbf{Field} & \textbf{Specification} \\
\midrule
Method & e.g.\ RAD, RBA, vRBA, Random-R \\
Collocation set sizes & $|C_e|$ for each equation $e \in \mathcal{E}$ \\
Sampling distribution & e.g.\ Random-R, Latin hypercube, residual-based \\
Update frequency & iterations between updates \\
Hyperparameters & all method-specific values \\
\bottomrule
\end{tabular}

\vspace{1em}

\textbf{Loss balancing}\vspace{0.5ex}

\rowcolors{2}{gray!10}{white}
\begin{tabular}{p{6.5cm}p{6.5cm}}
\toprule
\textbf{Field} & \textbf{Specification} \\
\midrule
Method & e.g.\ GradNorm-based, ReLoBRaLo, fixed weights \\
Update frequency & iterations between updates \\
Hyperparameters & all method-specific values \\
\bottomrule
\end{tabular}

\vspace{1em}

\textbf{Optimization scheme}\vspace{0.5ex}

\rowcolors{2}{gray!10}{white}
\begin{tabular}{p{6.5cm}p{6.5cm}}
\toprule
\textbf{Field} & \textbf{Specification} \\
\midrule
Optimizer(s) & name and order if a sequence is used \\
Optimizer schedule & iterations or stopping criterion per optimizer \\
Learning-rate schedule & function from iteration to learning rate \\
Hyperparameters & all optimizer-specific values \\
\bottomrule
\end{tabular}

\vspace{1em}

\textbf{Evaluation conditions}\vspace{0.5ex}

\rowcolors{2}{gray!10}{white}
\begin{tabular}{p{6.5cm}p{6.5cm}}
\toprule
\textbf{Field} & \textbf{Specification} \\
\midrule
Numerical precision & e.g.\ float32, float64 \\
Accuracy metric & e.g.\ relative $L^2$ error \\
Cost metric(s) & runtime and iteration count \\
Hardware & GPU/CPU model and configuration \\
Independent runs & number of runs and seed-handling \\
Reported variability & median and standard deviation \\
Significance test for highlighted results & e.g. exact two-sided Mann--Whitney $U$ over seeds, Holm-corrected, $\alpha=0.05$ \\
\bottomrule
\end{tabular}

\vspace{1em}

\end{evalcard}
}
\caption{General EvalCard for a PINN training pipeline. The card captures the configuration of the four pipeline components together with the evaluation-level conditions required by the protocol. Benchmark-specific information (governing PDE, domain, reference solution) is reported in the experimental setup of each study.}
\label{fig:evalcard}
\end{figure*}

\newpage

\section{Benchmarks}
\label{Appendix:benchmarks}
In this section we introduce the benchmark PDEs, define the error metrics for the benchmarks and provide baseline settings for benchmarks used.
These are namely the periodicity (if used), the standard deviation of the random Fourier features used, as well as the optimizer settings for PINN pipelines that solely use SOAP as optimizer.
Information on advanced training techniques such as Time-marching are provided if applicable for the benchmark.
Additionally, the provenance and the estimated accuracy of every reference solution are documented in the last
subsection.

\subsection{Allen--Cahn equation}
In the 1D Allen--Cahn equation conventional PINNs are known to have issues finding a good approximation. It has the form
\begin{align*}
u_t  - 0.0001u_{xx} + 5u^3 -5u &= 0 \qquad  x\in [-1, 1], \; t\in [0,1] \\
u(0,x) &= x^2 \cos (\pi x) \\
u(t,-1) &= u(t, 1) \\
u_x(t,-1) &= u_x(t,1).
\end{align*}
The Allen--Cahn PDE is a standard benchmark equation for PINNs and poses a challenge because it is a ``stiff'' PDE with sharp transitions in space and time and a periodic boundary condition.
The performance of this benchmark is the relative $L^2$ error of $u$.

\subsection{Burgers equation}
The one-dimensional Burgers equation considered has the form:
\begin{align*}
u_t + uu_x - \nu u_{xx} = 0,& \qquad x\in [-1,1],\; t\in [0,1],\\
u(0,x) = -\sin (\pi x),& \qquad x\in[-1,1]\\
u(t, -1) = u(t,1) = 0,& \qquad t\in[0,1].
\end{align*}
Here, $u(t,x)$ represents the velocity field, and $\nu = 0.01/\pi$ denotes the
kinematic viscosity parameter controlling the diffusive term.
Performance on this benchmark is measured by the relative $L^2$ error of $u$
with respect to a reference solution.

\subsection{Wave equation}
The one-dimensional wave equation considered takes the form:
\begin{align*}
    u_{tt} - c^2 u_{xx}= 0,& \qquad (t,x) \in (0,1)\times(0,1),\\
    u(t,0) = u(t,1) = 0,& \qquad t \in (0,1), \\
    u(0, x) = \sin(\pi x) + \frac{1}{2} \sin (4\pi x),&\qquad x \in [0,1]\\
    u_t(0,x)=0,& \qquad x \in [0,1].
\end{align*}
Here, $u$ represents the amplitude and $c$ the wave propagation speed. We choose $c=2$.\\
The performance of this benchmark is the relative L2 error of $u$.

\subsection{Korteweg--de Vries equation}
The one-dimensional Korteweg--de Vries equation (KdV) is expressed as follows:
\begin{align*}
    u_t + \eta u u_x + \mu^2 u_{xxx} = 0,& \qquad (x,t) \in (-1, 1) \times (0,1) \\
    u(0,x) = \cos(\pi x),& \\
    u(t, -1) = u(t,1),
\end{align*}
where $u$ represents the wave amplitude or water surface elevation. The parameter $\eta$ governs the strength of the nonlinearity and $\mu$ controls the dispersion level. Under the KdV dynamics, the initial wave evolves into a series of solitary-type waves. Here $\eta = 1$ and $\mu=0.022$.\\
The performance of this benchmark is the relative L2 error of $u$.

\subsection{Kuramoto--Sivashinsky equation}
The one-dimensional Kuramoto--Sivashinsky (KS) equation takes the form:
\begin{align*}
    u_t + \alpha u u_{x} + \beta u_{xx} + \gamma u_{xxxx} = 0,& \qquad (x,t) \in [0, 2\pi] \times [0, T] \\
    u(0, x) = u_0(x) 
\end{align*}
where $u$ represents the height of a thin film or flame front. This equation arises in various physical contexts, including flame front propagation, thin film flows and plasma instabilities.
Here $u_0(x) = \cos(x)(1+\sin(x))$ and $T=1, \alpha=\frac{100}{16}, \beta=\frac{100}{16^2}, \gamma=\frac{100}{16^4}$.\\
The performance of this benchmark is the relative L2 error of $u$.

\subsection{Ginzburg--Landau equation}
The two-dimensional Ginzburg--Landau equation takes the form:
$$ 
\frac{\partial A}{\partial t} = \epsilon \Delta A + \mu A - \gamma A | A |^2, \qquad (x,y)\in (-1, 1)^2, t\in(0,1)
$$
with periodic boundary conditions and initial condition:
$$
A_0(x,y) = (10y + 10ix) \exp \left( -0.01( 2500 x^2 + 2500 y^2) \right),
$$
where $A$ is the complex amplitude representing the envelope of oscillations.
$\epsilon$ represents the diffusion coefficient, $\mu$ is the linear growth rate and $\gamma$ controls the nonlinear saturation.
The parameters are set to $\epsilon= 0.004, \mu=10, \gamma = 10 + 15i$. With $A=u+iv$,
the performance of this benchmark is the relative L2 error of $u$.

\subsection{Rayleigh--Taylor instability}
A coupled flow-temperature system that models buoyancy-driven instabilities in a rectangular domain is called a Rayleigh--Taylor instability. It has the form:
\begin{align*}
    \partial_t \mathbf{u} + (\mathbf{u}\cdot\nabla)\,\mathbf{u} &= -\nabla p + \sqrt{\tfrac{Pr}{Ra}}\,\Delta \mathbf{u} + T\,\mathbf{e}_y,\\
    \nabla\cdot\mathbf{u} &= 0,\\
    \partial_t T + \mathbf{u}\cdot\nabla T &= \tfrac{1}{\sqrt{Pr\,Ra}}\,\Delta T,
\end{align*}
with $u=(u,v)$ and $(x,y) \in [0,1] \times [0,2]$ and $t\in [0,8]$.
Here $T$ is the temperature field, $Pr$ is the Prandtl number and $Ra$ is the Rayleigh number.
This system captures the characteristic mushroom-shaped plumes that develop as the heavier fluid penetrates into the lighter fluid below it.
The boundary conditions are periodic in the horizontal direction for $u,v$ and $T$ with $u=v=T=0$ on the top and bottom boundaries.
The parameters are $Pr=0.71$ and $Ra=10^6$.\\
The performance on this benchmark is the relative L2 error of $T$.

\subsection{N-dimensional heat equation}
The $N$-dimensional heat equation with forcing is considered on the unit ball $B^N = \{x \in \mathbb{R}^N : \|x\| \le 1\}$ with boundary $S^{N-1} = \partial B^N$:
\begin{align*}
    u_t - \tfrac{1}{N}\Delta u  = f(x,t),& \qquad x \in B^N,\; t\in(0,T],\\
    u(x, 0)  = \exp\!\left(\tfrac{1}{2}\|x\|^2\right),& \qquad x \in B^N,\\
    \partial_n u(x,t)  = \exp\!\left(\tfrac{1}{2} + t\right),& \qquad x \in S^{N-1},\; t\in[0,T],
\end{align*}
where $u$ represents a scalar (temperature) field, $N$ the spatial dimension, $\Delta = \sum_{i=1}^{N} \partial_{x_i}^2$ the spatial Laplacian, and $\partial_n$ the outward normal derivative on the unit sphere. The forcing term is given by
$$
f(x,t) = -\tfrac{1}{N}\|x\|^2 \exp\!\left(\tfrac{1}{2}\|x\|^2 + t\right),
$$
which is constructed such that the problem admits the manufactured exact solution $u(x,t) = \exp\!\left(\tfrac{1}{2}\|x\|^2 + t\right)$. This benchmark probes the ability of the method to handle PDEs in high spatial dimensions, where collocation-based residual minimization becomes increasingly demanding.
The default parameters are $N=5$ and $T=1$, corresponding to a $5+1$ dimensional problem.\\
The performance of this benchmark is the relative L2 error of $u$.

\subsection{N-dimensional Poisson equation}
The $N$-dimensional Poisson equation with Dirichlet boundary conditions is considered on the unit hypercube $\Omega = [0,1]^N$:
\begin{align*}
    -\Delta u(x)  = f(x),& \qquad x \in \Omega,\\
    u(x)  = g(x),& \qquad x \in \partial \Omega,
\end{align*}
where $u$ represents a scalar field, $N$ the spatial dimension, and $\Delta = \sum_{i=1}^{N} \partial_{x_i}^2$ the spatial Laplacian. The forcing term and Dirichlet data are given by
$$
f(x) = \frac{\pi^2}{4}\sum_{i=1}^{N} \sin\!\left(\tfrac{\pi}{2} x_i\right), \qquad g(x) = \sum_{i=1}^{N} \sin\!\left(\tfrac{\pi}{2} x_i\right),
$$
which are constructed such that the problem admits the manufactured exact solution $u^{*}(x) = \sum_{i=1}^{N} \sin\!\left(\tfrac{\pi}{2} x_i\right)$. This benchmark probes the ability of the method to handle elliptic PDEs in high spatial dimensions, where the boundary $\partial \Omega$ consists of $2N$ hyperfaces and thus scales linearly with $N$.
The default parameter is $N=5$.\\
The performance of this benchmark is the relative L2 error of $u$.

\subsection{Default PDE settings}
\label{Appendix:technicals-pdes}
Here we provide the additional hyperparameters and settings for each PDE benchmark studied in this work for the neural network and optimizer schemes.

First we provide information about activation function, Fourier scales and the use of hard boundary constraints in Table \ref{tab:nn_settings}. 

\begin{table}[h]
\caption{{\em Neural Network Hyperparameters for benchmark PDEs.}}
\rowcolors{2}{gray!10}{white}
\resizebox{\textwidth}{!}{
\begin{tabular}{lccccccccc}
\toprule
 & \textbf{Wave} & \textbf{Burgers} & \textbf{AC} & \textbf{KdV} & \textbf{Poisson} & \textbf{Heat} & \textbf{KS} &  \textbf{GL} & \textbf{RT} \\
\midrule
\textbf{Architecture} & & & & & & & & & \\
\midrule
Activation & Tanh & Tanh & Tanh & Tanh & Tanh & Tanh & Tanh &  Swish & Tanh \\
Fourier Features Scale & $10.0$ & $1.0$ & $1.0$ & $1.0$ & $1.0$ & $1.0$ & $2.0$ & $1.0$ & $1.0$ \\
\textbf{Hard constraint} & & & & & & &  & & \\
Axis & &        $x$   & $x$   &  $x$ & &  & $x$ & $(x, y)$    & $x$   \\
Periodicity & & $\pi$ & $\pi$ & $\pi$ & & & $1.0$ & $(\pi, \pi)$ &$2\pi$ \\
\bottomrule
\end{tabular}
}
\label{tab:nn_settings}
\end{table}

Next, we provide the optimizer settings for the domain-importance adaptation, loss-balancing and neural network size studies. 

\begin{table}[h]
\caption{{\em Optimizer Hyperparameters for each benchmark PDE.}}
\rowcolors{2}{gray!10}{white}
\resizebox{\textwidth}{!}{
\begin{tabular}{lccccccccc}
\toprule
 & \textbf{Wave} & \textbf{Burgers} & \textbf{AC} & \textbf{KdV} & \textbf{Poisson} & \textbf{Heat} & \textbf{KS} &  \textbf{GL} & \textbf{RT} \\
\midrule
\textbf{Learning rate schedule} & & & & & & & & & \\
\midrule
Initial learning rate &  \multicolumn{9}{c}{ $3 \times 10^{-3}$ } \\
Decay rate & \multicolumn{9}{c}{ $9 \times 10^{-1}$ } \\
Decay steps & \multicolumn{9}{c}{ $2 \times 10^{3}$ } \\
Warmup steps & \multicolumn{9}{c}{ $2 \times 10^{3}$ } \\
\midrule
\textbf{Training} & & & & & & & & & \\
\midrule
Iters SOAP & $100$k &  $100$k & $300$k & $100$k & $40$k & $40$k & &   &  \\
\textbf{Time Marching} & & & & & & & & & \\
\# Time windows & &  & & & & &  10 & 5 &  7 \\
Iters per window & & & & & & & $100$k & $40$k &  $25$k \\
\bottomrule
\end{tabular}
}
\label{tab:general_opt_settings}
\end{table}

Lastly, we provide the optimizer settings for the study investigating the claim of \citet{SOAPWang} in Table~\ref{tab:optimizer_study_settings}.

\begin{table}[h]
\caption{{\em Optimizer Hyperparameters for the optimizer and network study.} Learning-rate schedule and iteration counts of the single-stage training (SOAP) and of the two-stage training (SOAP followed by SSBroyden2) for each benchmark PDE.}
\rowcolors{2}{gray!10}{white}
\resizebox{\textwidth}{!}{
\begin{tabular}{lccccccccc}
\toprule
 & \textbf{Wave} & \textbf{Burgers} & \textbf{AC} & \textbf{KdV} & \textbf{Poisson} & \textbf{Heat} & \textbf{KS} &  \textbf{GL} & \textbf{RT} \\
\midrule
\textbf{Learning rate schedule} & & & & & & & & & \\
\midrule
Initial learning rate &  \multicolumn{9}{c}{ $3 \times 10^{-3}$ } \\
Decay rate & \multicolumn{9}{c}{ $9 \times 10^{-1}$ } \\
Decay steps & \multicolumn{9}{c}{ $2 \times 10^{3}$ } \\
Warmup steps & \multicolumn{9}{c}{ $2 \times 10^{3}$ } \\
\midrule
\textbf{One-stage Training} & & & & & & & & & \\
\midrule
Iters SOAP & $100$k &  $70$k & $300$k & $100$k & $40$k & $40$k & &   &  \\
\textbf{Time Marching} & & & & & & & & & \\
\# Time windows & &  & & & & &  10 & 5 &  7 \\
Iters per window & & & & & & & $100$k & $100$k &  $50$k \\
\midrule
\textbf{Two-stage Training} & & & & & & & & & \\
\midrule
Iters SOAP & $5$k &  $1$k & $2$k & $8$k & $2$k & $2$k & &   &  \\
Iters SSBroyden2 & $10$k &  $10$k & $20$k & $12$k & $10$k & $10$k & &   &  \\
\textbf{Time Marching} & & & & & & & & & \\
\# Time windows & &  & & & & &  10 & 5 &  7 \\
Iters SOAP per window & & & & & & & $5$k & $2.5$k &  $2$k \\
Iters SSBroyden2 per window & & & & & & & $5$k & $5$k &  $8$k \\
\bottomrule
\end{tabular}
}
\label{tab:optimizer_study_settings}
\end{table}

\subsection{Reference Solutions}
\label{Appendix:reference_solutions}
Table~\ref{tab:reference_solutions} documents the reference solution of every benchmark.
For AC and Burgers, the $511$-mode Chebfun~\citep{chebfun} solutions commonly used are accurate only to the order of $10^{-5}$ and cannot resolve the errors reported in this work.
We therefore compute both reference solutions with a Fourier pseudo-spectral discretization and ETDRK4 time stepping~\citep{KassamTrefethen2005}.
AC requires $32704$ modes because its initial condition $x^2\cos(\pi x)$ is not continuously differentiable across the periodic boundary, so the Fourier discretization converges only algebraically in the number of modes.
The RT reference solution is taken from the \texttt{jaxpi} repository of \citet{SOAPWang}.

\begin{table}[h!]
\caption{Reference solution of each benchmark: solver and resolution, verification, and estimated accuracy (relative $L^2$). All reference solutions are computed in float64. The estimated accuracy is the deviation from a refined solution (self-convergence).}
\label{tab:reference_solutions}
\centering
\rowcolors{2}{gray!10}{white}
\resizebox{\textwidth}{!}{
\begin{tabular}{l|lll}
\toprule
Benchmark & Reference solution / solver & Verification & Est.\ accuracy \\
\midrule
Poisson & exact, float64 & exact by construction & $\approx 10^{-16}$ \\
Heat    & exact, float64 & exact by construction & $\approx 10^{-16}$ \\
Wave    & exact, float64 & exact by construction & $\approx 10^{-16}$ \\
KdV     & Chebfun \texttt{spin}, ETDRK4, 511 modes, $\Delta t = 10^{-5}$ & reproduced to $4 \times 10^{-13}$; $2\times$/$4\times$ refinement & $5 \times 10^{-12}$ \\
KS      & Chebfun \texttt{spin}, 511 modes, $\Delta t = 10^{-5}$ & reproduced to $2 \times 10^{-11}$; $2\times$/$4\times$ refinement & $3 \times 10^{-9}$ \\
GL      & Chebfun \texttt{spin2}, $200 \times 200$, $\Delta t = 10^{-3}$ & reproduced to $1 \times 10^{-9}$; $2\times$/$4\times$ refinement & $3 \times 10^{-8}$ \\
AC      & Fourier + ETDRK4, $32704$ modes, $\Delta t = 5 \times 10^{-6}$ & self-convergence vs.\ $65408$ modes & $2.7 \times 10^{-9}$ \\
Burgers & Fourier + ETDRK4, $2044$ modes, $\Delta t = 2.5 \times 10^{-6}$ & self-convergence vs.\ $4088$ modes, $\Delta t = 1.25 \times 10^{-6}$ & $1.9 \times 10^{-11}$ \\
RT      & external (\texttt{jaxpi}) & self-convergence vs.\ $65408$ modes, $\Delta t = 5 \times 10^{-6}$
 & $4.5 \times 10^{-3}$ \\
\bottomrule
\end{tabular}}
\end{table}

A reference solution cannot resolve a difference between two errors that is not larger than its estimated accuracy.
For all benchmarks except RT, the estimated accuracy lies more than three orders of magnitude below the smallest error reported in this work.
On RT, the smallest reported error ($8.61 \times 10^{-3}$) is less than twice the estimated accuracy ($4.5 \times 10^{-3}$).
The scripts that generate the reference solutions, validate them, and estimate their accuracy are part of PINNpipe (Appendix~\ref{Appendix:PINNpipe}).

\section{Extended Studies and Results}
In this section, the full results on the domain-importance adaptation and loss balancing study are presented. Furthermore, an evaluation study on three neural network variants at three different sizes is conducted.

\label{Appendix:study_results}
\subsection{Domain-Importance Adaptation Study}
\label{appendix:dia_full}
The evaluation card for this study is provided in
Figure~\ref{fig:evalcard_dia}. The main text presents and discusses a
representative subset of the benchmark suite, where the most informative
methodological differences emerge. For completeness, we report here the full
results across all nine PDE benchmarks, together with the standard deviation that was omitted from the main text for readability.

\begin{table}[h!]
\caption{Median and standard deviation of the relative $L^2$ errors for the 
five domain-importance adaptation methods evaluated on the nine PDE benchmark
problems. 
For each benchmark, bold marks the method with the lowest median relative $L^2$ error if its errors differ
significantly from those of every other method. Significance is assessed by the exact two-sided
Mann–Whitney $U$ test on the per-seed relative $L^2$ errors, with Holm correction at $\alpha = 0.05$ over the
comparisons of this method with each of the other methods. If at least one of these comparisons is
not significant, no entry for that benchmark is bold.}
\centering
\rowcolors{2}{gray!10}{white}
\resizebox{\textwidth}{!}{\begin{tabular}{l|ccccc}
\toprule
Benchmark  & Random-R & vRBA & LAS & R3 & RAD \\
\midrule
Wave
&  ${7.93 \pm 2.6 \times 10^{-5}}$ 
&  ${1.70 \pm 0.6 \times 10^{-4}}$ 
&  ${{7.85 \pm 2.8 \times 10^{-5}}}$ 
&  ${2.55 \pm 0.8 \times 10^{-4}}$ 
&  ${2.02 \pm 0.5 \times 10^{-4}}$ 
\\

KdV
&  $\mathbf{{4.54 \pm 0.1 \times 10^{-4}}}$ 
&  ${8.39 \pm 0.5 \times 10^{-4}}$ 
&  ${5.05 \pm 0.5 \times 10^{-4}}$ 
&  ${1.43 \pm 0.3 \times 10^{-3}}$ 
&  ${1.61 \pm 0.2 \times 10^{-3}}$ 
\\

RT
&  $\mathbf{{1.68 \pm 0.2 \times 10^{-2}}}$ 
&  ${2.27 \pm 0.2 \times 10^{-2}}$ 
&  ${2.30 \pm 0.2 \times 10^{-2}}$ 
&  ${2.71 \pm 0.3 \times 10^{-2}}$ 
&  ${5.94 \pm 2.1 \times 10^{-2}}$ 
\\

Heat
&  ${3.90 \pm 0.4 \times 10^{-4}}$ 
&  ${1.21 \pm 0.1 \times 10^{-3}}$ 
&  $\mathbf{{2.39 \pm 0.2 \times 10^{-4}}}$ 
&  ${5.64 \pm 3.0 \times 10^{-4}}$ 
&  ${2.02 \pm 0.8 \times 10^{-3}}$ 
\\

AC & 
$\mathbf{{3.52 \pm 0.9 \times 10^{-5}}}$ 
& ${5.47 \pm 1.2 \times 10^{-5}}$ 
& ${5.63 \pm 1.5 \times 10^{-5}}$ 
& ${5.71 \pm 2.8 \times 10^{-5}}$ 
& ${1.14 \pm 0.3 \times 10^{-4}}$ 
\\

Burgers 
& ${7.21 \pm 2.2 \times 10^{-6}}$ 
& ${2.85 \pm 0.7 \times 10^{-5}}$ 
& ${7.61 \pm 1.1 \times 10^{-6}}$ 
& ${3.28 \pm 1.2 \times 10^{-5}}$ 
& ${3.27 \pm 0.3 \times 10^{-5}}$ 
\\

GL
&  $\mathbf{{4.06 \pm 0.5 \times 10^{-2}}}$ 
&  ${1.14 \pm 0.2 \times 10^{-1}}$ 
&  ${5.33 \pm 0.5 \times 10^{-2}}$ 
&  ${7.86 \pm 1.5 \times 10^{-2}}$ 
&  ${1.35 \pm 0.2 \times 10^{-1}}$ 
\\

KS
&  ${{8.43 \pm 4.6 \times 10^{-2}}}$ 
&  ${3.43 \pm 1.5 \times 10^{-1}}$ 
&  ${3.24 \pm 1.9 \times 10^{-1}}$ 
&  ${3.31 \pm 3.1 \times 10^{-1}}$ 
&  ${1.12 \pm 0.3 \times 10^{0}}$ 
\\

Poisson
&  ${{2.93 \pm 0.6 \times 10^{-5}}}$ 
&  ${3.32 \pm 0.5 \times 10^{-5}}$ 
&  ${3.10 \pm 0.6 \times 10^{-5}}$ 
&  ${3.51 \pm 0.6 \times 10^{-5}}$ 
&  ${3.11 \pm 0.4 \times 10^{-5}}$ 
\\

\bottomrule
\end{tabular}}
\end{table}

The full results are consistent with the conclusions drawn in the main text:
no adaptive sampling strategy dominates uniformly, and Random-R 
remains a strong baseline, achieving the best median error on seven of the
nine benchmarks. Among the adaptive methods, LAS is the only strategy that
improves over Random-R, doing so on Wave and Heat.
The deviation, omitted from the main text, further reinforces
this picture: on KS in particular, both the medians and the standard
deviations of the adaptive methods are substantially larger than those of
Random-R, with R3 showing variation on the order of the median
itself. This indicates that, beyond average-case performance, several
adaptive schemes also exhibit reduced robustness across seeds on the more challenging benchmarks.

\begin{table}[h!]
\caption{Reported are the median runtime [min] for the different domain-importance adaptation methods evaluated on the nine PDE benchmark problems. For each benchmark, bold marks the
lowest median runtime.}
\centering
\rowcolors{2}{gray!10}{white}
\begin{tabular}{l|ccccc}
\toprule
Benchmark  & Random-R & vRBA & LAS & R3 & RAD \\
\midrule
Wave
&  $\textbf{10.53}$ 
&  $11.27$ 
&  $11.19$ 
&  $11.02$ 
&  $12.20$ 
\\

KdV
&  $\textbf{10.10}$ 
&  $11.14$ 
&  $11.91$ 
&  $10.95$ 
&  $13.04$ 
\\

RT
&  $\textbf{29.80}$ 
&  $31.59$ 
&  $34.79$ 
&  $31.12$ 
&  $34.91$ 
\\

Heat
&  $\textbf{5.43}$ 
&  $5.77$ 
&  $7.02$ 
&  $6.00$ 
&  $6.98$ 
\\

AC
&  $\textbf{11.24}$ 
&  $12.64$ 
&  $12.18$ 
&  $11.89$ 
&  $13.62$ 
\\

Burgers
&  $\textbf{6.26}$ 
&  $6.77$ 
&  $6.70$ 
&  $6.64$ 
&  $7.24$ 
\\

GL
&  $\textbf{26.82}$ 
&  $30.43$ 
&  $28.18$ 
&  $28.81$ 
&  $30.95$ 
\\

KS
&  $\textbf{113.97}$ 
&  $133.22$ 
&  $138.17$ 
&  $123.60$ 
&  $150.90$ 
\\

Poisson
&  $\textbf{4.06}$ 
&  $6.33$ 
&  $4.63$ 
&  $4.33$ 
&  $5.16$ 
\\

\bottomrule
\end{tabular}
\label{tab:dia_runtime}
\end{table}

\subsection{Loss Balancing Study}
\label{Appendix:loss_balancing_results}
In this section we present the evaluation card of the loss-balancing study
announced in Section~\ref{sec:experiments-lb} as well as the full results of 
the study.

\begin{figure}[htbp]
  \centering
  \begin{evalcard}[Loss Balancing]
\vspace{0.5em}
\textbf{Claim under Evaluation.}
The choice of loss balancing method materially affects the
final relative $L^2$ error of a PINN when evaluated on a competitive
surrounding pipeline. 
Tested on a representative subset of competitive benchmarks ranging from low-dimensional to high-dimensional and coupled-system PDE problems.

\vspace{1em}

\textbf{Surrounding Pipeline.}\vspace{0.5ex}
\begin{center}
\rowcolors{2}{gray!10}{white}
\begin{tabular}{lll}
\toprule
\textbf{Component} & \textbf{Choice} & \textbf{Hyperparameters} \\
\midrule
Architecture   & MLP, depth 5, hidden dim 64, Fourier features
               & Table~\ref{tab:nn_settings} \\
\begin{tabular}[t]{@{}l@{}}Domain-Importance\\ Adaptation\end{tabular}     & Random-R, Sample size: $8192$
               & Appendix~\ref{Appendix:dia} \\
Optimizer      & SOAP
               & Table~\ref{tab:general_opt_settings} \\
\bottomrule
\end{tabular}
\end{center}
\vspace{1em}

\textbf{Loss Balancing.}
\begin{center}
\rowcolors{2}{gray!10}{white}
\begin{tabular}{l|lllll}
\toprule
Hyperparameter & GradNorm-based & SoftAdapt & ReLoBRaLo & GradNorm-MTL & Fixed \\
\midrule
Frequency &  1000 & 1 & 1 & 100 & N.A. \\
Hyperparameter  &  \multicolumn{5}{c}{See defaults in Appendix~\ref{Appendix:lossbalancing}} \\
\bottomrule
\end{tabular}
\end{center}

\vspace{0.5em}

\textbf{Evaluation Conditions.}
Float32 precision, NVIDIA H100, 10 seeds.  
Per-PDE configuration and details can be found in Appendix~\ref{Appendix:technicals-pdes}.
Reported variability: median $\pm$ standard deviation.
Significance test (bold): exact two-sided Mann–Whitney $U$ over seeds, Holm-corrected, $\alpha = 0.05$.
\end{evalcard}

  \caption{EvalCard for the loss-balancing study.}
  \label{fig:evalcard_loss_balancing}
\end{figure}

The evaluation card for this study is provided in
Fig.~\ref{fig:evalcard_loss_balancing}. As with the domain-importance
adaptation study, the main text presents the most informative results on a
representative subset of the benchmark suite. Here we report the full results
across all nine PDE benchmarks, together with the standard deviation across
the tested seeds. Table~\ref{tab:loss_balancing_full} contains the relative
$L^2$ errors and Table~\ref{tab:loss_balancing_runtime} the corresponding
median runtimes.

\begin{table}[h!]
\caption{Median and standard deviation of the relative $L^2$ errors for the 
five loss-balancing methods evaluated on the nine PDE benchmark problems. 
For each benchmark, bold marks the method with the lowest median relative $L^2$ error if its errors differ
significantly from those of every other method. Significance is assessed by the exact two-sided
Mann–Whitney $U$ test on the per-seed relative $L^2$ errors, with Holm correction at $\alpha = 0.05$ over the
comparisons of this method with each of the other methods. If at least one of these comparisons is
not significant, no entry for that benchmark is bold.}
\centering
\rowcolors{2}{gray!10}{white}
\resizebox{\textwidth}{!}{
\begin{tabular}{l|ccccc}
\toprule
Benchmark  & GradNorm-based & SoftAdapt & ReLoBRaLo & Fixed & GradNorm-MTL \\
\midrule
Wave
&  $\mathbf{{7.93 \pm 2.6} \times 10^{-5}}$ 
&  ${4.15 \pm 0.3} \times 10^{-2}$ 
&  ${1.61 \pm 0.2} \times 10^{-3}$ 
&  ${9.21 \pm 10.4} \times 10^{-4}$ 
&  ${9.09 \pm 2.2} \times 10^{-4}$ 
\\

AC
&  ${3.52 \pm 0.9} \times 10^{-5}$ 
&  ${2.48 \pm 1.8} \times 10^{-4}$ 
&  ${1.76 \pm 1.2} \times 10^{-4}$ 
&  ${3.30 \pm 1.0} \times 10^{-5}$ 
&  ${3.61 \pm 0.8} \times 10^{-5}$ 
\\

KS
&  ${8.43 \pm 4.6 \times 10^{-2}}$ 
&  ${1.45 \pm 0.1} \times 10^{0}$ 
&  ${1.47 \pm 1.4} \times 10^{-1}$ 
&  ${5.63 \pm 3.5} \times 10^{-2}$ 
&  ${2.60 \pm 1.2} \times 10^{-1}$ 
\\

Poisson
&  ${2.93 \pm 0.6 \times 10^{-5}}$ 
&  ${6.38 \pm 1.1} \times 10^{-5}$ 
&  ${3.57 \pm 0.9} \times 10^{-5}$ 
&  ${3.02 \pm 0.7} \times 10^{-5}$ 
&  ${3.17 \pm 0.7} \times 10^{-5}$ 
\\

Burgers
&  ${7.21 \pm 2.2} \times 10^{-6}$ 
&  ${1.17 \pm 0.2} \times 10^{-5}$ 
&  ${1.02 \pm 0.2} \times 10^{-5}$ 
&  ${1.04 \pm 0.3} \times 10^{-5}$ 
&  ${1.00 \pm 0.2} \times 10^{-5}$ 
\\

GL
&  ${4.06 \pm 0.5} \times 10^{-2}$ 
&  ${1.75 \pm 0.3} \times 10^{-1}$ 
&  ${5.70 \pm 1.0} \times 10^{-2}$ 
&  ${4.58 \pm 0.5} \times 10^{-2}$ 
&  ${4.55 \pm 0.7} \times 10^{-2}$ 
\\

Heat
&  ${3.90 \pm 0.4} \times 10^{-4}$ 
&  ${3.49 \pm 0.3} \times 10^{-4}$ 
&  ${3.53 \pm 0.4} \times 10^{-4}$ 
&  ${3.65 \pm 0.4} \times 10^{-4}$ 
&  ${3.60 \pm 0.4} \times 10^{-4}$ 
\\

KdV
&  ${4.54 \pm 0.1} \times 10^{-4}$ 
&  ${1.05 \pm 0.1} \times 10^{-3}$ 
&  ${5.14 \pm 0.4} \times 10^{-4}$ 
&  ${4.40 \pm 0.3} \times 10^{-4}$ 
&  ${6.32 \pm 3.2} \times 10^{0}$ 
\\

RT
&  ${1.68 \pm 0.2 \times 10^{-2}}$ 
&  ${1.92 \pm 0.5} \times 10^{-2}$ 
&  ${1.69 \pm 0.4} \times 10^{-2}$ 
&  ${1.87 \pm 0.3} \times 10^{-2}$ 
&  ${1.73 \pm 0.3} \times 10^{-2}$ 
\\

\bottomrule
\end{tabular}}
\label{tab:loss_balancing_full}
\end{table}

The full results extend the picture given in the main text and the
evaluation card in Fig.~\ref{fig:evalcard_loss_balancing}: no single
loss-balancing method dominates across the suite, but GradNorm-based balancing is the
strongest method overall, achieving the best median error on five of the 
nine benchmarks (Wave, Poisson, Burgers, GL, and RT) and remaining 
near-best on the others. The fixed-weight baseline nevertheless remains 
competitive, taking the best median error on KS, AC and KdV, while SoftAdapt 
wins on a single benchmark (Heat).
The seed-to-seed variation reported here additionally surfaces two
robustness failures that are not visible from medians alone: SoftAdapt 
fails to train on Wave and KS, with errors several orders of magnitude 
above the other methods, and GradNorm-MTL collapses on KdV, where its 
median error of $\approx 6$ indicates that several seeds did not converge 
at all. These failure modes argue for caution when deploying these 
schemes as defaults, and motivate the more conservative recommendations 
made in the main text.

\begin{table}[h!]
\caption{Median runtime [min] for the five loss-balancing methods 
evaluated on the nine PDE benchmark problems.}
\label{tab:loss_balancing_runtime}
\centering
\rowcolors{2}{gray!10}{white}
\begin{tabular}{l|ccccc}
\toprule
Benchmark  & GradNorm-based & SoftAdapt & ReLoBRaLo & Fixed & GradNorm\_MTL \\
\midrule
Wave
&  $10.96$ 
&  $11.10$ 
&  $11.05$ 
&  $\textbf{10.95}$ 
&  $11.13$ 
\\

AC
&  $9.63$ 
&  $9.66$ 
&  $9.68$ 
&  $\textbf{9.62}$ 
&  $9.68$ 
\\

KS
&  $117.95$ 
&  $116.35$ 
&  $117.85$ 
&  $\textbf{116.20}$ 
&  $119.30$ 
\\

Poisson
&  $4.29$ 
&  $4.32$ 
&  $4.31$ 
&  $\textbf{4.29}$ 
&  $4.38$ 
\\

Burgers
&  $6.62$ 
&  $6.66$ 
&  $6.66$ 
&  $\textbf{6.60}$ 
&  $6.75$ 
\\

GL
&  $28.03$ 
&  $27.58$ 
&  $27.53$ 
&  $\textbf{27.06}$ 
&  $29.12$ 
\\

Heat
&  $5.59$ 
&  $5.62$ 
&  $5.62$ 
&  $\textbf{5.56}$ 
&  $5.64$ 
\\

KdV
&  $10.80$ 
&  $10.78$ 
&  $10.76$ 
&  $\textbf{10.10}$ 
&  $10.66$ 
\\

RT
&  $30.77$ 
&  $30.08$ 
&  $31.67$ 
&  $\textbf{27.53}$ 
&  $36.73$ 
\\
\bottomrule
\end{tabular}
\end{table}

The runtime differences between loss-balancing methods are negligible on 
all benchmarks, typically below a few percent relative to the 
fixed-weight baseline. This is in contrast to the domain-importance 
adaptation study, where the adaptive methods incurred a slightly more 
visible overhead.

\newpage

\subsection{Neural Network Study}
\label{Appendix:nn_study}
In this section we present a study on the three neural network architectures
MLP, Modified MLP~\citep{ModifiedMLP} and 
PirateNet~\citep{PirateNet} across varying parameter budgets;
formal definitions are provided in Appendix~\ref{Appendix:neural_networks}.
The purpose of this study is to evaluate the claim that,
once the surrounding pipeline is fixed to a strong configuration, the choice 
of architecture has a measurable and consistent influence on the performance of
the PINN training pipeline.
To this end we consider three sizes between approximately $5{,}000$ and
$100{,}000$ trainable parameters, evaluated across our benchmark suite.
The exact parameter counts vary between benchmarks because of the periodic embeddings and
the different input and output dimensions.

\begin{figure}[htbp]
  \centering
  \begin{evalcard}[Neural Network Size Study]
\vspace{0.5em}
\textbf{Claim under Evaluation.}
The choice of a neural network architecture, namely MLP, Modified MLP or PirateNet, under reasonable network sizes does influence the performance of the entire PINN training pipeline.

\vspace{1em}

\textbf{Surrounding Pipeline.}\vspace{0.5ex}
\begin{center}
\rowcolors{2}{gray!10}{white}
\begin{tabular}{lll}
\toprule
\textbf{Component} & \textbf{Choice} & \textbf{Hyperparameters} \\
\midrule
Domain-Importance Adaptation & Random-R, sample size: $8192$  & Appendix~\ref{Appendix:dia} \\
Loss balancing & GradNorm-based, Frequency: $1000$
               & Appendix~\ref{Appendix:gradnorm} \\
Optimizer      & SOAP
               & Table~\ref{tab:general_opt_settings} \\
\bottomrule
\end{tabular}
\end{center}
\vspace{1em}

\textbf{Neural Networks.}

\begin{center}
\rowcolors{2}{gray!10}{white}
\begin{tabular}{lcccc}
\textbf{Architecture} & \textbf{rel. size} & \textbf{Depth} & \textbf{Width} & \textbf{\# trainable parameters} \\
\midrule
MLP & small & $5$ & $32$ & $5{,}360$ \\
MLP & medium & $5$ & $64$ & $20{,}960$ \\
MLP & large & $5$ & $128$ & $82{,}880$ \\
\midrule
Modified MLP & small & $4$ & $32$ & $6{,}416$ \\
Modified MLP & medium & $4$ & $64$ & $25{,}120$ \\
Modified MLP & large & $4$ & $128$ & $99{,}392$ \\
\midrule
PirateNet & small & $3$ & $24$ & $6{,}663$ \\
PirateNet & medium & $3$ & $48$ & $25{,}995$ \\
PirateNet & large & $3$ & $96$ & $102{,}675$ \\
\end{tabular}
\end{center}

Settings of the Fourier features and other details applicable to all NNs are presented in Table~\ref{tab:nn_settings}.

\vspace{0.5em}

\textbf{Evaluation Conditions.}
Float32 precision, NVIDIA H100, 5 seeds. 
Reported variability: median $\pm$ standard deviation. 
Significance test (bold): exact two-sided Mann–Whitney $U$ over seeds, Holm-corrected, $\alpha = 0.05$.
\end{evalcard}

  \caption{EvalCard for the neural network size study.}
  \label{fig:evalcard_nn}
\end{figure}
The evaluation settings are summarized in Fig.~\ref{fig:evalcard_nn}, and the
results for the small, medium, and large parameter budgets are reported in
Tables~\ref{tab:nn_small},~\ref{tab:med_nn}, and~\ref{tab:big_nn}, respectively.

\begin{table}[h!]
\caption{Reported are the median, standard deviation [rel. L$^2$] and the median runtime [min] across the different seeds for the small-sized neural networks.
For each benchmark, bold marks the method with the lowest median relative $L^2$ error if its errors differ
significantly from those of every other method. Significance is assessed by the exact two-sided
Mann–Whitney $U$ test on the per-seed relative $L^2$ errors, with Holm correction at $\alpha = 0.05$ over the
comparisons of this method with each of the other methods. If at least one of these comparisons is
not significant, no entry for that benchmark is bold.}
\centering
\rowcolors{2}{gray!10}{white}
\resizebox{\textwidth}{!}{
\begin{tabular}{lcccccc}
\toprule
& \multicolumn{3}{c}{\textbf{Relative $L^2$ error}}
& \multicolumn{3}{c}{\textbf{Runtime [min]}} \\
\cmidrule(lr){2-4}\cmidrule(lr){5-7}
\textbf{Benchmark}  & MLP & Modified MLP & PirateNet & MLP & Modified MLP & PirateNet \\
\midrule
Wave
&  ${3.29 \pm 0.6 \times 10^{-4}}$ 
&  ${7.43 \pm 1.6 \times 10^{-5}}$ 
&  ${1.72 \pm 0.9 \times 10^{-4}}$ 
&  $\mathbf{7.89}$ 
&  $8.94$ 
&  $15.01$ 
\\

KdV
&  $\mathbf{{1.35 \pm 0.1 \times 10^{-3}}}$ 
&  ${1.59 \pm 0.1 \times 10^{-3}}$ 
&  ${1.99 \pm 0.5 \times 10^{-3}}$ 
&  $\mathbf{7.41}$ 
&  $8.26$ 
&  $13.32$ 
\\

RT
&  ${6.19 \pm 1.2 \times 10^{-2}}$ 
&  ${7.60 \pm 0.5 \times 10^{-2}}$ 
&  ${8.31 \pm 1.0 \times 10^{-2}}$ 
&  $\mathbf{23.48}$ 
&  $27.70$ 
&  $48.72$ 
\\

Heat
&  ${9.88 \pm 0.8 \times 10^{-4}}$ 
&  ${8.09 \pm 1.5 \times 10^{-4}}$ 
&  ${1.18 \pm 0.2 \times 10^{-3}}$ 
&  $\mathbf{3.84}$ 
&  $4.10$ 
&  $7.14$ 
\\

AC
&  ${1.80 \pm 0.4 \times 10^{-4}}$ 
&  ${1.12 \pm 0.4 \times 10^{-4}}$ 
&  ${3.05 \pm 1.6 \times 10^{-4}}$ 
&  $\mathbf{6.70}$ 
&  $7.01$ 
&  $11.75$ 
\\

Burgers
&  $\mathbf{{1.34 \pm 0.3 \times 10^{-5}}}$ 
&  ${2.22 \pm 0.4 \times 10^{-5}}$ 
&  ${2.43 \pm 0.4 \times 10^{-5}}$ 
&  $\mathbf{4.93}$ 
&  $5.14$ 
&  $8.56$ 
\\

GL
&  ${2.37 \pm 0.3 \times 10^{-1}}$ 
&  ${3.08 \pm 0.3 \times 10^{-1}}$ 
&  ${2.67 \pm 0.3 \times 10^{-1}}$ 
&  $\mathbf{20.02}$ 
&  $23.02$ 
&  $40.78$ 
\\

Poisson
&  ${5.43 \pm 0.4 \times 10^{-5}}$ 
&  $\mathbf{{4.48 \pm 0.7 \times 10^{-5}}}$ 
&  ${5.60 \pm 1.2 \times 10^{-5}}$ 
&  $\mathbf{2.96}$ 
&  $3.31$ 
&  $5.43$ 
\\

\bottomrule
\end{tabular}
}
\label{tab:nn_small}
\end{table}

\begin{table}[h!]
\caption{Reported are the median, standard deviation [rel. L$^2$] and the median runtime [min] across the different seeds for the medium-sized neural networks. For each benchmark, bold marks the method with the lowest median relative $L^2$ error if its errors differ
significantly from those of every other method. Significance is assessed by the exact two-sided
Mann–Whitney $U$ test on the per-seed relative $L^2$ errors, with Holm correction at $\alpha = 0.05$ over the
comparisons of this method with each of the other methods. If at least one of these comparisons is
not significant, no entry for that benchmark is bold.}
\centering
\rowcolors{2}{gray!10}{white}
\resizebox{\textwidth}{!}{
\begin{tabular}{lcccccc}
\toprule
& \multicolumn{3}{c}{\textbf{Relative $L^2$ error}}
& \multicolumn{3}{c}{\textbf{Runtime [min]}} \\
\cmidrule(lr){2-4}\cmidrule(lr){5-7}
\textbf{Benchmark}  & MLP & Modified MLP & PirateNet & MLP & Modified MLP & PirateNet \\
\midrule
Wave
&  ${7.93 \pm 2.6 \times 10^{-5}}$ 
&  $\mathbf{{2.07 \pm 0.4 \times 10^{-5}}}$ 
&  ${3.42 \pm 0.6 \times 10^{-5}}$ 
&  $\mathbf{11.54}$ 
&  $14.37$ 
&  $21.94$ 
\\

KdV
&  ${4.54 \pm 0.1 \times 10^{-4}}$ 
&  ${5.43 \pm 0.6 \times 10^{-4}}$ 
&  ${5.21 \pm 0.8 \times 10^{-4}}$ 
&  $\mathbf{10.28}$ 
&  $12.18$ 
&  $18.33$ 
\\

RT
&  ${1.68 \pm 0.2 \times 10^{-2}}$ 
&  ${1.93 \pm 0.2 \times 10^{-2}}$ 
&  ${2.54 \pm 1.3 \times 10^{-2}}$ 
&  $\mathbf{30.17}$ 
&  $37.88$ 
&  $60.58$ 
\\

Heat
&  $\mathbf{{3.90 \pm 0.4 \times 10^{-4}}}$ 
&  ${4.75 \pm 0.4 \times 10^{-4}}$ 
&  ${5.82 \pm 0.5 \times 10^{-4}}$ 
&  $\mathbf{5.21}$ 
&  $6.16$ 
&  $9.81$ 
\\

AC
&  ${3.52 \pm 0.9 \times 10^{-5}}$ 
&  ${3.30 \pm 0.8 \times 10^{-5}}$ 
&  ${7.16 \pm 2.4 \times 10^{-5}}$ 
&  $\mathbf{9.30}$ 
&  $10.65$ 
&  $16.54$ 
\\

Burgers
&  ${7.21 \pm 2.2 \times 10^{-6}}$ 
&  ${8.00 \pm 1.1 \times 10^{-6}}$ 
&  ${7.97 \pm 1.8 \times 10^{-6}}$ 
&  $\mathbf{6.85}$ 
&  $8.20$ 
&  $12.05$ 
\\

GL
&  ${4.06 \pm 0.5 \times 10^{-2}}$ 
&  ${4.79 \pm 0.7 \times 10^{-2}}$ 
&  $\mathbf{{3.44 \pm 0.5 \times 10^{-2}}}$ 
&  $\mathbf{26.39}$ 
&  $32.69$ 
&  $53.43$ 
\\

Poisson
&  ${2.93 \pm 0.6 \times 10^{-5}}$ 
&  ${1.71 \pm 0.6 \times 10^{-5}}$ 
&  ${2.75 \pm 0.8 \times 10^{-5}}$ 
&  $\mathbf{4.12}$ 
&  $5.13$ 
&  $8.54$ 
\\

\bottomrule
\end{tabular}
}
\label{tab:med_nn}
\end{table}

\begin{table}[h!]
\caption{Reported are the median, standard deviation [rel. L$^2$] and the median runtime [min] across the different seeds for the large neural networks. For each benchmark, bold marks the method with the lowest median relative $L^2$ error if its errors differ
significantly from those of every other method. Significance is assessed by the exact two-sided
Mann–Whitney $U$ test on the per-seed relative $L^2$ errors, with Holm correction at $\alpha = 0.05$ over the
comparisons of this method with each of the other methods. If at least one of these comparisons is
not significant, no entry for that benchmark is bold.}
\centering
\rowcolors{2}{gray!10}{white}
\resizebox{\textwidth}{!}{
\begin{tabular}{lcccccc}
\toprule
& \multicolumn{3}{c}{\textbf{Relative $L^2$ error}}
& \multicolumn{3}{c}{\textbf{Runtime [min]}} \\
\cmidrule(lr){2-4}\cmidrule(lr){5-7}
\textbf{Benchmark}  & MLP & Modified MLP & PirateNet & MLP & Modified MLP & PirateNet \\
\midrule
Wave
&  ${3.56 \pm 1.3 \times 10^{-5}}$ 
&  ${1.67 \pm 0.4 \times 10^{-5}}$ 
&  ${1.83 \pm 0.2 \times 10^{-5}}$ 
&  $\mathbf{20.98}$ 
&  $30.97$ 
&  $37.43$ 
\\

KdV
&  ${3.15 \pm 0.2 \times 10^{-4}}$ 
&  ${3.37 \pm 0.1 \times 10^{-4}}$ 
&  ${2.90 \pm 0.4 \times 10^{-4}}$ 
&  $\mathbf{17.22}$ 
&  $21.54$ 
&  $30.74$ 
\\

RT
&  ${1.00 \pm 0.1 \times 10^{-2}}$ 
&  ${1.35 \pm 0.6 \times 10^{-2}}$ 
&  ${1.44 \pm 0.4 \times 10^{-2}}$ 
&  $\mathbf{48.18}$ 
&  $60.93$ 
&  $94.52$ 
\\

Heat
&  $\mathbf{{3.84 \pm 0.2 \times 10^{-4}}}$ 
&  ${6.89 \pm 0.5 \times 10^{-4}}$ 
&  ${7.06 \pm 0.5 \times 10^{-4}}$ 
&  $\mathbf{8.17}$ 
&  $11.14$ 
&  $16.23$ 
\\

AC
&  ${2.29 \pm 0.3 \times 10^{-5}}$ 
&  ${2.29 \pm 0.3 \times 10^{-5}}$ 
&  ${2.72 \pm 0.6 \times 10^{-5}}$ 
&  $\mathbf{15.84}$ 
&  $19.06$ 
&  $27.98$ 
\\

Burgers
&  ${5.63 \pm 1.2 \times 10^{-6}}$ 
&  ${5.26 \pm 1.2 \times 10^{-6}}$ 
&  ${4.51 \pm 1.2 \times 10^{-6}}$ 
&  $\mathbf{11.79}$ 
&  $15.22$ 
&  $20.20$ 
\\

GL
&  $\mathbf{{1.12 \pm 0.2 \times 10^{-2}}}$ 
&  ${3.34 \pm 0.8 \times 10^{-2}}$ 
&  ${1.98 \pm 0.4 \times 10^{-2}}$ 
&  $\mathbf{44.05}$ 
&  $55.17$ 
&  $87.52$ 
\\

Poisson
&  ${2.10 \pm 0.4 \times 10^{-5}}$ 
&  ${9.28 \pm 1.1 \times 10^{-6}}$ 
&  ${1.19 \pm 0.3 \times 10^{-5}}$ 
&  $\mathbf{8.12}$ 
&  $10.88$ 
&  $16.14$ 
\\

\bottomrule
\end{tabular}
}
\label{tab:big_nn}
\end{table}

\paragraph{Discussion.}
The results across all three sizes tell a surprisingly consistent story that
only partially supports the claim. While architecture choice does influence
performance -- Modified MLP reliably outperforms on problems with smooth,
oscillatory solutions (Wave, AC, Poisson across most sizes), and the plain MLP
tends to dominate on steeper or more irregular problems (KdV, RT, GL at small
and large scale) -- the expected advantage of more sophisticated architectures
at larger parameter counts does not materialize. PirateNet, despite its
inductive biases designed for difficult PINN problems~\citep{PirateNet},
does not achieve consistent gains over the plain MLP at any scale, winning in
only $3$ out of $24$ benchmark--size combinations while incurring a runtime
penalty of roughly $1.8$--$2\times$ relative to MLP.

The most pronounced architectural sensitivity appears on GL and RT, the most
challenging benchmarks in our suite, yet even there the rankings are not stable
across sizes: PirateNet achieves the best GL result at medium scale but falls
behind at both small and large scale.

Taken together, the evidence supports the claim only weakly and in a
problem-dependent manner: architecture choice produces measurable but
inconsistent accuracy differences, with no single architecture dominating
across benchmarks and sizes. Crucially, the plain MLP remains competitive or
superior at large scale while being substantially cheaper to train,
which justifies the plain MLP as the anchor.

\newpage

\subsection{Optimizer, Precision and Architecture Ablation}
\label{Appendix:ablation_nn_optim}
Recall that the optimizer and network study (Section~\ref{experiments:optimizers}) compares two configurations, P1 and P2, that differ in three axes at once: the network architecture, the numerical precision, and the optimizer scheme.
Specifically, P1 is a PirateNet trained with SOAP in float32, and P2 is an MLP trained with the two-stage scheme of SOAP followed by SSBroyden2 in float64 (Fig.~\ref{fig:evalcard_optimizer}).
The comparison of P1 and P2 therefore reflects the joint effect of all three axes, and the difference in relative $L^2$ error cannot be attributed to a single axis from this comparison alone.
To separate the effects of the three axes, we add three configurations that each differ from P1 or from P2 in a single axis.
Concretely, these are a PirateNet trained with SOAP in float64, an MLP trained with SOAP in float32, and an MLP trained with SOAP in float64.
The first two differ from P1 in the numerical precision and in the network architecture, respectively, and the third differs from P2 in the optimizer scheme.

Each experiment is a corner of a cube $\{\text{MLP}, \text{PirateNet}\} \times \{\text{float32}, \text{float64}\} \times \{\text{SOAP}, \text{SOAP}\to\text{SSBroyden2}\}$.
Two corners of the cube are joined by an \emph{edge} if they differ in a single axis.

A comparison along an edge isolates the effect of that axis if the two arms agree in all settings outside the cube.
We omit the corners that are combination of the two-stage scheme with a PirateNet or with float32.
These corners are not feasible, for the reasons given in the last paragraph of this subsection.
The relative $L^2$ errors and the runtimes of the five arms on the nine benchmarks of Appendix~\ref{Appendix:benchmarks} are reported in Tables~\ref{tab:ablation_error} and~\ref{tab:ablation_runtime}, respectively.

\begin{table}[h!]
\caption{Median and standard deviation over seeds of the relative $L^2$ errors for the five arms of the single-axis ablation evaluated on the nine PDE benchmark problems. Arms marked $\dagger$ are P1 and P2 of the optimizer and network study (Fig.~\ref{fig:evalcard_optimizer}).}
\label{tab:ablation_error}
\centering
\rowcolors{2}{gray!10}{white}
\resizebox{\textwidth}{!}{
\begin{tabular}{l|ccccc}
\toprule
Benchmark
& \shortstack{PirateNet, SOAP\\float32 (P1)$^{\dagger}$}
& \shortstack{PirateNet, SOAP\\float64}
& \shortstack{MLP, SOAP\\float32}
& \shortstack{MLP, SOAP\\float64}
& \shortstack{MLP, SOAP$\to$SSBroyden2\\float64 (P2)$^{\dagger}$} \\
\midrule
Wave
&  ${2.82 \pm 0.8 \times 10^{-5}}$
&  ${2.23 \pm 0.28 \times 10^{-5}}$
&  ${7.72 \pm 2.58 \times 10^{-5}}$
&  ${6.90 \pm 1.44 \times 10^{-5}}$
&  $\mathbf{2.36 \pm 0.8 \times 10^{-7}}$
\\
KdV
&  ${1.93 \pm 0.1 \times 10^{-4}}$
&  ${1.33 \pm 0.11 \times 10^{-4}}$
&  ${4.81 \pm 0.39 \times 10^{-4}}$
&  ${5.35 \pm 0.38 \times 10^{-4}}$
&  $\mathbf{2.31 \pm 1.2 \times 10^{-5}}$
\\
RT
&  ${8.80 \pm 0.8 \times 10^{-3}}$
&  ${8.76 \pm 0.06 \times 10^{-3}}$
&  ${1.11 \pm 0.15 \times 10^{-2}}$
&  ${1.14 \pm 0.24 \times 10^{-2}}$
&  ${8.61 \pm 0.1 \times 10^{-3}}$
\\
Heat
&  ${2.62 \pm 0.2 \times 10^{-4}}$
&  ${3.25 \pm 0.21 \times 10^{-4}}$
&  ${1.52 \pm 0.42 \times 10^{-4}}$
&  ${2.12 \pm 1.27 \times 10^{-4}}$
&  $\mathbf{1.25 \pm 0.6 \times 10^{-8}}$
\\
AC
&  ${2.04 \pm 0.5 \times 10^{-5}}$ 
&  --$^{a}$ 
&  ${3.61 \pm 0.98 \times 10^{-5}}$ 
&  ${6.65 \pm 2.01 \times 10^{-5}}$ 
&  $\mathbf{{9.20 \pm 3.4 \times 10^{-6}}}$ 
\\
Burgers
&  ${2.51 \pm 0.9 \times 10^{-5}}$ 
&  ${2.32 \pm 0.77 \times 10^{-5}}$ 
&  ${4.21 \pm 2.99 \times 10^{-5}}$ 
&  ${6.56 \pm 2.76 \times 10^{-5}}$ 
&  $\mathbf{{8.59 \pm 10.8 \times 10^{-8}}}$ 
\\
GL
&  ${4.77 \pm 1.6 \times 10^{-3}}$
&  --$^{a}$
&  ${9.58 \pm 0.61 \times 10^{-3}}$
&  ${1.49 \pm 0.04 \times 10^{-2}}$
&  ${3.68 \pm 1.2 \times 10^{-3}}$
\\
Poisson
&  ${4.82 \pm 0.4 \times 10^{-6}}$
&  ${1.69 \pm 0.24 \times 10^{-5}}$
&  ${4.94 \pm 3.31 \times 10^{-5}}$
&  ${5.76 \pm 2.25 \times 10^{-5}}$
&  $\mathbf{1.14 \pm 0.4 \times 10^{-7}}$
\\
KS
&  ${4.51 \pm 0.7 \times 10^{-2}}$
&  --$^{a}$
&  ${1.30 \pm 0.48 \times 10^{-1}}$
&  ${1.15 \pm 0.38\times 10^{-1}}$
&  $\mathbf{7.26 \pm 3.6 \times 10^{-3}}$
\\
\bottomrule
\end{tabular}}
\par\vspace{0.3em}
{\footnotesize\raggedright $^{a}$Training diverged. \par}
\end{table}

\begin{table}[h!]
\caption{Median runtime [min] for the five arms of the single-axis ablation evaluated on the nine PDE benchmark problems. Arms marked $\dagger$ are P1 and P2 of the optimizer and network study (Fig.~\ref{fig:evalcard_optimizer}).}
\label{tab:ablation_runtime}
\centering
\rowcolors{2}{gray!10}{white}
\resizebox{\textwidth}{!}{
\begin{tabular}{l|ccccc}
\toprule
Benchmark
& \shortstack{PirateNet, SOAP\\float32 (P1)$^{\dagger}$}
& \shortstack{PirateNet, SOAP\\float64}
& \shortstack{MLP, SOAP\\float32}
& \shortstack{MLP, SOAP\\float64}
& \shortstack{MLP, SOAP$\to$SSBroyden2\\float64 (P2)$^{\dagger}$} \\
\midrule
Wave    & $111.22$  & $187$    & $11$  & $16$ & $4.65$  \\
KdV     & $88.16$   & $88$     & $16$  & $14$ & $7.39$  \\
RT      & $513.50$  & $708$    & $89$  & $99$ & $70.27$ \\
Heat    & $49.39$   & $76$     & $8$   & $7$  & $7.29$  \\
AC      & $297.46$  & --$^{a}$ & $30$  & $41$ & $9.68$ \\
Burgers & $41.56$   & $94$     & $8$   & $9$  & $3.10$  \\
GL      & $519.22$  & --$^{a}$ & $66.7$ & $76.1$ & $18.02$ \\
Poisson & $25.30$   & $62$     & $4$   & $6$  & $5.04$  \\
KS      & $1100.85$ & --$^{a}$ & $191$ & $176$ & $70.33$ \\
\bottomrule
\end{tabular}}
\par\vspace{0.3em}
{\footnotesize\raggedright $^{a}$Training diverged; runtime not reported.\par}
\end{table}

First, none of the error reduction from P1 to P2 is attributable to the change from float32 to float64 as a main effect.
For the MLP trained with SOAP, float64 leaves the relative $L^2$ error essentially unchanged or marginally higher (Table~\ref{tab:ablation_error}).
More precisely, it raises the error on seven of the nine benchmarks (KdV, RT, Heat, AC, Burgers, GL, Poisson) and lowers it on Wave and KS.
For the PirateNet trained with SOAP, float64 changes the error by at most small factors in either direction wherever training completes.
Specifically, it lowers the error on KdV, raises it on Heat and Poisson, and leaves it within one standard deviation of P1 on Wave, RT, and Burgers.
However, the decrease on KdV is specific to the PirateNet: for the MLP, which P2 uses, float64 raises the error on KdV.
Three failures occur for the PirateNet in float64. On Allen–Cahn the physics-informed initialisation's least-squares fit retains near-singular features and produces output weights of order 1e7.

Second, the MLP trained with SOAP alone in float64 differs from P2 only in the optimizer scheme.
In terms of accuracy, replacing SOAP alone by the two-stage scheme (SOAP followed by SSBroyden2) reduces the relative $L^2$ error of this MLP by a factor between $1.3$ on RT and $1.7 \times 10^{4}$ on Heat.
On eight benchmarks, this factor exceeds the factor between the errors of P1 and P2, and on Heat it is nearly as large.

In terms of runtime, replacing SOAP alone by the two-stage scheme reduces the runtime on GL, Wave, KdV, RT, AC, Burgers, Poisson, and KS and leaves it essentially unchanged on Heat.
For example, the runtime decreases from $9$ to $3.1$~min on Burgers and from $176$ to $70$~min on KS.

Third, the change of architecture alone is not the source of the error reduction from P1 to P2.
In particular, with SOAP alone in float64, the MLP has a higher error than the PirateNet on five of the six benchmarks on which both runs completed (Wave, KdV, RT, Burgers, Poisson) and a lower error on Heat.
However, once the quasi-Newton SSBroyden2 stage is added, the same MLP reaches errors up to four orders of magnitude lower than under SOAP alone.

Against the PirateNet trained with SOAP in float64, the same MLP in P2 attains a lower error on all six of these benchmarks.
We therefore attribute the higher error of the MLP under SOAP alone, relative to the PirateNet, to the difficulty of training the MLP with a first-order optimizer, not to a limited expressivity.

Taken together, the error reduction from P1 to P2 is attributable to the quasi-Newton stage.
In terms of precision, float64 is not an independent contributor to this reduction but a prerequisite of the quasi-Newton stage.
We note that \citet{SOAPWang} exclude quasi-Newton methods as impractical for complex PDEs in part because these methods require float64.
In terms of memory, the moderate size of the MLP of P2 is what makes the quasi-Newton stage feasible.
Specifically, SSBroyden2, the optimizer of the quasi-Newton stage, stores a dense approximation of the inverse Hessian, so the memory of SSBroyden2 grows as $\mathcal{O}(d^2)$ in the number $d$ of trainable parameters; see Appendix~\ref{Appendix:optimizer}.
This quadratic growth restricts SSBroyden2 to networks with a moderate number of trainable parameters.
For the MLP of P2, the peak VRAM of the quasi-Newton stage lies between $10.23$ and $10.68$~GiB across the nine benchmarks, which is well within the memory of modern accelerators.
The peak VRAM per benchmark is reported in Appendix~\ref{Appendix:vram_results}.

\section{Video RAM Results}
\label{Appendix:vram_results}
The peak Video RAM usage reported in Table~\ref{tab:vram_results} is obtained using the built-in memory tracking functionality of the \texttt{JAX} library.
Specifically, we report the maximum value returned by
\texttt{jax.local\_devices()[0].memory\_stats()['peak\_bytes\_in\_use']},
which corresponds to the peak VRAM consumption observed during training.
All measurements are collected after completion of the training run.

The reported VRAM values for the PirateNet--SOAP reference pipeline are
produced by executing the publicly released code provided by the original
authors.
The VRAM measurements for the proposed pipeline can be reproduced by
running our published implementation under the same experimental setup.

\begin{table}[h]
\centering
\caption{Peak Video RAM usage occupied by the training pipelines in JAX.}
\rowcolors{2}{gray!10}{white}
\label{tab:vram_results}
\begin{tabular}{lccc}
\toprule
\textbf{Benchmark} & \textbf{PirateNet-SOAP} & \multicolumn{2}{c}{\textbf{SOAP-SSBroyden}} \\
&& SOAP phase & SSBroyden Phase \\
\midrule
Burgers 
& $1.46$ GiB & $0.25$ GiB & $10.23$ GiB \\
Korteweg--de Vries 
& $1.53$ GiB & $0.49$ GiB & $10.55$ GiB \\
Allen--Cahn 
& $1.95$ GiB & $0.41$ GiB & $10.23$ GiB \\
Kuramoto--Sivashinsky 
& $2.76$ GiB & $0.70$ GiB & $10.31$ GiB \\
Wave 
& $3.23$ GiB & $0.79$ GiB & $10.55$ GiB \\
Poisson 
& $3.71$ GiB & $0.41$ GiB & $10.32$ GiB\\
Heat
& $6.67$ GiB & $0.54$ GiB & $10.36$ GiB \\
Ginzburg--Landau 
& $8.89$ GiB & $1.00$ GiB & $10.61$ GiB \\
Rayleigh--Taylor instability 
& $19.22$ GiB & $1.94$ GiB & $10.68$ GiB \\

\bottomrule
\end{tabular}
\end{table}

The other studies, which use networks of a size similar to that of the SOAP-SSBroyden pipeline, have a VRAM consumption very similar to that of this pipeline during the SOAP phase.

Notably, the SOAP-SSBroyden pipeline consumes a nearly constant amount of memory in the SSBroyden phase, which is dominated by the dense second-order information. However, on challenging problems like the Rayleigh--Taylor benchmark, the PirateNet-SOAP pipeline requires even more memory due to its number of trainable parameters and the amount of equations considered in the benchmark.

\section{PINNpipe: The code package}
\label{Appendix:PINNpipe}

The experiments in this paper are implemented in \emph{PINNpipe} (\url{https://github.com/KevinvonBargen/pinnpipe}), a
JAX/Flax-based evaluation framework for physics-informed neural networks.
The framework was built specifically to support the comparative study
described in the main text: it treats optimizer choice, neural
architecture, adaptive sampling strategy, and loss-balancing scheme as
independently configurable components, allowing any combination of the
four to be evaluated on any benchmark without code changes. This design
directly reflects the orthogonality argument in
Section~\ref{sec:related}: where prior comparative studies conflate
components by treating them as a single comparison axis, PINNpipe makes
the four axes structurally separate at the API level.

The package is organized into a core library and a benchmark suite. The
core library provides base abstractions for PINN models, train states,
evaluators, and training loops, together with concrete
implementations of the components studied in the main text. Supported
architectures include standard multilayer perceptrons, modified MLPs,
and PirateNet-style models. The optimization interface covers
first-order and quasi-second-order training with Adam and SOAP, as well as SOAP/Broyden-style
hybrids. Adaptive domain sampling methods include Random-R,
grid sampling, RAD~\citep{RAR}, LAS~\citep{jeong2025mitigating}, R3~\citep{r3},
and vRBA~\citep{Toscano.2026}; loss-weighting schemes include fixed
weighting, GradNorm-based balancing~\citep{wang2023expert}, learning-rate annealing~\citep{ModifiedMLP},
ReLoBRaLo~\citep{relobralo}, SoftAdapt~\citep{SoftAdapt}, and
GradNorm-MTL~\citep{Gradnorm_mtl}.

The benchmark suite covers Allen--Cahn, Burgers, Ginzburg--Landau, Heat, KdV,
Kuramoto--Sivashinsky, Poisson, Rayleigh--Taylor, and wave problems.
Each benchmark directory exposes the same five experiment entry points
---neural architecture sweeps, loss-balancing sweeps, sampler sweeps,
SOAP optimizer sweeps, and SOAP/Broyden optimizer sweeps---so that
every benchmark can be exercised along every axis with identical
tooling. Problem-specific model definitions, evaluators,
configurations, and data-loading utilities live inside each benchmark
folder. Reference data required by the loaders is included where
needed.

\paragraph{Dependencies and prior artifacts.}
PINNpipe is built on JAX, NumPy~\citep{harris2020array}, and
Matplotlib~\citep{Hunter2007}. 

\section{Visual Results and error plots for the SSBroyden pipeline}
To complement the quantitative error metrics, we visualize the predictions produced by the SOAP-SSBroyden pipeline on the considered benchmark problems. Each figure compares the final learned solution with the reference solution and includes an absolute error map. Together, these results illustrate how accurately the trained models recover the underlying dynamics across the different PDE examples.

\begin{figure}[H]
  \centering
  \includegraphics[width=1\textwidth]{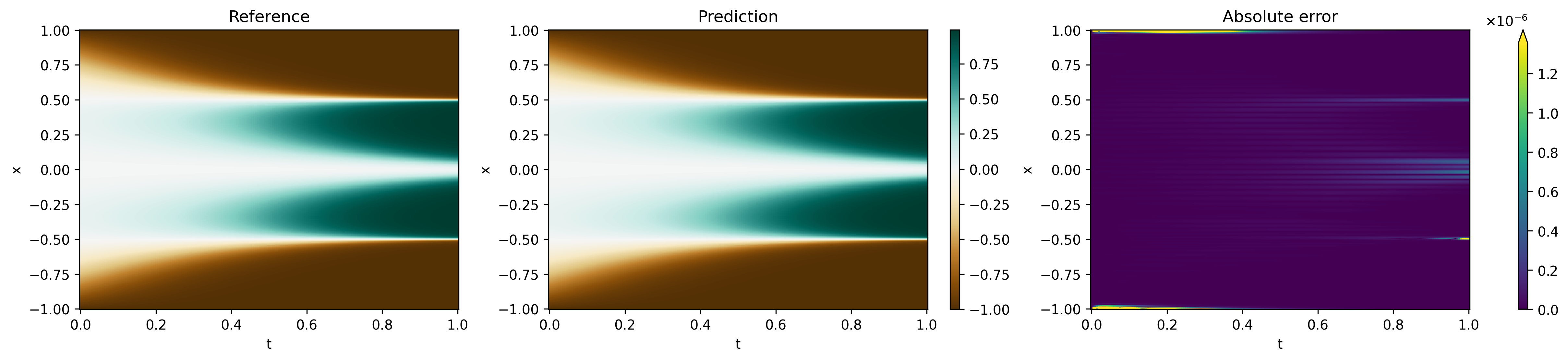}
  \caption{{\em Allen--Cahn Equation}. From left to right: reference solution, final prediction, and absolute error for the Allen--Cahn equation. The reference and prediction share a common color scale for direct comparison.}

  \label{fig:ac_pred}
\end{figure}

\begin{figure}[H]
  \centering
  \includegraphics[width=1\textwidth]{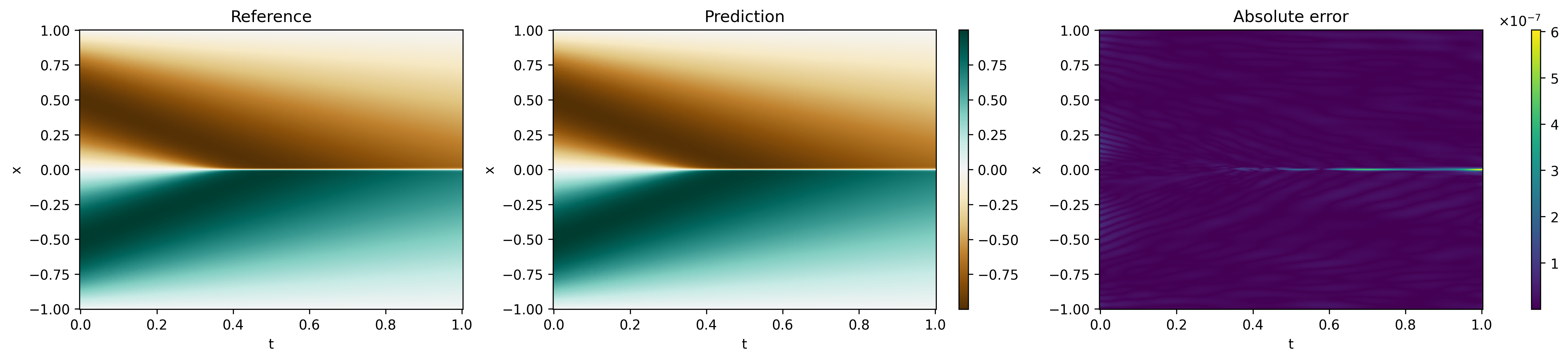}
  \caption{{\em Burgers Equation}. From left to right: reference solution, final prediction, and absolute error for the Burgers equation. The reference and prediction share a common color scale for direct comparison.}

  \label{fig:burgers_pred}
\end{figure}

\begin{figure}[H]
  \centering
  \includegraphics[width=1\textwidth]{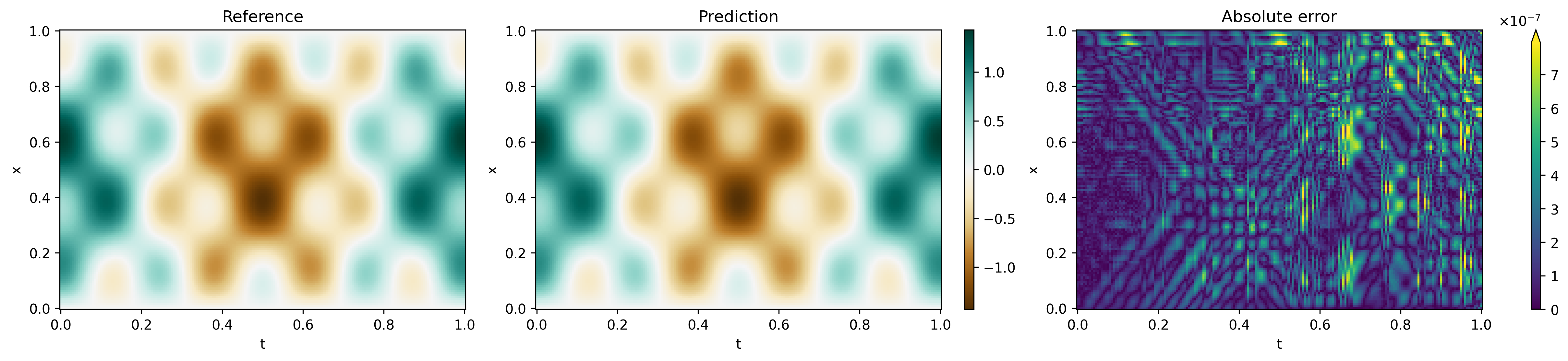}
  \caption{{\em Wave Equation}. From left to right: reference solution, final prediction, and absolute error for the Wave equation. The reference and prediction share a common color scale for direct comparison.}

  \label{fig:wave_pred}
\end{figure}

\begin{figure}[H]
  \centering
  \includegraphics[width=1\textwidth]{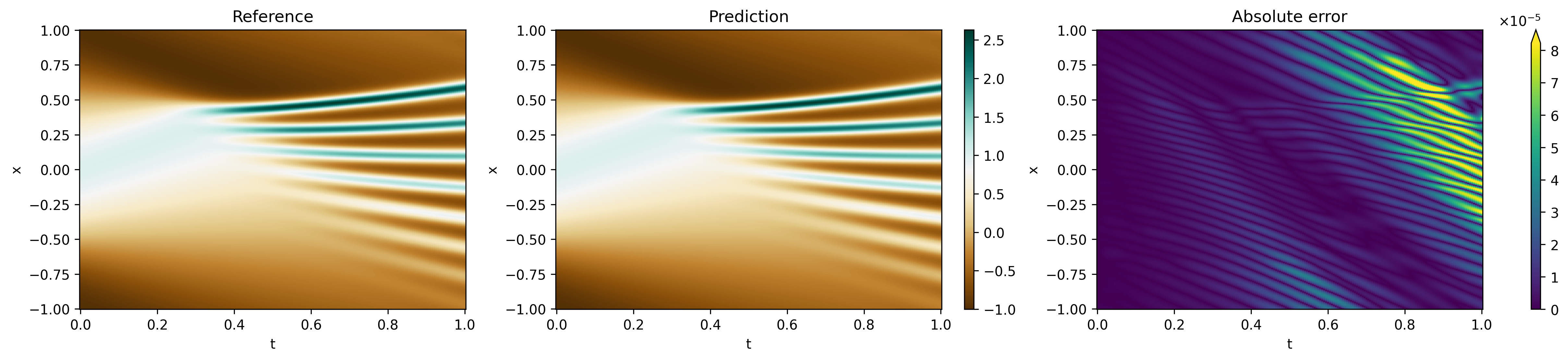}
  \caption{{\em Korteweg--de Vries Equation}. From left to right: reference solution, final prediction, and absolute error for the KdV equation. The reference and prediction share a common color scale for direct comparison.}
  \label{fig:kdv_pred}
\end{figure}

\begin{figure}[H]
  \centering
  \includegraphics[width=1\textwidth]{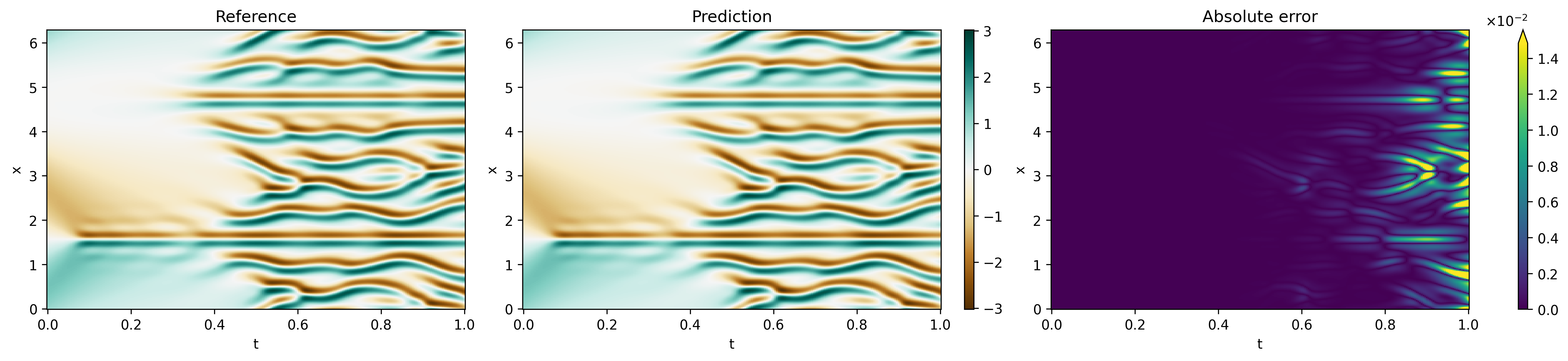}
  \caption{{\em Kuramoto--Sivashinsky Equation}. From left to right: reference solution, final prediction, and absolute error for the KS equation. The reference and prediction share a common color scale for direct comparison.}
  \label{fig:ks_pred}
\end{figure}

\begin{figure}[H]
  \centering
  \includegraphics[width=1\textwidth]{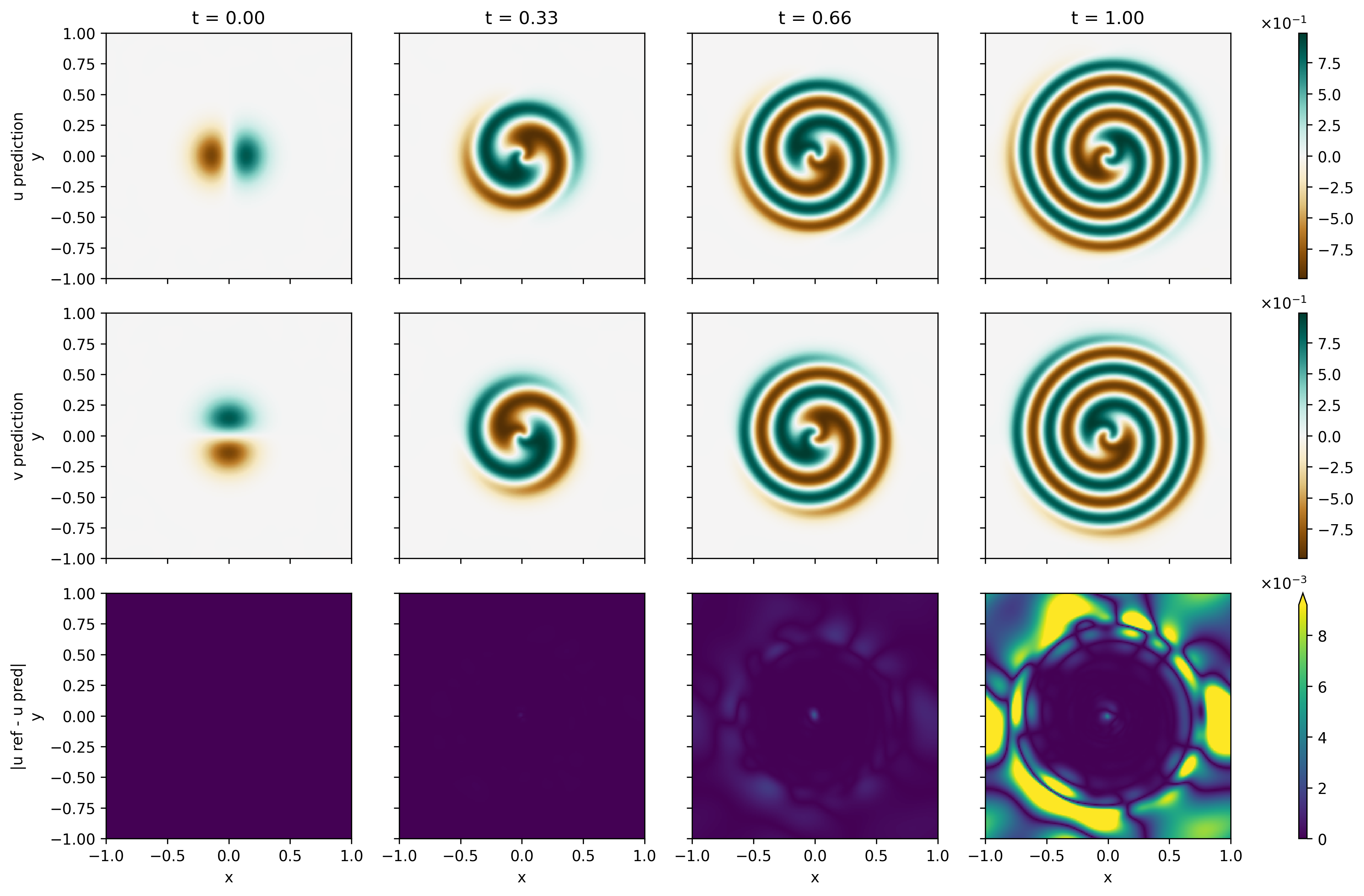}
  \caption{{\em Ginzburg--Landau Equation}. The columns show four selected time snapshots from left to right. The rows show the predicted components $u,v$, and the absolute $u$ error.}
  \label{fig:gl_pred}
\end{figure}

\begin{figure}[H]
  \centering
  \includegraphics[width=1\textwidth]{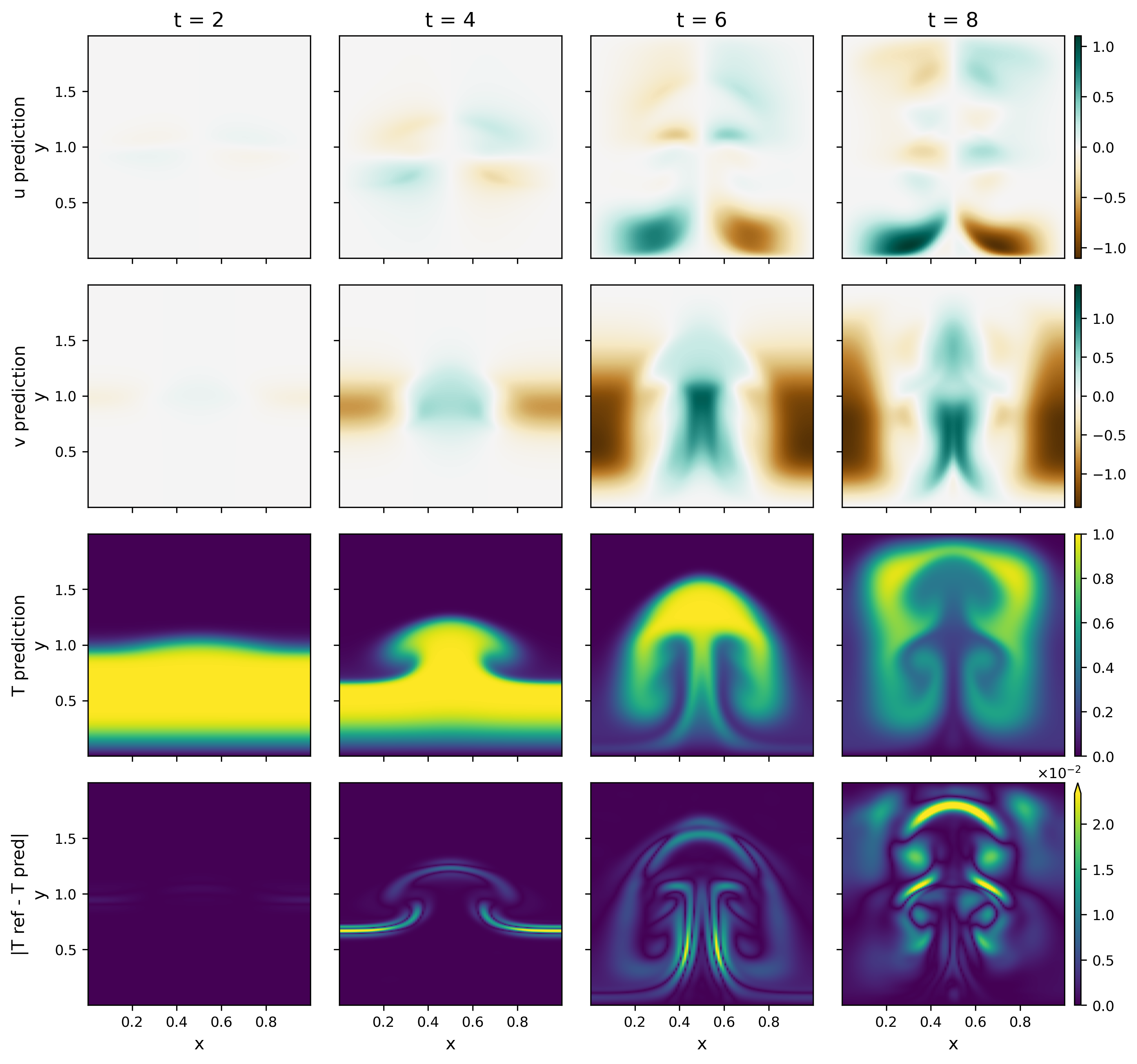}
  \caption{{\em Rayleigh--Taylor Instability}. From left to right, the columns show selected time snapshots of the flow. From top to bottom, the rows show the predicted horizontal velocity $u$, the predicted vertical velocity $v$, the predicted temperature field $T$, and the absolute temperature error.}

  \label{fig:rt_pred}
\end{figure}

\newpage
\section*{NeurIPS Paper Checklist}

\begin{enumerate}

\item {\bf Claims}
    \item[] Question: Do the main claims made in the abstract and introduction accurately reflect the paper's contributions and scope?
    \item[] Answer: \answerYes{} %
    \item[] Justification: The abstract and introduction accurately reflect the paper's contributions and scope: introducing a new evaluation standard for PINNs, presenting a four-component decomposition of the PINN training pipeline, illustrating it along multiple studies and providing the PINNpipe code library as well as revising commonly held assumptions.

\item {\bf Limitations}
    \item[] Question: Does the paper discuss the limitations of the work performed by the authors?
    \item[] Answer: \answerYes{} %
    \item[] Justification: The manuscript discusses limitations of the benchmarks covered as well as the specialization to PINNs solving forward problems.

\item {\bf Theory assumptions and proofs}
    \item[] Question: For each theoretical result, does the paper provide the full set of assumptions and a complete (and correct) proof?
    \item[] Answer: \answerYes{} %
    \item[] Justification: The decomposition of the four-component training pipeline of PINNs is based on established work in the literature.

    \item {\bf Experimental result reproducibility}
    \item[] Question: Does the paper fully disclose all the information needed to reproduce the main experimental results of the paper to the extent that it affects the main claims and/or conclusions of the paper (regardless of whether the code and data are provided or not)?
    \item[] Answer: \answerYes{} %
    \item[] Justification: The manuscript contains all information needed to reproduce the experiments: the EvalCard of each study (Section~\ref{sec:experiments}, Appendix~\ref{Appendix:study_results}), the benchmark definitions and the provenance and accuracy of their reference solutions (Appendix~\ref{Appendix:benchmarks}, Appendix~\ref{Appendix:reference_solutions}), the training settings (Appendix~\ref{Appendix:technicals-pdes}) and the optimizer, precision and architecture ablation (Appendix~\ref{Appendix:ablation_nn_optim}). The public PINNpipe repository (\url{https://github.com/KevinvonBargen/pinnpipe}) additionally contains all code, the instructions to reproduce the results and the reference-solution generation scripts.

\item {\bf Open access to data and code}
    \item[] Question: Does the paper provide open access to the data and code, with sufficient instructions to faithfully reproduce the main experimental results, as described in supplemental material?
    \item[] Answer: \answerYes{} %
    \item[] Justification: The code is publicly available in the PINNpipe repository (\url{https://github.com/KevinvonBargen/pinnpipe}) under an open-source license, together with instructions to reproduce all studies, the reference data required by the benchmarks, and the reference-solution generation scripts, so that the reference solutions and their accuracy can be regenerated independently (Appendix~\ref{Appendix:PINNpipe}, Appendix~\ref{Appendix:reference_solutions}).

\item {\bf Experimental setting/details}
    \item[] Question: Does the paper specify all the training and test details (e.g., data splits, hyperparameters, how they were chosen, type of optimizer) necessary to understand the results?
    \item[] Answer: \answerYes{} %
    \item[] Justification: The EvalCard of each study states the surrounding pipeline, the compared methods and the evaluation conditions. The method hyperparameters are the values proposed by the introducing authors, identical across benchmarks and seeds (Appendix~\ref{Appendix:dia}, Appendix~\ref{Appendix:lossbalancing}); the training and optimizer settings, the reference solutions and the optimizer, precision and architecture ablation are given in Appendix~\ref{Appendix:technicals-pdes}, Appendix~\ref{Appendix:reference_solutions} and Appendix~\ref{Appendix:ablation_nn_optim}.

\item {\bf Experiment statistical significance}
    \item[] Question: Does the paper report error bars suitably and correctly defined or other appropriate information about the statistical significance of the experiments?
    \item[] Answer: \answerYes{} %
    \item[] Justification: Every reported error is given as the median $\pm$ the standard deviation over independent seeds, where each seed is a complete training run under the stated conditions; the number of seeds of each study is stated in its EvalCard. A best entry is highlighted in bold only if its margin over every other method in its row is significant under the exact two-sided Mann--Whitney $U$ test on the per-seed relative $L^2$ errors, with Holm correction over the comparisons in that row, at $\alpha = 0.05$; otherwise no entry in the row is bold. Bold runtimes are descriptive and mark the lowest median (Section~\ref{sec:protocol}, Section~\ref{sec:Evalcard}).

\item {\bf Experiments compute resources}
    \item[] Question: For each experiment, does the paper provide sufficient information on the computer resources (type of compute workers, memory, time of execution) needed to reproduce the experiments?
    \item[] Answer: \answerYes{} %
    \item[] Justification: In each study the paper specifies the GPU used (H100).
    
\item {\bf Code of ethics}
    \item[] Question: Does the research conducted in the paper conform, in every respect, with the NeurIPS Code of Ethics \url{https://neurips.cc/public/EthicsGuidelines}?
    \item[] Answer: \answerYes{} %
    \item[] Justification: The research presented conforms to the NeurIPS Code of Ethics. The manuscript focuses on evaluation design, standards, and comparisons, as well as a common definition of the PINN training pipeline, and raises no ethical concerns.

\item {\bf Broader impacts}
    \item[] Question: Does the paper discuss both potential positive societal impacts and negative societal impacts of the work performed?
    \item[] Answer: \answerYes{} %
    \item[] Justification: While the work on an evaluation standard for PINNs solving forward problems is specific, it can be used as a guide to develop standards for the broader field of Physics-Informed Machine Learning and beyond.
    
\item {\bf Safeguards}
    \item[] Question: Does the paper describe safeguards that have been put in place for responsible release of data or models that have a high risk for misuse (e.g., pre-trained language models, image generators, or scraped datasets)?
    \item[] Answer: \answerNA{} %
    \item[] Justification:  The paper does not involve the release of sensitive datasets, or tools with a high risk of misuse. It focuses on physics-informed neural networks without foreseeable harmful applications.

\item {\bf Licenses for existing assets}
    \item[] Question: Are the creators or original owners of assets (e.g., code, data, models), used in the paper, properly credited and are the license and terms of use explicitly mentioned and properly respected?
    \item[] Answer: \answerYes{} %
    \item[] Justification: Appendix~\ref{Appendix:PINNpipe} cites and acknowledges the software and data PINNpipe builds on: JAX/Flax, NumPy and Matplotlib, the \texttt{jaxpi} framework and PINNacle for problem definitions and the Rayleigh--Taylor reference data, and Chebfun for the reference solutions.

\item {\bf New assets}
    \item[] Question: Are new assets introduced in the paper well documented and is the documentation provided alongside the assets?
    \item[] Answer: \answerYes{} %
    \item[] Justification: The new asset is the PINNpipe package, publicly released at \url{https://github.com/KevinvonBargen/pinnpipe}. Appendix~\ref{Appendix:PINNpipe} documents its structure, the implemented methods, the benchmark suite, the experiment entrypoints and the reference-solution generation scripts.

\item {\bf Crowdsourcing and research with human subjects}
    \item[] Question: For crowdsourcing experiments and research with human subjects, does the paper include the full text of instructions given to participants and screenshots, if applicable, as well as details about compensation (if any)? 
    \item[] Answer: \answerNA{} %
    \item[] Justification: The paper does not involve any human subjects or crowdsourcing in data
collection or evaluation.

\item {\bf Institutional review board (IRB) approvals or equivalent for research with human subjects}
    \item[] Question: Does the paper describe potential risks incurred by study participants, whether such risks were disclosed to the subjects, and whether Institutional Review Board (IRB) approvals (or an equivalent approval/review based on the requirements of your country or institution) were obtained?
    \item[] Answer: \answerNA{} %
    \item[] Justification:  The research does not involve human subjects, so IRB approval is not applicable.

\item {\bf Declaration of LLM usage}
    \item[] Question: Does the paper describe the usage of LLMs if it is an important, original, or non-standard component of the core methods in this research? Note that if the LLM is used only for writing, editing, or formatting purposes and does \emph{not} impact the core methodology, scientific rigor, or originality of the research, declaration is not required.
    \item[] Answer: \answerNA{} %
    \item[] Justification:  The paper does not indicate the use of LLMs as part of the core methods or
research.

\end{enumerate}

\end{document}